# Formal Properties of Language as Constraints on Neural Dynamics

Elliot Murphy[1,2]

**Affiliations**: 1. Vivian L. Smith Department of Neurosurgery, McGovern Medical School, UTHealth, Houston, TX, USA
2. Texas Institute for Restorative Neurotechnologies, UTHealth, Houston, TX, USA
**Permanent Address**: 1133 John Freeman Blvd, Houston, TX 77030, USA
**Contact**: elliot.murphy@uth.tmc.edu

**Abstract**
What must a neural system be capable of to implement language? Current research annotates stimuli with linguistic variables and tests which electrodes, voxels, or language-model layers predict neural activity. Yet predictive success leaves neural mechanisms under-constrained, and candidate composition rules may fail to preserve distinctions natural language requires. Here, we show that algebraic properties of language can constrain this search space by specifying invariants that candidate mechanisms must preserve, such as non-associative hierarchical grouping, binding that ignores the order of its parts, recursive closure, access to substructures, semantic role compatibility, and transitions through structured workspaces. This constraint-led approach is termed the Neural Admissibility Program (NAP). To illustrate it, syntactic structure building is analyzed algebraically and candidate mechanisms are proposed for each requirement: content-addressable workspace memory (e.g., expander Hopfield networks), graph-structured transient dynamics for scheduling structure-building operations (e.g., stable heteroclinic channels), and a phase-coupled sealing operation that records grouping. Simulations then show that a corrected form of the Marcolli–Berwick entropy-optimized binding gate preserves grouping only within a narrow band of commitment. As an alternative, we propose a novel neural binding operation we term 'Meld': two constituent populations converge through shared synapses, integrate sublinearly, and saturate. Meld is, to our knowledge, the closest neurally plausible composition law to the syntactic Merge computation. It preserves every invariant the NAP demands, is built from operations cortex is known to perform, and recovers hierarchical structure at every depth and temperature tested, decoding bracketing to within a few points of the best-performing rival law (ordered role-filler) while remaining order-blind. It predicts that the effective dimensionality of population activity separates alternative bracketings and that the composite depends on how far its constituents disagree. Importantly, the laws that decode bracketing most accurately are a priori inadmissible, indicating that neural decoding accuracy alone cannot adjudicate between mechanisms. By specifying how neural dynamics can remain faithful to linguistic structure, the NAP changes the criterion by which neural implementations of cognition are evaluated.

## 1. Introduction

A productive branch of modern cognitive neuroscience takes a range of naturalistic language stimuli (e.g., podcasts) and annotates these with linguistic (lexical, semantic, syntactic) and statistical variables. A temporal-response function, encoding model, representational similarity analysis, or language-model alignment analysis is then fitted, and the resulting maps are interpreted as evidence for the neural sensitivity of a region to a given linguistic property. This work has supplied indispensable spatiotemporal constraints, but has not yet delivered mechanistic theories of compositional syntax-semantics. Existing models can reconstruct words or continuous language from neural recordings, but predictive success alone does not identify the brain's representational format, compositional operations, or temporal dynamics (Brouwer, 2026). One major obstacle is that the relevant "parts list" of language has not been specified finely enough to support plausible linking hypotheses between neuroscience and linguistics (Poeppel et al., 2012).

Which properties of language should neuroscientists aim to explain? Consider a simple phrase like "the second green ball". Children usually select the second green item in an ordered array, not the second ball that also happens to be green (Hamburger & Crain, 1984). If the phrase were read as a linear string, "second" and "green" could combine conjunctively. English speakers instead converge on the hierarchical reading, showing that *constituency* is a necessary design feature of linguistic meaning (Mai & Martin, 2025; Yang et al., 2017): syntax "carves the path that semantics must blindly follow" (Uriagereka, 2008). The core feature of human language is not merely a list of features attached to a string (Chomsky, 1957; Chomsky et al., 2023; Dighiero-Brecht et al., 2026; Everaert et al., 2015; Friederici, 2017; Klinedinst, 2025; Senturia & Frank, 2023); rather, syntax is a generative system that builds objects with internal algebraic structure, and those properties can potentially help specify what a neural implementation must respect.

This strategy of using formal structure as a prior constraint on neural implementation, rather than inferring structure post hoc from neural data, is common in other domains of cognitive neuroscience. The hexagonal firing structure of entorhinal grid cells was understood, and in important respects anticipated, through the lens of the formal properties a metric for two-dimensional space must have. The toroidal topology of grid-cell population activity was later shown to be preserved across behavioral and sleep states (Constantinescu et al., 2016; Gardner et al., 2022; Hafting et al., 2005). The head-direction system is similarly understood through the ring attractor, a topological object whose structure was posited on computational grounds and subsequently confirmed in the Drosophila central complex (Kim et al., 2017). Efficient-coding and normative-Bayesian accounts derive predictions about population geometry from principles of information maximization and optimal inference (Barlow, 1972; Barlow & Levick, 1965; Simoncelli & Olshausen, 2001; Trenholm & Krishnaswamy, 2020). Sound localization offers a particularly striking case: Jeffress posited a delay-line and coincidence-detection architecture as the abstract solution to the interaural-time-difference problem in 1948, decades before the corresponding circuitry was identified in the barn owl's nucleus laminaris and found to instantiate the predicted place-code for azimuth almost exactly (Carr & Konishi, 1990; Jeffress, 1948). Meanwhile, maps of object space in inferotemporal cortex have successfully been built by embracing abstract dimensions like animacy (Bao et al., 2020). The same logic also shaped major advances in reinforcement learning, decision-making, and motor control (Gold & Shadlen, 2007; Montague et al., 1996; Ratcliff & McKoon, 2008; Schultz et al., 1997; Todorov & Jordan, 2002).

Using the formal properties of language to constrain admissible neural mechanisms is therefore continuous with a broader methodological tradition. When a capacity has rich formal structure, that structure constrains what its neural implementation must make dynamically available. Others have already offered constituency structure as a guide for language neuroscience (Ding et al., 2016; Gwilliams et al., 2025; Mai & Martin, 2025). Here, we extend this into a constraint-based (Ross, 2022, 2025) program for the neurobiology of language, which we will refer to as the Neural Admissibility Program (NAP). The guiding question of the NAP is not which neural signals correlate with linguistic variables, but which neural dynamics are admissible implementations of the formal invariants that define linguistic structure. Only by addressing these issues directly will we move from what van Bree terms the *premechanism stage* (exploring where and when mechanisms might be situated in space and time) to the *mechanism stage* that fills in the premechanism mold (exploring how computation emerges from organized interactions between parts) (Van Bree, 2024a) – where we can assume that a mechanism for a given linguistic phenomenon is defined by entities whose activities and interactions are organized in such a way that they produce that phenomenon (Glennan, 1996; Ross & Bassett, 2024).

We will assume that the relevant target for neurobiology is therefore not the string, but rather hierarchical constituency structure (Hornstein & Pietroski, 2009; Marcolli, Chomsky, et al., 2025; Marcolli, Huijbregts, et al., 2025; Pietroski, 2018). Recent work on free commutative non-associative magmas (Marcolli, Chomsky, et al., 2025), Hopf-algebraic workspaces (Marcolli & Skigin, 2025), colored operads and hypermagmas (Marcolli, Huijbregts, et al., 2025), and function-space embeddings (Marcolli & Berwick, 2026) provides an especially useful formal route into this problem, because each structure implies distinct neural admissibility conditions.

Since the 1990s, advances in neuroimaging have enabled researchers to explore *where* in the brain and *when* higher-order language processes are enabled (Fedorenko et al., 2024; Friederici, 2017; Matchin & Hickok, 2020). More advanced recording techniques, such as intracranial EEG, have provided acute spatiotemporal resolution – yet, even with these modern techniques, most of the advances for higher-order language have been refining the *where* and *when* questions, providing little headway into *how* natural language syntax and semantics are processed. We believe one of the reasons for the lack of mechanistic traction here is that the explanatory target is under-specified: 'sentence processing' and 'phrase meaning' are often not operationalized appropriately, and are not delineated into neurobiologically plausible units of representation or computation (Embick & Poeppel, 2015).

Large language models (LLMs) are now dominant in the modern cognitive neuroscience of language. Embeddings and surprisal estimates predict measurable neural variance during naturalistic comprehension, but predictive alignment alone gives little mechanistic traction (Dentella et al., 2024; Murphy et al., 2025): systems with different causal organizations can share the same mapping, and a model can predict neural activity while misrepresenting its generating mechanism (Guest et al., 2026; Ku et al., 2026; Qian et al., 2024). Deep models may align with comprehension by capturing distributional, semantic, contextual, or long-distance statistical regularities (Hadidi et al., 2026). Those facts are useful, but they do not show that grammar or meaning is identical with surface-context distributions. Nastase and colleagues (Nastase et al., 2026) exemplify the opposing, alignment-first program, moving from LLM-brain representational correspondence toward claims about shared computational principles and mechanistic models of language. Yet prediction, shared geometry, and mechanism are distinct: similar or linearly mappable population spaces can arise from common stimulus structure while leaving the causal operations that generate and use those states underdetermined (Murphy, 2026b).

Two general programs for the neurobiology of language can therefore be distinguished (Figure 1). One reads structure from neural data by fitting surface-statistical models post hoc (Caucheteux & King, 2022; Goldstein et al., 2024, 2025). The other (here, the NAP) treats formal properties as constraints any neural implementation must satisfy. The first has empirical scope, and both approaches can help identify neural signatures and feature spaces, but causal-mechanistic models ask in advance which formal properties the neural data must preserve. Indeed, much of the 'statistics-first' literature *already* adjudicates between explicit structural hypotheses (e.g., constituency versus dependency parsers; syntactic surprisal derived from formal grammars) and is in that respect structure-first in spirit, and the structure and statistics of natural language are jointly represented across cortical sites (Weissbart & Martin, 2024). Hence, we anticipate a mutually supportive approach by which both programs can help refine each other's regions of interest, feature spaces, and criteria for causal evidence. Data-driven models will remain indispensable for discovering candidate representations, controlling confounds, and comparing the neural adequacy of rival implementations.

As we outline below, the NAP has three commitments: (i) linguistic theory specifies formal invariants; (ii) a neural mechanism is admissible only if it preserves those invariants up to a stated equivalence; (iii) neural data adjudicate among admissible mechanisms rather than merely locating correlates of linguistic variables.

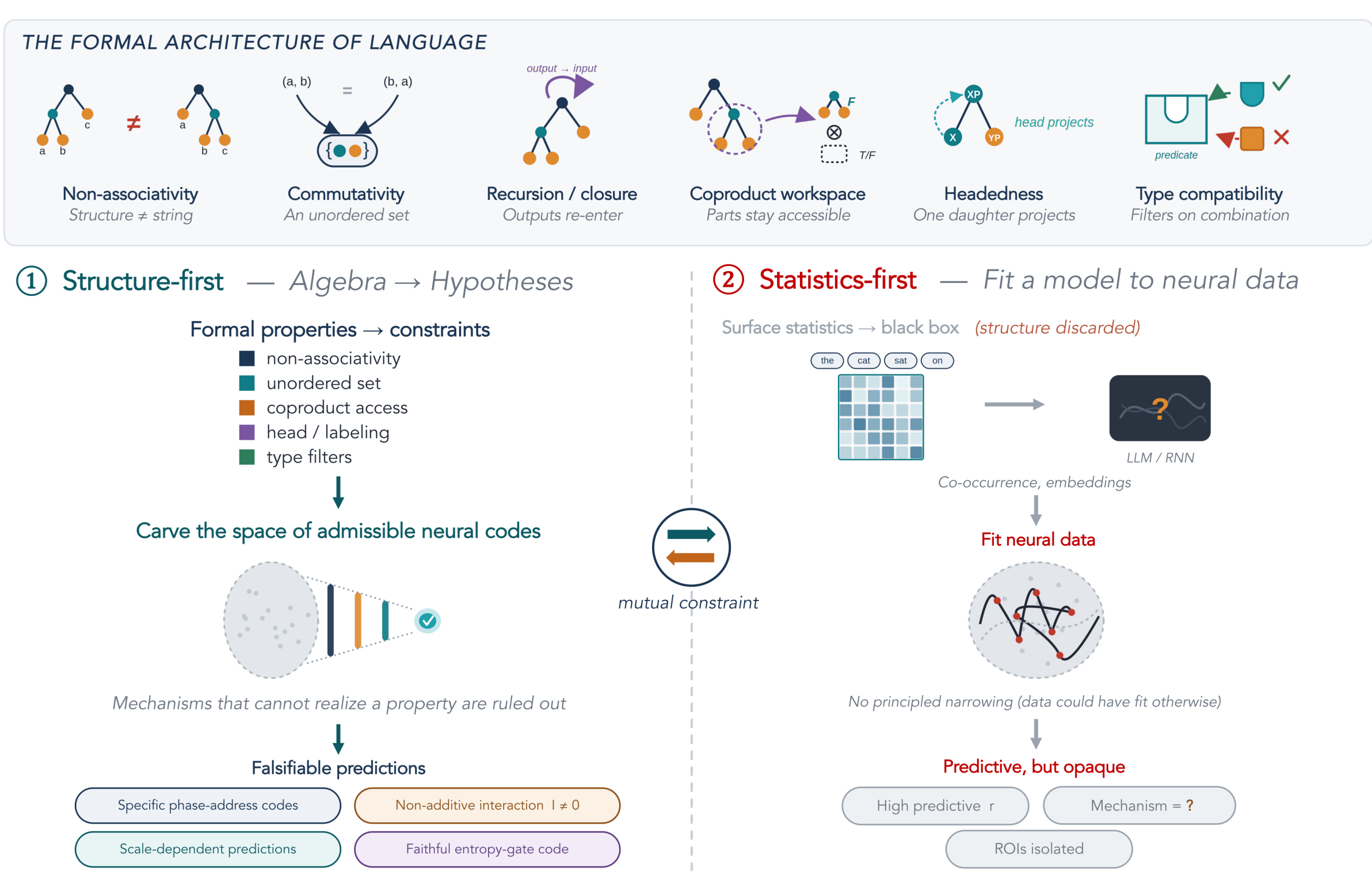

**Figure 1: Two programs for the neurobiology of language.** Top: six algebraic properties of structure-building: non-associativity (irreducible bracketing), commutativity (combination yields an unordered set), recursion/closure

(Merge applies to its own output), the coproduct workspace (any substructure remains addressable), headedness (the head projects its label), and type compatibility (only matching types combine). Structure-first approaches (here, the NAP) use these properties to constrain admissible neural mechanisms and derive predictions. Statistics-first approaches fit surface statistics or distributional embeddings to neural activity, often yielding prediction without mechanism.

We can state what can be termed the core Neural Admissibility Criterion (NAC) that operationalizes the NAP in general terms: Let a theory $F$ specify a space of computational objects $S$, operations $M_i$, and invariants $I_j$. A candidate neural implementation consists of an encoding $\mathcal{N}: S \to X$ into neural state space $X$ and physiological operations $G_i$. It is neurally admissible only if the relevant diagrams commute up to an explicitly stated equivalence and the invariants $I_j$ remain recoverable under $\mathcal{N}$. Neural evidence then adjudicates among admissible implementations; mere correlation with a variable defined by $F$ is not sufficient to establish implementation, while the absence of a simple correlation need not exclude it. Below, we instantiate the NAC by deriving neural consequences of linguistic invariants and using them to distinguish candidate physiological mechanisms.

Our central commitment is methodological: we do not assume that compositional structure is coded at one neurophysiological scale. Neural signatures of higher-order linguistic structure have been recovered (to varying degrees of psycholinguistic resolution) at essentially every scale at which we can currently record. Distributed frontotemporal dynamics track sentence-level structure across cortical areas (Woolnough et al., 2023); local broadband gamma indexes phrasal and morphosyntactic composition (Murphy, Woolnough, et al., 2022; Murphy, Rollo, et al., 2024; Nelson et al., 2017); low-frequency activity entrains to phrase- and constituent-level structure beyond lexical and acoustic confounds (Coopmans et al., 2022; Lu et al., 2023); and population- and single-neuron activity is sensitive to lexico-semantic features (Cai et al., 2026; Jamali et al., 2024; Lakretz et al., 2026; Yan et al., 2026). None has yet been shown to be the primary causal driver rather than a correlate or consequence. Hence, our primary aim is to let those algebraic admissibility conditions, rather than any prior bias toward a favored neural signature, guide the search space.

One reason the NAP has become possible to pursue is that recent work has made both sides of the linking problem unusually explicit. On the formal side, we build on the reformulation of syntax as a free commutative non-associative magma (Marcolli, Chomsky, et al., 2025), its extension to Hopf-algebraic workspaces and Merge Markov chains (Marcolli & Skigin, 2025), colored operads and hypermagmas for typed filtering and headedness (Marcolli, Huijbregts, et al., 2025), and the function-space embedding that realizes composition as an entropy-optimized Rényi gate – a nonlinear combination rule that mixes its two inputs at a weighting fixed by minimizing a free-energy trade-off, and which is commutative and non-associative as a result (Marcolli & Berwick, 2026). On the neural side, we draw on inter-areal dynamics linked to symbolic properties of language such as phase-amplitude coupling (Kazanina & Tavano, 2023; Weissbart & Martin, 2024); recently introduced content-addressable memories built on bipartite expander graphs (Chaudhuri & Fiete, 2019) and Hopfield networks (Manin & Marcolli, 2024; Yampolskaya & Mehta, 2026), and their higher-order simplicial extensions (Burns & Fukai, 2023); combinatorial threshold-linear networks (CTLNs), whose attractors and sequential limit cycles follow from parameter-independent graph rules (Curto et al., 2019, 2024; Parmelee, Alvarez, et al., 2022); the spatial-computing account of low-frequency dynamics as a control field that sculpts where high-frequency content can occur (Chen et al., 2026; Singer & Effenberger, 2025); the separability of phase- and frequency-tuned neuronal populations (Jourahmad et al., 2026); and population-geometry measures such as effective dimensionality (Gao et al., 2017; Jazayeri & Ostojic, 2021) and variance quenching (Churchland et al., 2010).

We utilize these theoretical and empirical advances in principled ways in order to demonstrate proof-of-concept explanatory connections via a language-relevant phase-address code, an order-free hash-key, and a sealing operator that jointly enforce the magma's non-associativity and commutativity in neural terms (§5); a pullback of phrase-structure and semantic role filters whose signature is a specific non-additive interaction (§4.5); a route from the proposed binding gate to a measurable population quantity, under which the effective dimensionality of the composing population should distinguish alternative bracketings, subject to one stated assumption about how composition weights are expressed in population activity (§9.2); a division of labor in which an expander-Hopfield store, a threshold-linear sequencer, and a composition gate realize the memory, the derivation, and the binding step of Merge (§6.7); and, from applying the admissibility conditions to the binding step itself, a neural binding operation we term 'Meld' (§4.9.1), which is built from operations that cortex is known to perform, satisfies every condition the program produces, and recovers hierarchical structure in simulation at every depth and temperature tested while remaining blind to the order of its parts. Crucially, Meld adds no primitive to the theory of grammar: Merge is unchanged, and Meld is a claim about its physical realization. In each of these cases, a formal property of language becomes a concrete claim about what a neural mechanism must do to implement it.

## 2. Hierarchical Constituency Structure

Core phenomena of linguistic structure and meaning depend on hierarchical objects (Adger, 2003; Chomsky, 2013; Collins & Kayne, 2023; Collins & Stabler, 2016; Hornstein, 2024; Senturia & Frank, 2023). Binding, scope, ellipsis, displacement, selection, reconstruction, islands, and compositional interpretation depend on constituents, heads (e.g., 'run' in [$_{VP}$[$_{V}$ *run*][$_{NP}$ *home*]]), and local operations that create nonlocal dependencies. A neural model that only tracks linear distance, dependency edges, or distributional predictability can be empirically useful while missing the computational object that makes the sentence interpretable. For example, a dependency parser can attach useful relational features to words in naturalistic stimuli (Berwick & Stabler, 2019; Brennan et al., 2016; Lopopolo et al., 2021; Stanojević et al., 2023). Those features can be used to *predict* activity in language cortex, but a dependency representation is not, by itself, an account of the constituent-building operation that generated the expression (Figure 2).

The point here is not that dependency relations are wrong but that a dependency graph is a description of an utterance rather than a grammar that licenses one, and the difference matters as soon as the missing components are supplied. Müller shows that the rule systems dependency grammars need in order to license their own graphs closely resemble phrase-structure rules, and that once semantics is added the extra representational machinery required to state the meaning of a head together with its dependents reintroduces the nodes it was meant to eliminate (Müller, 2026). He also shows that headless constructions and discontinuity require either empty heads or constituency-like adjacency constraints. Dependency structure is therefore not a lighter alternative to constituency so much as a *projection* of it, with the licensing conditions left implicit. Vigneaux and colleagues sharpen this distinction from within the probing literature itself (Vigneaux et al., 2026): structural-probe accuracy across dependency relations is strongly predicted by linear dependency distance and the similarity-aware entropy of lexical heads, with greater lexical diversity impairing recovery. Thus, decodability of a dependency relation from an LLM state need not index an abstract syntactic operation; it can depend substantially on surface distance and lexical geometry. The NAP therefore treats recoverable dependency structure as evidence about a representation, but not as an account of constituent-building.

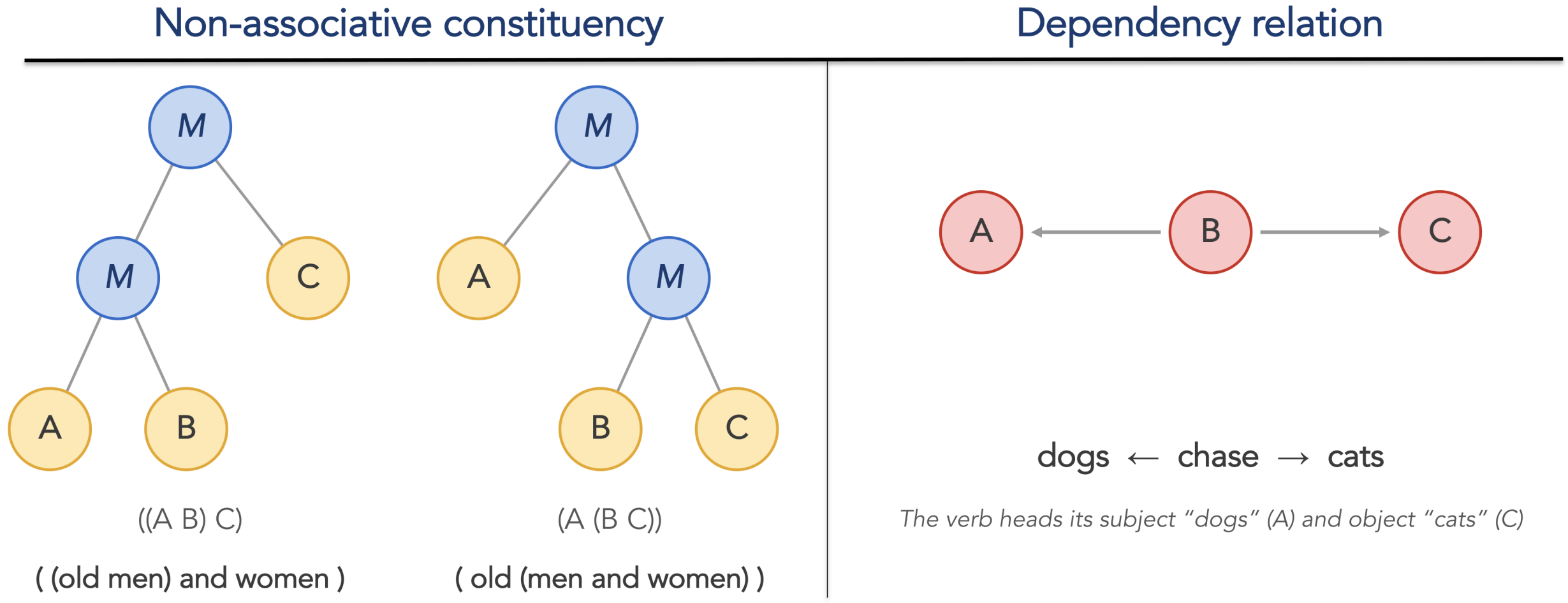

**Figure 2**: **Dependency relations and constituency structure answer different questions.** The same terminals can support different non-associative groupings. A neural mechanism for syntactic composition must preserve at least some grouping distinctions rather than only track terminal relations.

Riveland and colleagues similarly distinguish compositional behavior from compositional implementation, placing candidate mechanisms on a continuum of module expressivity and "syntactic complexity" (Riveland et al., 2026). This provides a useful taxonomy, but its deliberately broad notion of syntax does not determine which recombination rules are admissible for human language: additive or pattern-matching solutions can occupy the continuum while still failing to preserve non-associative constituency. The NAP asks a stricter question – not how compositional a system appears, but whether its dynamics preserve the specific algebraic invariants that linguistic composition requires.

In current models of linguistic knowledge (of the kind that are used to derive features like node closure/opening and dependencies), the key operation, often termed 'Merge', is a binary structure-building operation that constructs syntactic objects recursively (Chomsky, 1995). This process is architecturally separated from 'externalization' (mapping linguistic information to sensorimotor interfaces) and interpretive 'filtering' (conditions set by conceptual systems on how to semantically assess the meaning of phrases, such as thematic roles of nouns, and how to assign a phrase its head).

This target also prevents a common error: treating every sentence-level effect as either semantic or grammatical. During real-time processing, sentences can differ from word lists because of coherence, prosody, task demands, discourse, attention, world knowledge, predictability, or memory (Murphy & Woolnough, 2024). The relevant question becomes not whether compositional syntax-semantics lives in a broad, distributed sentence network – it surely does (Murphy et al., 2023; Woolnough et al., 2023) – but whether particular neural mechanisms implement distinct or overlapping linguistic computations. Concurrently, it would also be a mistake for neuroscientists to search for a local neural signature of every phenomenon in a linguist's inventory, many of which will ultimately reflect artificial psychological constructs of a complex and often arbitrary theoretical space, as argued elsewhere (Boeckx, 2014).

Meanwhile, encoding models are useful tools for localization, timing, and hypothesis generation (Caucheteux et al., 2023; Hale et al., 2022; Weissbart & Martin, 2024). They can test whether an electrode is better predicted by surprisal, node closure, dependency distance, phoneme rate, or semantic coherence. But an encoding coefficient is not an operation: a coefficient for node closure does not show how a neural system closes a node, stores the result, or makes it available to interpretation by other cognitive systems.

As we will explore, human language is governed by formal constraints that fundamentally differ from other cognitive systems. Syntax instantiates a free non-associative algebra over the lexicon. In contrast, vector addition, convolution, and reinforcement-learning updates are typically associative and structure-flattening. We will argue that what is distinctive about certain recent developments in mathematical linguistics is that they allow us to formalize linguistic representations and workspaces in terms that can be mapped onto neural dynamics.

## 3. Design Features of Language

We now summarize the algebraic properties of language that will do most of the theoretical work. Much of what we review here will require invoking structures from category theory and Hopf algebra, as part of a recent move within linguistics (Marcolli, Chomsky, et al., 2025) to strengthen prior theories that have historically relied on naïve set theory and basic properties of recursive function theory (Chomsky, 1995, 2013, 2014).

### *3.1. Syntactic objects as a free commutative non-associative magma*

Let $SO_0$ be the finite inventory of lexical items and syntactic features relevant to the derivation of a given phrasal structure (Marcolli, Chomsky, et al., 2025). The set SO of syntactic objects can be idealized as the free commutative non-associative magma generated by $SO_0$ under a binary operation, Merge ($\mathcal{M}$). Informally, $\mathcal{M}$ takes two syntactic objects and returns a new syntactic object. Commutativity means that $\mathcal{M}$(A,B) and $\mathcal{M}$(B,A) define the same unordered set-like object at the level of core structure building (e.g., the underlying meaning of a phrase will be the same for speakers of languages with different surface word orders). Commutativity matters because the same abstract relation between two elements should survive differences in word order: a Spanish–English bilingual represents the same underlying combination whether they hear 'red boat' or 'barco rojo'. Non-associativity means that $\mathcal{M}(\mathcal{M}$(A,B),C) and $\mathcal{M}$(A,$\mathcal{M}$(B,C)) remain distinct. The operation is closed: applying it to syntactic objects yields another syntactic object. In compact form:

$$\text{SO}=\text{Magma}_{c,na}(\text{SO}_0,\mathcal{M}),\ \text{with}\ \mathcal{M}(A,B)=\mathcal{M}(B,A),\ \text{but}\ \mathcal{M}(\mathcal{M}(A,B),C)\neq\mathcal{M}\big(A,\mathcal{M}(B,C)\big).$$

One linking hypothesis here offers a *dissociation*: the order code belongs to sensorimotor, prosodic, and language-specific linearization processes, whereas the grouping code should remain recoverable after those processes are modeled explicitly, and should be cued in syntax-sensitive language sites (Desbordes et al., 2023; Dighiero-Brecht et al., 2026; Lakretz et al., 2026; Zacharopoulos et al., 2026).

This rules out two extremes. A bag-of-words vector is commutative but too associative: once A, B, and C collapse into one sum, the system cannot recover whether A first combined with B or B with C. A sequential recurrent code may preserve order without preserving unordered set formation.

The status of commutativity should be clarified at this point. As a constraint on real-time processing, it is patently weak. A parser unavoidably receives its input in linear order, and nothing prevents that order from being represented somewhere in the neural state, so an order signal found in the complete recorded population does not by itself disqualify a candidate mechanism. But as a constraint on the composed object its status remains strong (and it is in this second role that it carries an important discriminating result below; §4.9.1, §7.2). The requirement should not be that order is absent from the system entirely, but that it be functionally factored out of the representation available to recursive composition: the composition law must be invariant to whatever order information the state carries, so that the composite it produces does not depend on it. A state may contain an order tag in a dimension the law ignores, and an external decoder may read that tag without the computation using it. The strongest operational test is therefore chance decoding of daughter order within a predefined constituency-preserving subspace (§6.3, §7.2), while bracketing remains decodable there. A law that keeps order in the composed object, as an ordered role-filler law does, represents more than the competence object (i.e., knowledge of syntax-semantics) contains, and is inadmissible as a realization of Merge unless order is discarded *before the composite is consumed*. We accordingly treat commutativity not as a co-equal real-time partner of non-associativity, closure, and the coproduct, but as a condition on the composed state, which is where its empirical relevance lies.

### *3.2. Workspaces, coproducts, and Hopf algebra*

Natural language does not simply produce a finished tree-structure. It operates over workspaces: collections of syntactic objects (e.g., lexico-semantic features, morphemes, partial or complete phrases) that can be combined, extracted, copied, moved, or reassembled. In the algebraic formulation adopted here, workspaces are *forests of binary rooted trees* (Marcolli & Skigin, 2025). The vector space spanned by these workspaces carries a commutative graded connected Hopf algebra structure. The product of two workspaces is their disjoint union. The coproduct extracts accessible substructures and the remaining quotient structure:

$$\Delta(T)=\sum_{c} F_c \otimes T/F_c,$$

where the sum ranges over the admissible cuts $C$ of $T$ – sets of edges no two of which lie on the same root-to-leaf path – so that $F_C$ is a genuine forest of accessible subterms and $T/F_C$ is the quotient obtained after extraction and copy cancellation (i.e., copy of a syntactic object). Merge actions on workspaces can then be represented by operators of the form:

$$\mathcal{M}_{S,S'}=\mu\circ(B\otimes\mathrm{id})\circ\delta_{S,S'}\circ\Delta,$$

where $\Delta$ extracts candidate terms, $\delta$ selects the target pair, B grafts the selected objects, and $\mu$ reassembles the workspace. These operators are the elementary moves of a dynamical system rather than isolated rewrite rules. Summing them over all admissible selections yields a single generator, $K = \sum_{S,S'} M_{S,S'}$, whose iteration defines the derivational dynamics; below, we assemble $K$ explicitly in §4.7 and analyze the Markov chain it induces in §4.8. Note that the schematic form above is most transparent for External Merge, where $\delta_{S,S'}$ selects two independent accessible terms and $B$ grafts them. Internal Merge requires slightly more care, as Marcolli and colleagues have stressed: there, one argument to the gate is a subterm extracted by $\Delta$ and the other is the quotient in which that subterm has been cancelled, so the selection is not of an independent pair drawn from the extracted forest, and part of the work is carried by the copy-cancellation convention in $T/F_C$ (Marcolli & Berwick, 2026). We will adopt the extended formula throughout.

These observations form a major part of the reason why neural language models must support substructure access, not only global 'sentence embeddings'. A system that builds hierarchical objects must retain structured access to their parts. If a neural mechanism only accumulates *a global sentence representation* (e.g., a typical 'ramping' activation signature), it does not satisfy the workspace requirement (Figure 3A). A Hopf-algebraic perspective therefore pushes neural theories toward mechanisms of composition-decomposition duality: build an object, keep its internal components addressable, and make them available for flexible re-composition.

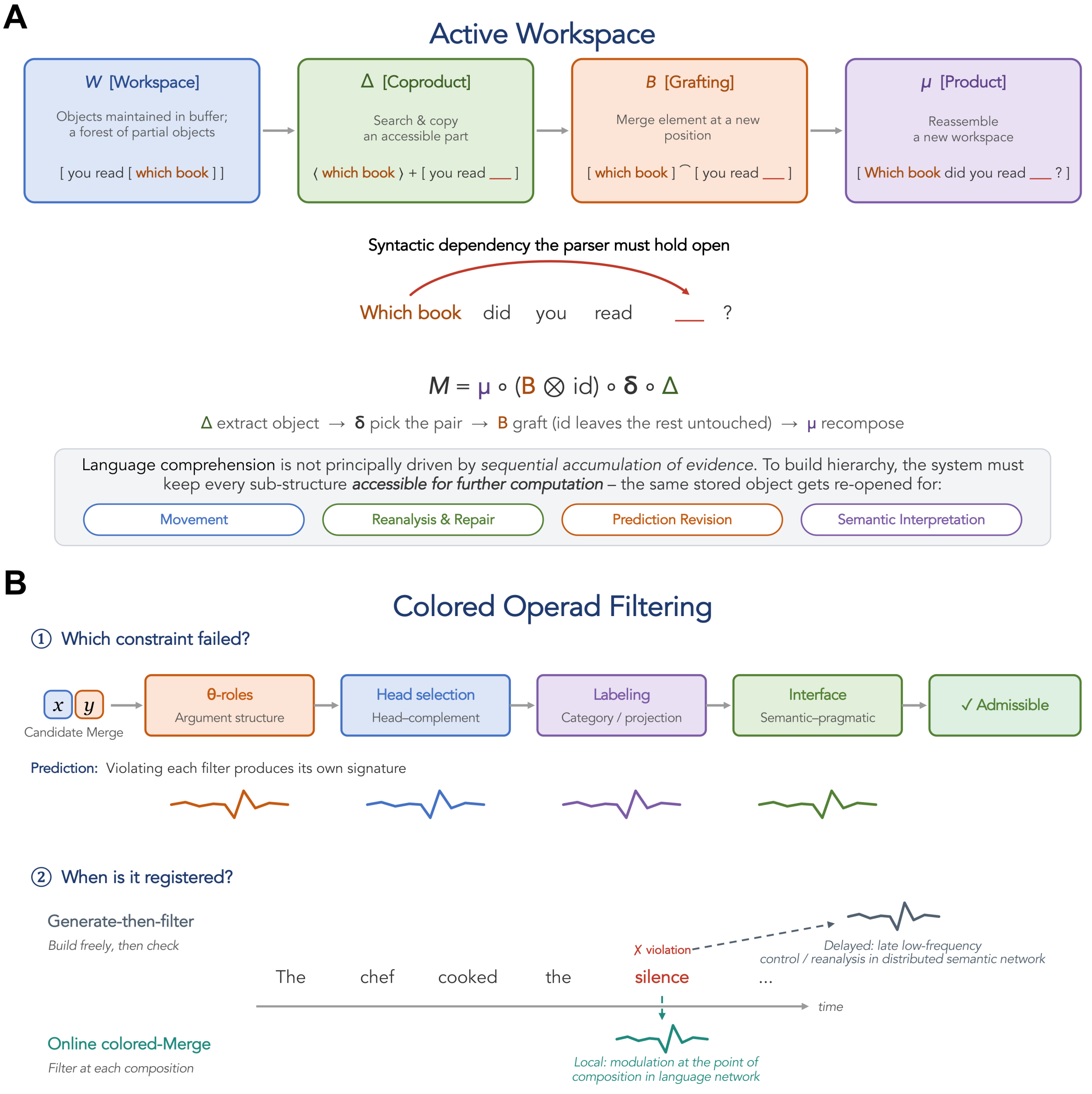

**Figure 3: Workspace and filtering dynamics. (A)** Merge as an operation on an active workspace. The Hopf-algebraic formulation recasts Merge not as simple pairwise composition but as a multi-step cycle over a workspace of partial syntactic objects, illustrated here with the formation of a simple *wh*-question. From an already-built object (*W*: [you read [which book]]), the coproduct (Δ) extracts an accessible subterm, copying out *which book* and leaving a gapped remainder; grafting (*B*) re-merges that subterm at a new position; and the product (μ) reassembles the result into a new workspace (*Which book did you read* __?). The merge operator $\mathcal{M} = \mu \circ (B \otimes \mathrm{id}) \circ \delta \circ \Delta$ thus reads right-to-left as extract, select, graft, recompose, with id leaving the unselected structure untouched. **(B)** Colored-operad filtering. Free Merge over-generates; colored operads admit only type-compatible combinations. Some immediate neural questions concern which filter fails and when filtering applies: generate-then-filter predicts delayed reanalysis, whereas online filtering predicts local modulation at composition.

*3.3. Markov chains on workspaces and entropy-optimized dynamics*

Recent work on Merge dynamics formulates the action of Merge on workspaces as a Hopf algebra Markov chain (Marcolli, Chomsky, et al., 2025; Marcolli & Skigin, 2025). This offers a principled way to treat derivations as state transitions over workspaces rather than as arbitrary symbolic rewrite sequences. 'Internal Merge' (merging an element already contained in a workspace with the larger object containing it, e.g., $\mathrm{Merge}_{\mathrm{Int}}(what, [John\ read\ what]) = [what_i[John\ read\ t_i]]$) is, on its own, an ergodic dynamical system: on each connected component of workspaces of fixed lexical content, it is strongly connected and aperiodic, with the uniform distribution as its unique stationary distribution (Marcolli & Skigin, 2025); 'External Merge' (merging two independent workspace elements, e.g., $\mathrm{Merge}_{\mathrm{Ext}}(read, books) = [read\ books]$) contributes structure-building transitions that move the dynamics toward more connected workspaces.

This view suggests that real-time parsing should not be modeled as a flat accumulation of feature load. It should instead be modeled as a sequence of transitions through a structured state space. Some transitions open a workspace by adding available objects, or closing local structures, or retrieving an accessible term, or filtering an object for interpretability. This claim is naturally compatible with probabilistic parsing, but it differs from treating syntax as only non-Markovian sequence statistics. The relevant Markov property is not over the word string but over a *structured workspace state*; the fact that Merge is Markovian is already discussed by Chomsky (Chomsky, 2021). A neural signal may therefore look history-sensitive at the string level while being Markovian over a richer internal state.

*3.4. Colored operads, theta roles, and labeling*

The free generation of syntactic objects over-generates and results in a combinatorial explosion. Natural language does not require the interpretation of every possible binary tree over a lexical array. Objects must pass *filters* involving head-complement structure, theta-role assignment (Agent, Location, etc.), labeling (Hornstein, 2024), and interface viability. Colored operads provide a way to formalize such filters (Marcolli, Huijbregts, et al., 2025). In an operad, operations compose inputs into outputs. In a colored operad, the inputs and outputs carry types or colors, and only compatible operations are admitted. A bud generating system can propagate these colors through a derivation.

The point for our purposes is that the brain must distinguish successful from unsuccessful combinations in a way that is more specific than generic statistical anomaly detection. Admissible and inadmissible structures should diverge according to the violated filter. Colored Merge also clarifies an algorithmic issue: One can first freely generate structures and then filter them, or one can combine structure building and filtering online (Figure 3B). Theta-role violations should be detectable via the means through which the cortical language network exports information to general semantic systems (Fairhall & Caramazza, 2013), whereas phrase structure/labeling violations should be more locally detectable via interactions between core frontotemporal language sites (Murphy, Woolnough, et al., 2022; Murphy, Rollo, et al., 2024).

### *3.5. Hypermagmas and head functions*

Headedness introduces an additional complication (Chomsky, 2013; Hornstein, 2024; Murphy, 2015a; Pan et al., 2024). A bare magma can build unordered binary objects, but natural language must also determine which component labels and controls further interpretation. Recent work extends the magma structure to a hypermagma when head functions are incorporated (Marcolli, Huijbregts, et al., 2025), because head-related relations are not ordinary binary combinations. Whereas c-command (i.e., the relation in which a node's first branching parent also dominates another node) can be characterized in relation to the magma structure, m-command (i.e., the broader relation in which a node commands material within the domain of its maximal projection) and extended projections require richer structure.

The implication is that composition and labeling should not be treated as the same event. Existing research already points to some separation between semantic composition and headedness (Yang et al., 2026; Zacharopoulos et al., 2026; Zhao et al., 2025). A pair of objects may be bound before the system determines which object supplies the label, and which phrase structure or theta-role constraints are satisfied. Hypermagmas model cases where the grammar has built a structure but has not yet settled which daughter controls it (as in small-clauses like 'John in the garden', 'John angry'). If this distinction is neurally realizable, then head selection should show a competition phase, followed by stabilization. In rapid comprehension, these processes may naturally overlap in time, but they need not be identical in scale: Local broadband $\gamma$ activity may index the availability of a lexical or combinatorial object (Murphy, 2024, 2025), while low-frequency phase organization and inter-areal coupling may index selection, projection, and workspace stabilization.

### *3.6. The syntax-semantics interface as structured transformation*

The syntax-semantics interface is often described loosely as "semantic integration", "binding", or "unification" (Hagoort, 2005, 2017). Yet, compositional meaning is not just the incremental addition of lexical meanings. It depends on the phrasal head, argument positions, available scope relations, amongst other things. Algebraic models of the syntax-semantics interface therefore treat semantic interpretation as a structured transformation from syntactic objects into semantic spaces, not as unconstrained vector addition (Marcolli, Chomsky, et al., 2025; Monteza & Hermansyah, 2025). A semantic embedding that distinguishes *dog* from *cat* but collapses every difference between active and passive, raising and control, quantifier-scope alternatives, or different bracketing structures is not a neural code for higher-order syntax-semantics. Conversely, a neural population encoding lexical features may contribute to compositional meaning only when field-level or inter-areal dynamics assign it a structural role (Barrett & Miller, 2026; Chen et al., 2026; Martin, 2020).

The coproduct matters for interpretation as well as for movement. Interpreting a completed object requires decomposing it into the parts over which the interpretation function is actually defined – for example, the argument positions a head selects, or the domain in which a variable is bound – and $\Delta$ is precisely the operation that supplies those decompositions together with the residual quotient in which the extracted part is to be interpreted. Compositional semantics in this setting is therefore not read off a finished tree by traversal, but is computed over the same admissible-cut decompositions that $\Delta$ makes available to Merge, which is what makes the interface a *structured transformation* rather than an aggregation. This refines the neural expectation stated in §3.2: the substructure-access signature should not be restricted to filler-gap and reanalysis contexts, but should also arise wherever interpretation itself demands access to a sub-object of an already-built structure (e.g., quantifier scope, binding, coercion) in the absence of any displacement.

Table 1 summarizes some physical implications of the formal approach we have so far developed. The next section will turn to more precisely characterizing these algebraic properties.

| Algebraic property | Formal role | Neural admissibility condition | Options for falsification |
| --- | --- | --- | --- |
| Non-associativity | $\mathcal{M}(\mathcal{M}(A,B),C)$ differs from $\mathcal{M}(A,\mathcal{M}(B,C))$ | The code must preserve grouping history or depth at later composition points | A neural signature collapses all same-terminal bracketings into an undifferentiated sum |
| Closure | Outputs of Merge are syntactic objects eligible for further Merge | Composition should create a reusable representational state, not only a transient response | Phrase-level activity cannot be reactivated, decoded, or used in subsequent composition |
| Workspace coproduct | Substructures can be extracted from a workspace | Neural dynamics must support addressable subparts for movement, reanalysis, and interpretation | Only global sentence embeddings are recoverable; no structured subconstituent access is detectable |
| Colored operads | Filters enforce theta-role, head, and label compatibility | Violation types should show filter-specific signatures, and combined theta-role + head violations should interact non-additively | All violations reduce to generic difficulty, surprise, or an additive sum of anomaly effects |
| Hypermagma/head functions | Head selection and extended projection enrich bare composition | Binding and head selection should be separable, with competition-to-resolution dynamics especially for exocentric cases where Dom(h) fails to close under Merge | No selective competition/resolution signature for exocentric head ambiguity beyond generic difficulty |
| Commutativity | $\mathcal{M}(A,B)=\mathcal{M}(B,A)$ at the unordered structure-building level | Invariance of the completed object to daughter order | The composition code is fully predicted by surface order, with no separable grouping projection |
| Hopf algebra Markov dynamics | Derivations are transitions over structured workspaces | Neural state transitions should reflect build, filter, and closure operations | State dynamics track only word rate or lexical surprisal |
| Function-space embedding | Discrete objects can be faithfully represented in a wavelet-like function space | Neural waves are plausible only if they preserve recoverable structure, not merely correlate with rhythm | Oscillatory signatures track acoustic/prosodic envelopes but not algebraic contrasts |

**Table 1: Algebraic properties and their neural admissibility consequences.**

## 4. Algebraic Admissibility for Neural Enforcement

The preceding section introduced relevant algebraic objects mostly verbally. We now state them more explicitly.

Beginning at the most general level of multi-element structures, let $SO_0$ again denote a finite set of lexical atoms and syntactic features. Let SO denote the closure of $SO_0$ under a binary operation $\mathcal{M}$. Let $X$ be a neural state space and $\mathcal{N}: SO \to X$ an encoding of syntactic objects into neural states. The neurobiological question is whether there exists a physiological composition operation $G: X \times X \to X$ such that formal composition and neural composition agree, at least approximately:

$$\mathcal{N} \circ \mathcal{M} \approx G \circ (\mathcal{N} \times \mathcal{N}).$$

Here, G is not one particular frequency band, cortical region, or decoding model, but a placeholder for the physiological operation or family of operations that would make neural composition structurally faithful. If no such G is found, the neural signal may still *correlate* with linguistic structure, but it has not yet been shown to implement it.

### *4.1. Free commutative non-associative magmas and the irreducibility of bracketing*

The most basic object we can deal with in this space is the free commutative non-associative magma generated by $\mathrm{SO}_0$:

$$\mathrm{SO}{=}\mathrm{Magma}_{c,na}(\mathrm{SO}_0,\mathcal{M}).$$

This means that SO is the smallest set containing $\mathrm{SO}_0$ and closed under a single binary operation $\mathcal{M}$:SO×SO→SO, with commutativity but without associativity. The canonical model of this structure is the set $T_{\mathrm{SO}_0}$ of full non-planar binary rooted trees whose leaves are labeled by elements of $\mathrm{SO}_0$. Non-planarity corresponds to commutativity; non-associativity corresponds to the preservation of hierarchical grouping. Thus, the distinction between $\mathcal{M}(\mathcal{M}(A,B),C)$ and $\mathcal{M}(A,\mathcal{M}(B,C))$ is the formal residue and consequence of the reality of constituency structure.

A neural implementation must therefore satisfy a minimal bracketing-separation condition. Let $\mathcal{N}$:SO→$X_N$ be a neural encoding into a state space $X_N$. For a fixed lexical multiset $\{\alpha,\beta,\gamma\}$, a syntax-sensitive encoding must not collapse all bracketings

$$\mathcal{N}\big(\mathcal{M}(\mathcal{M}(\alpha,\beta),\gamma)\big)\neq\mathcal{N}\Big(\mathcal{M}\big(\alpha,\mathcal{M}(\beta,\gamma)\big)\Big),$$

unless the grammar or the conceptual interfaces have independently identified those objects. Hence, when lexical content and transition statistics are controlled, a putative syntactic signal must retain information about binary grouping – indeed, recent experimental efforts have moved in this direction (Chen et al., 2025; Coopmans et al., 2022; Ding et al., 2016; Lu et al., 2023; Lyu et al., 2025; Zou et al., 2026), separating more clearly syntactic prediction from integration (Iaia & Tavano, 2026; Murphy, Woolnough, et al., 2022).

### *4.2. Abstract syntax as an order-forgetting quotient*

Natural language syntax-semantics is not modality-specific, and so externalization can be regarded as a passage from abstract non-planar trees to planar or sequential outputs. If $T^{pl}_{\mathrm{SO}_0}$ denotes planar binary rooted trees, then there is a ‘forgetful’ projection

$$\Pi{:}T^{pl}_{\mathrm{SO}_0}\rightarrow T_{\mathrm{SO}_0}$$

that removes planar embedding. Trivially, many ordered, externalized versions of a phrase structure can correspond to the same abstract syntactic object. A language-specific externalization map can be treated as a section $\sigma_L{:}T_{\mathrm{SO}_0}\rightarrow T^{pl}_{\mathrm{SO}_0}$ with $\Pi\circ\sigma_L$=id. Crucially, $\sigma_L$ is not a magma morphism in general: narrow syntax is not recovered by assuming that the order of words is the generating algebra. We cannot reconstruct compositional meaning simply by looking at the word string as if the string itself were the generative algebra. The neural code for the syntactic object may be *inferred* and *cued* through the channel, but it should not be identified with that channel. We note that treating linear order as entirely a property of externalization is itself a theoretical commitment: antisymmetry-based approaches and pair-Merge analyses locate some asymmetry within narrow syntax (Kayne, 1994; Safir, 2024), and a reader who adopts those frameworks would therefore be advised to weaken the strict order-as-quotient assumption made here.

### *4.3. Syntax as composable families of operations*

A magma specifies one binary operation. An operad specifies a family of composable operations with variable arity. The conceptual shift is worth stating plainly for readers who have not previously encountered the formalism. In the magma setting, a tree is an *object*: the output of repeated application of $M$. In the operad setting, the same tree is promoted to the role of an *operation*, with its leaves as inputs and its root as output, so that a tree with $n$ leaves is an $n$-ary operation awaiting $n$ arguments. Composition is then substitution: grafting one tree into a leaf of another is the composite of the two operations, and the operad axioms are the demand that this substitution be associative and unital. What was a finished structure becomes a rule for building structures, and the operad is the collection of all such rules closed under substitution[1].

In the category of sets, an operad O is a collection $\{O(n)\}_{n\geq 1}$, where $O(n)$ contains n-input, one-output operations, together with composition maps (Giraudo, 2018; Marcolli, Huijbregts, et al., 2025)

$$\gamma{:}O(n)\times O(k_1)\times\cdots\times O(k_n)\to O(k_1+\cdots+k_n)$$

satisfying associativity of substitution:

$$\gamma\big(\gamma(T;T_1,\ldots,T_n);T_{1,1},\ldots,T_{n,k_n}\big)$$
$$=\gamma\left(T;\gamma\big(T_1;T_{1,1},\ldots\big),\ldots,\gamma\big(T_n;\ldots,T_{n,k_n}\big)\right).$$

The Merge operad $\mathcal{M}$ has $\mathcal{M}(n)$ equal to the set of abstract non-planar binary rooted trees with *n* leaves. Its insertion operations ($\circ_i$) graft one tree into a leaf of another:

$$\circ_i{:}\mathcal{M}(n)\times\mathcal{M}(m)\to\mathcal{M}(n+m\text{-}1),\quad (T,T')\mapsto T\circ_i T'.$$

The set of syntactic objects is then an algebra over this operad:

$$\gamma_{\mathrm{SO}}{:}\mathcal{M}(n)\times\mathrm{SO}^n\to\mathrm{SO}.$$

This operadic view separates the abstract operation-type from its implementation. A neural process would count as an algebra over the Merge operad *only* if it provides concrete operations whose compositions respect operadic substitution.

### *4.4. Colored operads and bud-generating systems*

Infamously (Bošković, 2018; Chomsky, 2013; Ginsburg, 2024; Maier et al., 2023), and as mentioned above, free Merge over-generates. Interface viability requires *filters*: theta-role compatibility, head-complement structure, 'phasehood', labeling, and movement restrictions.

[1] I thank Matilde Marcolli for highlighting to me the conceptual shift implied by these formulations.

Colored operads formalize this by allowing operations only when input and output colors match (Marcolli, Huijbregts, et al., 2025). Let $\Omega$ be a finite set of colors. A colored operad consists of sets

$$O(c;c_1,\ldots,c_n),\quad c,c_i\in\Omega$$

with composition only when the output color of an inserted operation matches the required input color of the receiving operation:

$$\gamma:O(c;c_1,\ldots,c_n)\times O\big(c_1;c_{1,1},\ldots,c_{1,k_1}\big)\times\cdots\to O\big(c;c_{1,1},\ldots,c_{n,k_n}\big).$$

A bud operad generated from an ordinary operad O and color set $\Omega$ is

$$B_\Omega(O)(n)=\Omega\times O(n)\times\Omega^n.$$

A bud generating system is a tuple

$$B=(O,\Omega,R,I,T).$$

where $R$ is a finite set of allowed local coloring rules, $I$ is a set of initial colors, and $T$ is a set of terminal colors. The language $L(B)$ consists of operations derivable from initial units using rules in $R$ and ending with terminal input colors:

$$L(B) = \{x = (c,\tau,\vec{c})\in B_\Omega(O)(n) \mid 1_c \Rightarrow_B x,\ c\in I,\ \vec{c}\in T^n\}.$$

Here, $\tau \in O(n)$ is the underlying $n$-ary operation, $\vec{c} = (c_1, \ldots, c_n)$ is its input-color vector, $1_c$ is the colored unit of color $c$, and $\Rightarrow_B$ denotes derivability by a finite sequence of rules in $B$.

The critical point is that L(B) is generally not a submagma of SO. That is, even if A and B are individually well-colored, $\mathcal{M}$(A,B) need not be well-colored. Hence, structure-building and filtering/labeling should not be assumed to have identical physical mechanisms.

*4.5. Combining phrase structure and theta filters by pullback*

The strongest prediction of the colored-filter framework is not that theta-role and phrase-structure violations dissociate. The substantive claim concerns *how* the two filters compose. If phrase-structure coloring and theta-role coloring are combined by a pullback (fiber product) over a shared interface/color space, rather than merely placed side by side, the combined filter is not the direct sum of two independent anomaly detectors. It is an *interaction object*: a derivation is admissible only if both colorings are simultaneously satisfied relative to the same compatibility space.

Before we proceed, we note that phase theory remains a formal theory of cyclic spell-out/locality (Chomsky, 2001; Citko, 2014; Legate, 2003), but several psycholinguistic findings converge with its core motivation. Comprehenders appear to build long-distance dependencies cyclically, with clause-boundary/intermediate-gap effects and online sensitivity to islands (Frazier & Clifton, 1989; Gibson & Warren, 2004; Keine, 2020; Phillips, 2006; Stowe, 1986). The evidence is strongest for CP-like/clausal locality domains and weaker for the specific claim that $v^*$P is a psychologically explicit phase, although related processing evidence from Japanese supports the psychological reality of VP/vP-internal structure (Koizumi & Tamaoka, 2010). For our purposes, given that this is a methodological and theoretical neuroscience discussion, we do not make strong commitments to phrase- versus phase-structure filters, and rely only on the existence of cyclic interface updating.

To see why a pullback is the right construction, and not a naive intersection, note that phrase-structure filters and theta-role filters are not two names for the same coloring. Let $B_\Phi$ be a bud system for phrase structure, with colors in $\Omega_\Phi$, and let $B_\Theta$ be a bud system for theta-role assignment, with colors in $\Omega_\Theta$. Because the two systems classify derivations using different color vocabularies, the expression

$$L(B_\Phi) \cap L(B_\Theta)$$

is not, by itself, well-formed: the two languages are not automatically subsets of a single space of colored syntactic objects. What is required is a common refinement in which one derivational object carries enough information to be read both as a phrase-structure-colored object and as a theta-role-colored object. We therefore take the refined color set to be a relation,

$$\hat{\Omega} \subseteq \Omega_\Theta \times \Omega_\Phi,$$

whose elements are exactly the jointly realizable (theta-color, phrase-color) pairs, with $\pi_\Theta$ and $\pi_\Phi$ the two coordinate projections. Because $\pi = (\pi_\Theta, \pi_\Phi)$ is injective by construction, a $\hat{\Omega}$-colored derivation is determined by its two projections. We write

$$\mathfrak{B}_\Omega(\mathcal{M}) := \bigsqcup_{n \geq 1} B_\Omega(\mathcal{M})(n)$$

for the set of $\Omega$-colored Merge trees over all arities. Thus $\hat{t} \in \mathfrak{B}_{\hat{\Omega}}(\mathcal{M})$, and $(\pi_\Theta)_*$ and $(\pi_\Phi)_*$ act by applying the corresponding color projection at every vertex while leaving the underlying Merge tree unchanged. The projections induce forgetful maps on colored trees,

$$(\pi_\Theta)_* : \mathfrak{B}_{\hat{\Omega}}(\mathcal{M}) \to \mathfrak{B}_{\Omega_\Theta}(\mathcal{M}), \quad (\pi_\Phi)_* : \mathfrak{B}_{\hat{\Omega}}(\mathcal{M}) \to \mathfrak{B}_{\Omega_\Phi}(\mathcal{M}),$$

that retain only the theta-role or only the phrase-structure coloring of an $\hat{\Omega}$-colored derivation $\hat{t}$. Viability is then defined in $\mathfrak{B}_{\hat{\Omega}}(\mathcal{M})$ as the intersection of the two preimages,

$$Viable = (\pi_\Theta)_*^{-1}(L(B_\Theta)) \cap (\pi_\Phi)_*^{-1}(L(B_\Phi)),$$

which is precisely the pullback of the inclusion $L(B_\Theta) \times L(B_\Phi) \hookrightarrow \mathfrak{B}_{\Omega_\Theta}(\mathcal{M}) \times \mathfrak{B}_{\Omega_\Phi}(\mathcal{M})$ along the joint map $((\pi_\Theta)_*, (\pi_\Phi)_*)$:

$$Viable = \mathfrak{B}_{\hat{\Omega}}(\mathcal{M}) \times_{\mathfrak{B}_{\Omega_\Theta}(\mathcal{M}) \times \mathfrak{B}_{\Omega_\Phi}(\mathcal{M})} (L(B_\Theta) \times L(B_\Phi)).$$

The substantive content of the construction lies entirely in whether $\hat{\Omega}$ is a proper subset of $\Omega_\Theta \times \Omega_\Phi$. If $\hat{\Omega} = \Omega_\Theta \times \Omega_\Phi$ – every theta-coloring compatible with every phrase-structure coloring – then every pair of separately admissible projections lifts, the pullback collapses to the Cartesian product $L(B_\Theta) \times L(B_\Phi)$, and the two filters are genuinely independent: a derivation is viable iff each filter passes it separately, and nothing is shared. If instead $\hat{\Omega} \subsetneq \Omega_\Theta \times \Omega_\Phi$, some combinations of theta-color and phrase-color are jointly unrealizable even though each is individually well-formed, and viability becomes an irreducible compatibility condition over one shared object rather than a conjunction of two verdicts.

The empirical claim of this section is therefore not that the filters combine by a pullback rather than by a product – indeed, every pullback over a full relation is a product – but that the compatibility relation $\hat{\Omega}$ is proper. Figure 4 depicts this schema (Figure 4A).

**A** Combined filter violations should interact, not add

$O_{\Theta \times_{\hat{\Omega}} \Phi}$

Combined filter: viable objects over $\hat{\Omega}$

$(\pi_\theta)_*$ $(\pi_\Phi)_*$

$O_\Theta$ , $B_\Theta$

Theta-role filter

$O_\Phi$ , $B_\Phi$

Phrase-structure filter

$$Viable = (\pi_\theta)_*^{-1}( L(B_\Theta) ) \cap (\pi_\Phi)_*^{-1}( L(B_\Phi) )$$

$$I_{\text{filter}} = R(\theta + \varphi) - R(\theta) - R(\varphi) + R(\emptyset) \neq 0$$

**B** Hopf-algebraic workspace dynamics

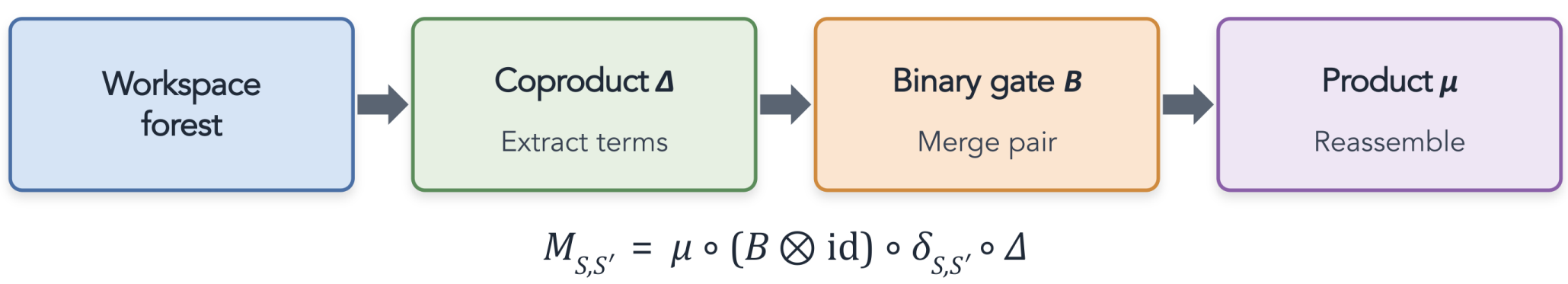

$$M_{S,S'} = \mu \circ (B \otimes \text{id}) \circ \delta_{S,S'} \circ \Delta$$

$$K = \sum_{S,S'} M_{S,S'} = \mu \circ (B \otimes \text{id}) \circ \Pi^{(2)} \circ \Delta$$

*The relevant Markovian object is not a sentence string, but a workspace transition system over forests*

**C** Derivational dynamics as a Markov chain over workspaces

*Branching opens options ↑ H* *Filters close options ↓ H*

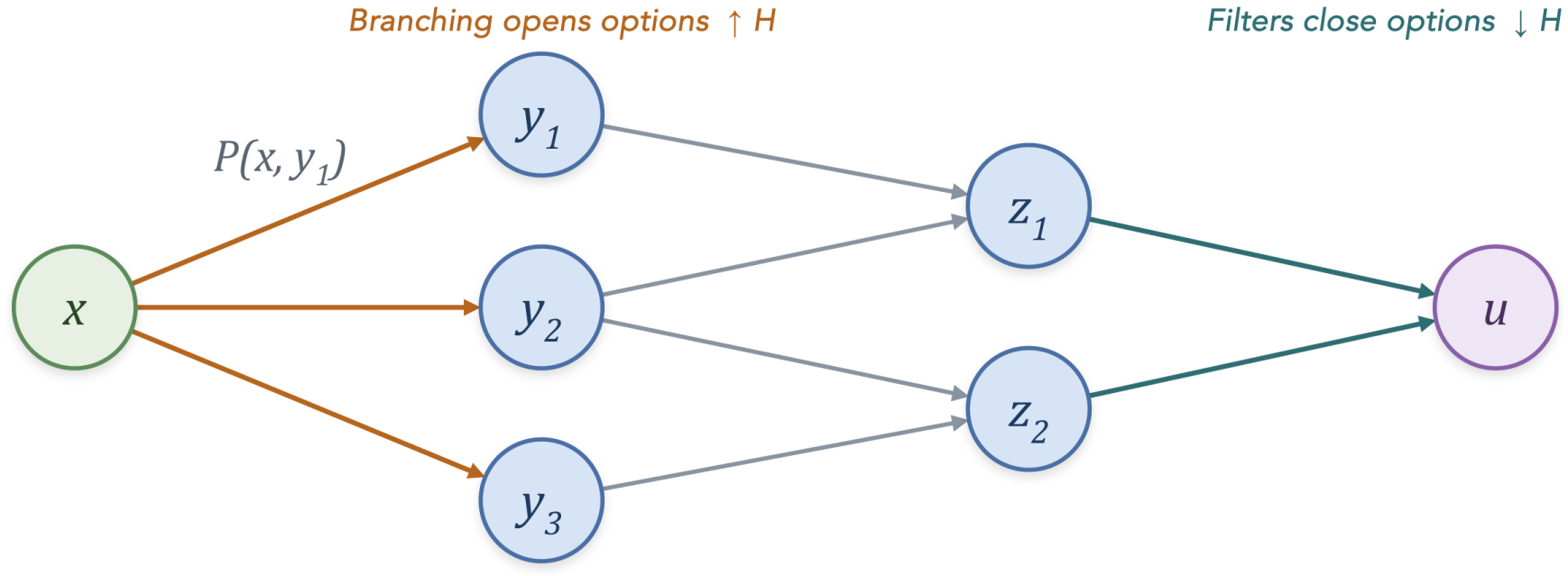

$$P_\beta(x, y) = \exp[ -\beta\, C(x, y) ]\, A(x, y) \;/\; Z_\beta(x) \;, \quad \sum_y P(x, y) = 1$$

$$h(P) = -\sum_x \pi(x) \sum_y P(x, y) \log P(x, y)$$

*Parsing as constrained stochastic flow through workspace states, not traversal of a single best-predicted tree*

**Figure 4. Formal properties of derivational dynamics. (A)** Phrase-structure and theta-role viability combine as a pullback, not as independent detectors. Each filter is a colored bud system: $\mathcal{O}_\Theta$ ($B_\Theta$, colors $\Omega_\Theta$) for theta-role coloring and $\mathcal{O}_\Phi$ ($B_\Phi$, colors $\Omega_\Phi$) for phrase-structure coloring, with admissible-derivation languages $L(B_\Theta)$, $L(B_\Phi)$ over the Merge magma $\mathcal{M}$. Since the filters use different color vocabularies they cannot simply be intersected; they combine over a refined color set $\hat{\Omega} \subseteq \Omega_\Theta \times \Omega_\Phi$ – the relation of jointly realizable color pairs – whose projections $\pi_\Theta, \pi_\Phi$ induce the forgetful maps $(\pi_\Theta)_*, (\pi_\Phi)_*$ (descending arrows). *Viable* is the pullback, shown here with its two projections. It has content only when $\hat{\Omega}$ is proper: were $\hat{\Omega}$ is the full product $\Omega_\Theta \times \Omega_\Phi$, every separately

admissible pair would lift and the filters would be independent. The predicted neural signature is the 2×2 interaction term $I_{\text{filter}}$. **(B)** Hopf-algebraic workspace dynamics. A Merge step as a coproduct-gate-product cycle over workspace *forests* (disjoint unions of syntactic objects) spanning a Hopf algebra. The coproduct Δ extracts terms, the binary gate $B$ merges a pair, and the product $\mu$ reassembles; ⊗ is the tensor product, $\mathrm{id}$ the identity, ∘ composition (right to left). $\mathcal{M}_{S,S'} = \mu \circ (B \otimes \mathrm{id}) \circ \delta_{S,S'} \circ \Delta$ is one Merge step on the pair $(S, S')$ (stated here for External Merge), with selector $\delta_{S,S'}$. Summing over pairs gives the total generator $K = \sum_{S,S'} \mathcal{M}_{S,S'}$, where $\Pi^{(2)} = \sum_{S,S'} \delta_{S,S'}$. The dynamics are here Markovian over forests, not sentence strings. **(C)** The External-Merge quotient of the workspace transition system. Nodes represent workspace states (forests of syntactic objects) and directed edges represent single External Merge transitions, with edge weights obtained by row-normalizing the External-Merge part of the degree-$l$ block of $K$. Row normalization here yields the simple random walk on the quotient, not the Doob transform $P(x, y)$ of §4.8, which requires an irreducible $K_l$; the quotient is drawn for its component-count structure, not as the chain on which the entropy rate is defined. Component-preserving Internal Merge transitions are collapsed within each sector, so every transition represented here reduces the number of forest components. Parsing is thus modeled as a constrained stochastic flow through alternative workspace states, rather than as traversal of a single best-predicted tree. The local branching entropy $H(x)$ – the uncertainty over the next External Merge transition from state $x$ – increases when an incoming word opens attachment possibilities and decreases as grammatical and interface filters eliminate them ($\uparrow H$, $\downarrow H$).

This compositional structure has a direct neural consequence: the combined response should be non-additive. Let $R(\theta)$, $R(\phi)$, and $R(\theta + \phi)$ denote appropriately matched neural responses to theta-only, phrase-only, and combined theta-and-phrase violations, and let $R(\varnothing)$ denote the licit (no-violation) baseline. A pullback-style filter predicts a non-zero interaction term,

$$I_{\text{filter}} = R(\theta + \phi) - R(\theta) - R(\phi) + R(\varnothing) \neq 0,$$

which is exactly the interaction contrast of a factorial design crossing the presence/absence of each violation type. A model in which the two filters operate independently predicts $I_{\text{filter}} \approx 0$: the response to a double violation is just the sum of the two single-violation responses, measured against baseline. The colored-filter model instead predicts that simultaneous violations reshape the ‘commitment’ or repair dynamics, because the system cannot satisfy the two colorings independently over the same local object. Critically, because $I_{\text{filter}}$ is a difference of differences, it is insensitive to additive contributions from lexical properties, word-by-word surprisal, or generic anomaly detection, provided these factors are matched across the four cells. A reliable departure of $I_{\text{filter}}$ from zero therefore constitutes a targeted test of compositional (pullback) structure.

A saturating nonlinearity in the neural response is by itself sufficient to make $I_{\text{filter}} \neq 0$, so the bare existence of an interaction is not diagnostic. The sign, however, is. Suppose the two violations contribute additively to a latent variable read out through a monotone transfer function *f*, so that $R(\theta + \varphi) = f(x_\theta + x_\varphi)$. Writing $a = x_\theta$ and $b = x_\varphi$, concavity of $f$ implies decreasing increments, so that the interaction can be written as follows. If *f* is compressive, then decreasing increments give $f(a + b) - f(b) \leq f(a) - f(0)$, so that

$$I_{\text{filter}} = R(\theta + \varphi) - R(\theta) - R(\varphi) + R(\emptyset) = f(a + b) - f(a) - f(b) + f(0) \leq 0.$$

An additive-latent account with a compressive readout can therefore produce only sub-additive interactions. A reliably super-additive interaction, $I_{filter} > 0$, is inconsistent with that entire family, for any concave *f*, without further assumptions. This matters because the attractor realization developed below in §6.7 predicts exactly that sign: incompatible type combinations are configurations with no surviving co-active fixed point, so combined filters eliminate fixed points super-additively rather than additively.

*4.6. Head functions and hypermagmas*

The head-function issue introduced above also requires sharpening. The critical empirical target is not headedness in general, but the failure of $\mathrm{Dom}(h)$ to be a submagma; i.e., the fact that the set of objects admitting a well-defined head function is not closed under Merge, so that merging two headed objects need not return a uniquely headed object. For example, in an endocentric head-complement structure such as [$_{VP}$ [$_{V}$ read] [$_{DP}$ the book]], the projecting head is determined locally: the verb *read* heads the resulting VP, and the head function can therefore be inherited from one of the immediate constituents. This differs from an exocentric {XP,YP} configuration (e.g., small-clause cases such as [$_{SC}$ [$_{DP}$ John] [$_{PP}$ in the garden]]), in which two already-headed objects combine but no antecedent head function by itself determines the head or label of the resulting object.

Here, we write $SO^h$ for the ambient set of headed syntactic objects, each a pair (A, $h_A$) of an underlying tree A ∈ SO and its head function $h_A$. Bare Merge is the operation $\mathcal{M}$: SO × SO → SO of §4.1, defined on underlying trees and blind to head assignment; merging two headed objects therefore requires extending their head functions to the new root. We write $h_+$ and $h_-$ for the two extensions of $h_A$ and $h_B$ to $\mathcal{M}$(A, B) that agree with $h_A$ on A and with $h_B$ on B and differ only at the new root, with

$$h_+(root) = h_A(root_A), \qquad h_-(root) = h_B(root_B).$$

The head-sensitive lift of Merge is then the map

$$\star\colon SO^h \times SO^h \to \mathcal{P}(SO^h),$$
$$(A, h_A) \star (B, h_B) = \{(\mathcal{M}(A,B), h_+), (\mathcal{M}(A,B), h_-)\},$$

whose value is a set of candidate headed objects rather than a single determined one: the structure is built, but its head and label remain unresolved. Note that ⋆ is not an operation on $SO^h$ – its codomain is $\mathcal{P}(SO^h)$, not $SO^h$ – which is precisely the failure of $\mathrm{Dom}(h)$ to be closed under Merge, now stated as a fact about typing rather than as an observation about examples.

Both extensions exist for endocentric and exocentric configurations alike, so ⋆ is set-valued in general. What distinguishes the two is whether the grammar independently collapses the set. In an endocentric head-complement structure, minimal search over the immediate constituents returns a unique projecting head and the set collapses at once. In an exocentric $\{XP, YP\}$ configuration, no antecedent head function determines the outcome, and the set persists until lexical or interface bias breaks the symmetry. The prediction is therefore graded rather than categorical: a competition-to-resolution signature should be present in both cases, with a measurably longer process for exocentric structures.

The relevant empirical observable is therefore a transiently multi-stable label/head state followed by a stable signature. Possible signatures of this may be found in local ensemble geometry or low-frequency ‘commitment’ bursts (Lundqvist et al., 2024), or with directed coupling from posterior temporal cortical regions carrying candidate labels to regions that stabilize the workspace.

Headedness cannot be reduced to the bare magma operation (Marcolli, Huijbregts, et al., 2025). A head function on a tree $T$ is a map,

$$h_T: V_{\text{int}}(T) \to L(T),$$

from internal (non-leaf) vertices to leaves, subject to a coherence condition under restriction: for every subtree $T' \subseteq T$ and every internal vertex $v \in V_{\text{int}}(T')$,

$$h_T(v) = h_{T'}(v).$$

This says that head assignment proceeds by descent and is constrained by restriction. From any internal vertex (the root included) the head is located by passing down one or the other of the two branches below it; and once that descent enters a subtree, the value assigned must agree with the value that the subtree's own head function already assigns. The head function of the full tree, restricted to any subtree, is therefore the head function of that subtree. Coherence is thus a constraint running downward from each node into the material it dominates, not an upward propagation from subtrees to the root: a head selected within a subtree does not thereby become the head of the tree containing it, but it does fix what the containing tree may assign at every vertex inside that subtree. But not every magma-generated tree admits a linguistically appropriate head/label, and (as the exocentric case shows) merging two headed objects may yield more than one admissible headed result. The head-sensitive structure is therefore naturally *hypermagmatic*, and this is what the lift ⋆ introduced above records: where the bare magma supplies a binary operation $\mathcal{M}$: SO × SO → SO, head assignment supplies

$$\star\colon\ SO^h \times SO^h \to \mathcal{P}(SO^h), \qquad (A, h_A) \star (B, h_B) \subseteq SO^h$$

where $\mathcal{P}(\mathrm{SO}^h)$ denotes the power set of $\mathrm{SO}^h$. The set-valued head label $\{h_+, h_-\}$ is the same object read at the level of head assignments: each choice of head in $\{h_+, h_-\}$ selects one element of $(A, h_A) \star (B, h_B)$. Hypermagma structure is thus a natural formal home for labeling ambiguity, exocentricity, and competition among head candidates.

*4.7. Workspaces and coproducts*

Structural ambiguities and syntactic movement require access to substructures. In recent literature, this has motivated 'arrays', 'numerations' and 'workspaces' (Chomsky, 1995; Collins & Stabler, 2016; Krivochen, 2023), most recently formalized as finite forests (Marcolli & Berwick, 2026; Marcolli & Skigin, 2025):

$$F = T_1 \sqcup \cdots \sqcup T_r, \quad T_i \in T_{\mathrm{SO}_0}.$$

Workspaces can be combined by disjoint union of forests: the operation simply places two workspace-forests side by side, preserving their separate components and any repeated copies. The coproduct decomposes a tree into accessible terms and quotients. In the form used by Marcolli and Berwick, we can write schematically

$$\Delta(T) = T \otimes 1 + 1 \otimes T + \sum_{C \in \mathrm{Adm}(T)} \Big(\bigsqcup_{v \in C} T_v\Big) \otimes T / \Big(\bigsqcup_{v \in C} T_v\Big),$$

where $\mathrm{Adm}(T)$ is the set of admissible cuts of $T$: collections of selected vertices, no two of which lie on the same root-to-leaf path. Selecting a vertex *v* is equivalent to selecting the edge that enters *v* from its parent, so this is the condition stated in §3.2 in terms of edges (we use whichever form is locally more convenient). The first two terms are primitive; each summation term contains the forest extracted by the cut and the corresponding quotient tree. The summation terms express admissible extraction of subtrees $T_v$ and the residual quotient after deletion or contraction. This is the algebraic reason a Hopf structure is more informative than a standard tree drawing: the coproduct explicitly formalizes access to substructure.

Given extracted accessible terms $S$ and $S'$, the Merge operator on workspaces is

$$\mathcal{M}_{S,S'} = \sqcup \circ (B \otimes \mathrm{Id}) \circ \delta_{S,S'} \circ \Delta$$

where $\delta_{S,S'}$ searches/selects the extracted pair and B maps $S \sqcup S'$ to $\mathcal{M}(S,S')$. We write

$$\Pi^{(2)} := \sum_{S,S'} \delta_{S,S'}$$

for the projector that retains the coproduct terms supplying exactly two accessible objects to the binary gate. Summing over possible pairs yields the core workspace transition operator

$$K = \sum_{S,S'} \mathcal{M}_{S,S'} = \sqcup \circ (B \otimes \mathrm{Id}) \circ \Pi^{(2)} \circ \Delta.$$

The schema assumed in this section is depicted in Figure 4 (Figure 4B). With respect to possible empirical relevance, $\Delta$ is the formal source of a strong prediction: retrieval and reactivation should not be modeled only as long-distance dependency length. They require neural access to a previously built sub-object while the current workspace is being recomposed. This predicts temporally specific reactivation of lexico-semantic features of moved (i.e., words that during real-time parsing offer filler-gap resolutions) or retrieved material.

### *4.8. Hopf algebra Markov chains and derivational dynamics*

We now return briefly to derivational dynamics. In this section, the operator $K$ can be viewed as inducing a directed graph over workspaces: each workspace is a vertex, and $K$ places a weighted edge from $x$ to $y$ whenever a single Merge step carries workspace $x$ to workspace $y$. $K$ is built from the Hopf-algebra operations as

$$K = \mu \circ (B \otimes \mathrm{id}) \circ \Pi^{(2)} \circ \Delta,$$

with the coproduct $\Delta$ extracting accessible sub-objects, the binary projector $\Pi^{(2)}$ selecting two-object extractions, the gate $B$ merging the selected pair while the identity leaves the remainder untouched, and the product $\mu$ reassembling the result into the updated forest. Compactly, $K = \sqcup \circ \mathcal{B} \circ \Delta$, where $\sqcup = \mu$ is the product (disjoint union of forests) and $\mathcal{B} = (B \otimes \mathrm{id}) \circ \Pi^{(2)}$ is the grading-preserving gate. Because the workspace bialgebra is graded and connected ($H_0 = k$), an antipode exists by the standard recursion and the Hopf structure is genuine rather than weak; no blockwise normalization is required.

The relevant grading is by total lexical material. Writing $H = \oplus_{l \geq 0} H_l$, with $H_l$ spanned by a finite basis $W_l$ of workspaces containing $l$ lexical items, Merge combines two objects into one without adding or deleting lexical items; it therefore preserves $l$. External Merge additionally reduces component count by one. Internal Merge leaves it unchanged. Consequently $K(H_l) \subseteq H_l$, and $K$ restricts to a non-negative matrix $K_l$ on $W_l$ with entries $K_l(x,y) \geq 0$. A grading-preserving non-negative operator does not yet define a Markov chain; it must be normalized.

Note first what $K_l$ is. Its entries count derivational moves: $K_l(x,y)$ is the number of distinct (admissible cut, selected pair) choices carrying workspace $x$ to workspace $y$ in one Merge step. $K_l$ is therefore the adjacency matrix of the Merge multigraph, in which parallel edges are distinct moves, and a derivation is an edge-path in that multigraph rather than a vertex-path in its underlying simple graph. Let us normalize by a positive eigenfunction: If $K_l$ is irreducible, the Perron–Frobenius theorem supplies a positive eigenvalue $\lambda$ (its spectral radius) and a strictly positive right eigenvector $\psi$ with

$$\sum_y K_l(x,y)\,\psi(y) \;=\; \lambda\,\psi(x), \quad \psi(x) \;>\; 0.$$

The ground-state transform (a Doob transform) then yields a stochastic matrix,

$$P(x,y) \;=\; \frac{\psi(y)}{\lambda\,\psi(x)}\,K_l(x,y), \quad \sum_y P(x,y) \;=\; 1.$$

Because $K_l$ is a multiplicity matrix rather than a matrix of free weights, this normalization is not one arbitrary choice among many: the resulting chain is exactly the maximal-entropy random walk on the Merge multigraph, the unique chain whose entropy rate attains the topological entropy $log\,\lambda$ and which assigns equal probability to all derivations of the same length and endpoints (Marcolli & Skigin, 2025). Explicitly, for any derivation $x_0 \;\to\; x_1 \;\to\; \cdots \;\to\; x_n$,

$$P(x_0 \;\to\; \cdots \;\to\; x_n) \;=\; \frac{\psi(x_n)}{\psi(x_0)} \cdot \lambda^{-n},$$

which depends only on the endpoints and the length.

A biologically interpretable deformation follows by Gibbs (Boltzmann) reweighting of the same operator, rather than by a separate construction. Let $C(x,y)$ be a cost combining grammatical, real-time processing, and neural-state contributions, let $\beta \;\geq\; 0$ be an inverse temperature, and set

$$K_{l,\beta}(x,y) := e^{-\beta C(x,y)} K_l(x,y).$$

Here $C(x,y) \geq 0$ is an edge cost on admissible Merge transitions. For a derivational path $\gamma = (x_0, \dots, x_n)$, the total cost is additive:

$$C(\gamma) = \sum_{i=0}^{n-1} C\,(x_i, x_{i+1}).$$

In the Marcolli-Skigin formulation, the standard terms encode Minimal Search, Minimal Yield, and complexity loss; importantly, these alone do not force convergence to completed single-tree workspaces. This requires the additional state potential

$$C_{\mathrm{Sh}}(x) := \mathcal{H}(P_x) = -\sum_{j=1}^{r(x)} P_x\,(j) \log P_x(j), \qquad P_x(j) := \frac{m_j}{\sum_{k=1}^{r(x)} m_k},$$

where $r(x)$ is the number of tree components in workspace $x$ and $m_j$ is the number of lexical items in its $j$th component. Hence $C_{\mathrm{Sh}}(x) = 0$ for a single-tree workspace and is positive for a fragmented forest. The symbol $\mathcal{H}$ is used here to distinguish this state potential from the local branching entropy $H(x)$ defined below.

Applying the same Doob transform to $K_{l,\beta}$ – with $(\lambda_\beta, \psi_\beta)$ its Perron pair – yields

$$P_\beta(x,y) = \frac{\psi_\beta(y)}{\lambda_\beta \, \psi_\beta(x)} \, e^{-\beta C(x,y)} \, K_l(x,y),$$

a one-parameter family of stochastic matrices. At $\beta = 0$, this recovers the maximal-entropy walk exactly; as $\beta \to \infty$, the Perron-Frobenius problem tropicalizes and stationary mass concentrates on the minimum-cost critical class – here, with the Shannon term, the completed single-tree sector. The temperature therefore interpolates between exploration and commitment along a single construction rather than switching between two.

We stress that this global normalization is not interchangeable with the row-wise alternative

$$P_\beta^{row}(x,y) := \frac{e^{-\beta C(x,y)} K_l(x,y)}{\sum_z e^{-\beta C(x,z)} \, K_l(x,z)}$$

which at $\beta = 0$ reduces to the simple random walk – uniform over admissible moves at each step – rather than to the maximal-entropy walk. The two coincide only on regular graphs, and they differ in a way that is empirically consequential here: the maximal-entropy walk concentrates on the most richly connected region of the workspace graph, whereas the simple random walk distributes mass by local out-degree. They therefore predict different distributions of derivational dwelling across workspace states, and the contrast is available to any analysis that can estimate state occupancy.

Either normalization yields a Markov chain on $W_l$ with, when it exists, a stationary distribution $\pi$ satisfying $\pi P = \pi$. Two entropy quantities organize the dynamics. The *local branching entropy* at a workspace $x$,

$$H(x) = -\sum_y P\,(x,y) \log P(x,y),$$

measures the uncertainty of the next Merge step from $x$. The *entropy rate* is its stationary average,

$$h(P) = \sum_x \pi\,(x)\, H(x) = -\sum_x \pi\,(x) \sum_y P\,(x,y) \log P(x,y).$$

The empirical implication is that parsing is not merely the traversal of a single best-predicted tree. It is a constrained stochastic flow through workspace states, in which the local branching entropy $H(x)$ falls when filters close derivational options and rises when an incoming word opens new ones (for example, newly available attachment sites; 'opening nodes', etc.), while $h(P)$ summarizes the long-run rate. This gives a principled way to ask whether broadband $\gamma$ activity, single-unit population diversity, or low-frequency activity track the local branching entropy $H(x)$ – that is, state-space branching – rather than raw syntactic difficulty.

Whether this chain is ergodic depends on which Merge operations are included, and the two operations act differently on forest-component count. External Merge combines two separate trees into one and reduces the number of components by one. Internal Merge re-merges a subterm within a single tree and leaves the component count unchanged. This is what reconciles the ergodicity of Internal Merge noted in §3.3 with the absorbing behavior here: on a fixed component-count sector, component-preserving Internal Merge is strongly connected and aperiodic with a uniform stationary distribution (Marcolli & Skigin, 2025), whereas the component-reducing drift supplied by External Merge carries the system across sectors toward the completed single-tree workspaces at which no external combination remains.

With only External and Internal Merge, the net dynamics are therefore absorbing in component count – though not in workspace state. The completed single-tree workspaces form a closed communication class rather than a set of absorbing points, and Internal Merge continues to act within it, so the limiting distribution is uniform over completed trees rather than concentrated on any one of them (Marcolli & Skigin, 2025). Adding what is known as 'Sideward Merge' closes the graph into a strongly connected, aperiodic, and hence ergodic chain, with a stationary distribution and nonzero entropy rate. Sideward Merge merges an accessible term extracted from one syntactic object with a second object, leaving the first as a quotient; when both merged terms are proper extracted subterms, the workspace ends up with more, and smaller, components – which is why its arrows return probability mass to less-connected states. Still, it is contested in the literature (Chomsky et al., 2023), and is invoked for a narrow range of phenomena, head-to-head movement being the standard case, and is standardly excluded on optimality grounds.

It is important to separate what depends on irreducibility from what does not. The row-normalized chain, and with it the local branching entropy $H(x)$ and – wherever a stationary distribution exists – the entropy rate $h(P)$, are defined for any generator, including the reducible External/Internal chain: $H(x)$ is a property of the out-transitions from $x$ and requires nothing global. The neural prediction stated above, that some signal tracks $H(x)$ rather than word-string difficulty, therefore does not depend on Sideward Merge. What does depend on it is the maximal-entropy characterization. The Doob transform requires a strictly positive $\psi$, which Perron-Frobenius supplies only for an irreducible generator. In the External/Internal regime $\psi$ may vanish on transient states, and the globally normalized chain is not guaranteed by this route. Sideward Merge enters only as the condition under which the unique global maximal-entropy walk is available, and the contrast drawn above between the two normalizations is a claim about that walk. A reader who rejects Sideward Merge would be free to keep $H(x)$ and keep entropy rates defined relative to a specified stationary measure on each closed class (for a reducible chain the stationary distribution need not be unique, so $h(P)$ is relative to that choice): what is lost is only the unique global maximal-entropy characterization and with it the identification of the $\beta = 0$ limit with the maximal-entropy walk. Figure 4 depicts the workspace transition system and these entropy signatures (Figure 4C).

This Markov-chain formulation replaces the standard metaphor of language comprehension as being a kind of *meter* (whereby difficulty accumulates across the sentence) with a *map*. At each moment the parser occupies a state – a workspace – and the incoming word changes not how full the meter is but which moves are available from where the system currently stands. The local branching entropy H(x) is how many ways the derivation can go next; it rises when a word opens attachment sites and falls when a filter closes them. The entropy rate h(P) is its long-run average. One immediate consequence for the neurobiology of language is a change of regressor. Instead of asking whether a signal tracks syntactic difficulty or node-related processing load, one asks whether it tracks H(x) – a property of the internal state space rather than of the word string. A signal that follows word rate or lexical surprisal but *not* H(x) is therefore tracking the statistics of the input, not the underlying derivation.

*4.9. Function-space realization: thermodynamic semirings and Rényi gates*

Perhaps the strongest bridge from these algebraic constraints toward genuine neural mechanisms comes from the function-space construction (Marcolli & Berwick, 2026), because it realizes the abstract composition law as an explicit operation on functions of the kind a neural system could plausibly carry. Following the Marcolli–Berwick formulation, assume lexical atoms and features are encoded as bounded functions on a bounded region $X \subset \mathbb{R}^D$ of a Euclidean space:

$$\iota: SO_0 \to C^\infty(X, \mathbb{R}) \subset L^\infty(X, \mathbb{R}).$$

Nothing in the construction requires $X$ to be a time axis. Taking $D = 4$ with coordinates $(t, x, y, z) \in \mathbb{R}^4$ makes $X$ a region of cortical spacetime, and the eventual tree encoding $\Psi_{S,\beta}(T)$, defined below, a spatiotemporal field rather than a waveform. Because the mixture parameter is fixed by a pointwise optimization over the domain, $\lambda_T$ is then a field $\lambda_T: X \to [0,1]^{|V^\circ(T)|}$ defined over cortex as well as over time. Here $V^\circ(T) = V_{\text{int}}(T)$ is the set of internal vertices of $T$, $L(T)$ denotes its leaf set, and we abbreviate the ambient function space as

$$\mathcal{F} := L^\infty(X, \mathbb{R}).$$

Importantly, various of the traveling-wave readings of cross-frequency coupling (Muller et al., 2026; Murphy, 2025) (e.g., §7.1), on which the direction and velocity of a low-frequency wave determine the order in which populations become excitable, is on this reading not an additional hypothesis grafted onto the function-space account: a spatially varying $\lambda_T$ over $X$ *is* a phase-gradient configuration, and coproduct extraction, composition and reassembly correspond to distinct gradient fields of $\lambda_T$ rather than to a generic increase in PAC. Likewise, the effective-dimensionality readout we will introduce in §9.2 becomes a spatiotemporal field $S_2\big(\lambda^\star(t, x, y, z)\big)$ rather than a scalar per trial, with the prediction that the direction of the composing population's dimensionality gradient should align with the wave direction over the same cortical patch.

One brief caveat should be highlighted here: Marcolli and Berwick's faithfulness result is a transversality statement (Marcolli & Berwick, 2026), and its genericity is stated with respect to a dense open set of maps $\varphi: \mathcal{SO}_0 \to C^\infty(X, \mathbb{R})$. It is therefore a claim in the smooth category, and a realization that encodes atoms in a non-smooth or finite-dimensional function class does not inherit it automatically.

The natural first attempt at a function-space realization is a superposition of the leaves' wavelets weighted by the tree, and its failure is what motivates the entropy term (Marcolli & Berwick, 2026). Assign to each internal vertex $v$ a mixing parameter $\lambda_v$ and combine its daughters linearly

$$\Lambda_v(\varphi, \psi) \;=\; \lambda_v\, \varphi \;+\; (1 - \lambda_v)\, \psi, \qquad \Lambda = (\lambda_v) \in [0,1]^{V^\circ(T)}.$$

Composing these down the tree distributes the weight across the leaves as a path product. Writing $\gamma_{v_0,\ell}$ for the path from the root $v_0$ to leaf $\ell$, and $e_v$ for the edge of that path with source $v$ oriented away from the root

$$T \mapsto \Lambda_T\big(\iota(\alpha)_{\ell \in L(T)}\big) = \sum_{\ell} a_\ell^T(\Lambda)\, \iota(\alpha_\ell), \qquad a_\ell^T(\Lambda) = \prod_{v \in \gamma_{v_0,\ell}} \epsilon_v,$$

with $\epsilon_v = \lambda_v$ when $e_v$ lies on the branch selected at $v$ and $\epsilon_v = 1 - \lambda_v$ otherwise, so that $A = \left(a_\ell^T(\Lambda)\right)_{\ell \in L(T)}$ is a probability distribution on the leaves.

This construction fails on three counts, and each failure identifies something a neural code would have to supply from elsewhere. First, nothing in the grammar fixes $\Lambda$; the map is a family of maps, not a map. Second, if $\Lambda$ is retained as a free variable then $\Lambda_T\big(\iota(\alpha)_{\ell \in L(T)}\big)$ lives in $\mathcal{F} \otimes \mathbb{R}[\Lambda]$ rather than in $\mathcal{F}$, so the output is not an object of the same type as the inputs and closure fails at the first step. Third, and most damagingly for present purposes, the distinction between the $\lambda_v$ side and the $1 - \lambda_v$ side of each vertex is a *planar* structure: it orders the daughters. The resulting map is therefore non-commutative, unlike free Merge, and by §4.2 it has smuggled externalization into narrow syntax.

The entropy optimization repairs all three at once. Minimizing a free-energy functional over Λ fixes the parameters variationally rather than by hand. The minimum is attained at a point of the simplex, so the output returns to $\mathcal{F}$, and because the objective depends on the daughters only through a term that the information functional renders symmetric, the planar asymmetry is quotiented away. But this is not the only repair: Fixing every λv at ½ also discharges all three failures – Λ is no longer free, the output is in $\mathcal{F}$, and averaging is symmetric – and does so with no entropy term at all. What the entropy term adds, and what a fixed-½ mean-pooling law lacks, is that λv is content-dependent: a specific function of the daughters' pointwise values rather than a constant. That content-dependence is, essentially, the empirical claim of this section. It is what separates the gate from mean-pooling in the simulations of §4.9.1, where the two laws share the order-blind, bracketing-decodable signature of §7.2 and part company only on contrasts that a fixed mixture cannot resolve. The nonlinearity anticipated above is the price of remaining inside the function space while discharging the order that the linear construction cannot avoid introducing.

Let $S$ be an information functional and $\beta > 0$ a thermodynamic (inverse-temperature) parameter – the same role $\beta$ plays in the Gibbs dynamics of §4.8 above. The target of the embedding is a function semiring,

$$\mathcal{R}_{S,\beta} = \big(L^\infty(X, \mathbb{R}), \oplus_{S,\beta}, +\big),$$

in which the additive operation is not ordinary pointwise vector addition but a deformed, entropy-optimized combination. For two inputs $x, y \in L^\infty(X, \mathbb{R})$, evaluated pointwise on $X$,

$$x \oplus_{S,\beta} y = \min_{\lambda \in [0,1]} \{\lambda\, x + (1 - \lambda)\, y - 1/\beta\; S(\lambda)\},$$

a free-energy-style trade-off between a convex mixture of the two inputs and an entropy term scaled by the temperature $1/\beta$. As $\beta \to \infty$ (low temperature) the entropy correction vanishes and $\oplus_{S,\beta}$ reduces to the tropical (min-plus) combination; as $\beta \to 0$ it becomes entropy-dominated. For a probability vector $p = (p_i)_{i=1}^n$, with $p_i \geq 0$ and $\sum_i p_i = 1$, the second Rényi entropy is

$$S_2(p) = -\log \sum_{i=1}^{n} p_i^2$$

so that for a binary mixture $S_2(\lambda) = -\log[\lambda^2 + (1-\lambda)^2]$, which, substituted for $S$, yields the Rényi gate applied at each Merge node. Two properties recommend the second order specifically. The first is expository: it is the order at which the entropy of a binary mixture takes the simple closed form used above, and at which the Marcolli–Berwick faithfulness result (their Thm 4.6) is stated. The second is substantive, and bounds the order from above. To consider a general Rényi order, fix a point of $X$ and let $f$ and $g$ denote the scalar values of the two input functions at that point. Let $\alpha > 0$, $\alpha \neq 1$, denote the Rényi order. The pointwise objective is then

$$J(\lambda) \ = \ \lambda f + (1-\lambda)g + \frac{1}{\beta(\alpha-1)}\log(\lambda^\alpha + (1-\lambda)^\alpha),$$

and set $u(\lambda) = \lambda^\alpha + (1-\lambda)^\alpha$, so that $\beta(\alpha-1)J''(\lambda) = (u''u - u'^2)/u^2$ with

$$u''(\lambda) \ = \ \alpha(\alpha-1)[\lambda^{\alpha-2} + (1-\lambda)^{\alpha-2}].$$

At the boundary $\lambda \to 0^+$, where $u \to 1$ and $u' \to -\alpha$, the curvature $J''(0^+) = (u''(0^+) - \alpha^2)/\beta(\alpha - 1)$ diverges to $+\infty$ for $\alpha < 2$, vanishes exactly at $\alpha = 2$, and equals $-\alpha/\beta(\alpha-1) < 0$ for $\alpha > 2$.

The second Rényi order is therefore the largest order at which the gate's variational problem can be convex in $\lambda$. The argument here is a boundary one (it evaluates $J''$ as $\lambda \to 0^+$) and so establishes that convexity fails above $\alpha$ = 2, not that it holds below; the exclusion is what the present purpose requires, since it is the discontinuity of the minimiser above $\alpha$ = 2, and not convexity as such, that disqualifies those orders as neural codes.

Above this order the objective is concave near the boundary and the minimiser is discontinuous: at $\alpha = 3$ the optimal mixture jumps by $0.29$ as the pointwise difference between the daughters crosses $\delta \approx 1.66$, and the jump grows with order (0.05 at $\alpha = 2.05$, 0.21 at $\alpha = 2.5$, 0.39 at $\alpha = 5$), whereas at $\alpha \leq 2$ the mixture reaches the boundary continuously. A gate whose output jumps under an infinitesimal perturbation of either daughter is not a candidate for implementation by bounded, noisy population dynamics. Non-associativity does not bracket the order from below.

It is worth commenting on what exactly is being deformed here, since the (min,+) semiring is not an unusual base for this construction. It is the algebra of dynamic programming and surfaces wherever a system computes a best path rather than a sum over paths. Shortest-path and Viterbi recursions are min-plus matrix products, and Bellman-type optimality equations are min-plus linear (Baccelli et al., 1992; Mohri, 2002). The point is immediate for parsing, where one and the same recursion returns the best derivation when evaluated over the tropical semiring and the inside probability when evaluated over the sum-product semiring (Goodman, 1999). The finite-$\beta$ family used here interpolates between exactly those two regimes, which supplies an independent reading of the temperature as the degree to which the system commits to a single derivation rather than maintaining a distribution over alternatives. Importantly, the semiring has a history in neural modeling. Lattice-algebraic and morphological neuron models replace the multiply-accumulate unit with max-plus or min-plus dendritic accumulation, based on the argument that thresholding and competitive selection are more plausible primitives for single-neuron computation than weighted linear summation (Maragos, 2017; Ritter & Urcid, 2003), and rectified-linear networks compute tropical rational functions so that their decision boundaries are tropical hypersurfaces (Charisopoulos & Maragos, 2018). The gate proposed here is consequently not a novel algebra imported into the neurosciences, but a finite-temperature deformation of one already entertained. What is distinct here is the claim that the deformation parameter and the choice of entropy functional are the linguistically contentful objects.

The deformation is a thermodynamic semiring in the sense of Marcolli and Thorngren (Marcolli & Thorngren, 2014): the tropical semiring's addition is deformed by an entropy functional at finite temperature while its multiplication is unchanged. Let a binary information measure be a continuous concave function $S: [0,1] \to \mathbb{R}_{\geq 0}$, and consider the induced operation on $\mathbb{R} \cup \{\infty\}$

$$x \oplus_{S,\beta} y \;=\; \min_{p\in[0,1]} \{p\, x + (1-p)\, y - \beta^{-1} S(p)\}$$

with tropical addition $x \oplus y = \min\{x, y\}$ recovered in the zero-temperature limit. The properties of $\oplus_{S,\beta}$ then stand in exact correspondence with axioms on $S$:

- $\oplus_{S,\beta}$ is **commutative** if and only if $S(p) = S(1-p)$;
- $\oplus_{S,\beta}$ has **additive unit** $0$ if and only if $S(0) = 0 = S(1)$;
- $\oplus_{S,\beta}$ is **associative** if and only if $S$ satisfies extensivity; for $0 \leq p < 1$ and $0 \leq q \leq 1-p$, with $p + q > 0$, extensivity is the identity

$$S(p) + (1-p)\, S\left(\frac{q}{1-p}\right) \;=\; S(p+q) + (p+q)\, S\left(\frac{p}{p+q}\right)$$

and all three hold together precisely when $S$ is the Shannon entropy.

This correspondence is worth stating in full because it is perhaps the clearest instance of the NAP's central claim available to us. The commutativity of Merge is not merely compatible with the gate; under this construction it directly *is* the symmetry of the information measure that the gate optimizes. Likewise, non-associativity is not merely tolerated – it is the failure of extensivity, which is to say the failure of the entropy of a grouped mixture to be recoverable from the entropies of its parts. The algebraic contrast that we stated in §4.1 as a general fact about trees now reappears here as an information-theoretic contrast about how a system may be permitted to score its own uncertainty when it combines two objects. A candidate neural gate that is symmetric in its daughters and non-extensive in its scoring is thereby a candidate Merge; one that is extensive is not, whatever else it does.

One brief qualification is needed here in advance of the following discussion. This correspondence is a statement about $\oplus_{S,\beta}$ as an operation on values, whose output is the minimum of the objective. The offset-corrected gate we will recommend in §4.9.1 keeps the argmin and discards the minimum value. It inherits commutativity from the symmetric optimization problem; its non-associativity must be established independently, and §4.9.1 shows that it remains non-associative even at the Shannon point, where $\oplus_{S,\beta}$ is associative. The biconditionals above therefore do not carry over to it, and the clearest-instance claim is made of the gate in its original form.

Specializing $S$ to the second Rényi entropy therefore selects, within the commutative and unital family, the non-associative deformation. There is an additional and independent motivation for the second order that our convexity argument does not supply. The second Rényi entropy is the standard concentration functional for optimizing expansions in redundant, non-orthogonal bases in signal analysis; optimizing a linear combination of wavelets with respect to it is the native use of the quantity. This matters here because the atoms of §6.3 are precisely a redundant, non-orthogonal dictionary, and the mutual-coherence prediction $\mu_{\text{syn}} < \mu_{\text{sem}}$ is a claim about how nearly that dictionary approaches a basis. The order of the gate and the geometry of the lexicon are consequently not independent parameters of the model, and the convexity bound of $\alpha \leq 2$ and the signal-analysis provenance of $\alpha = 2$ converge on the same value from different directions.

The curvature argument does not exclude orders below the Shannon point either: for $\alpha < 1$ both $(\alpha - 1)$ and $u''$ change sign, so $J''(0^+)$ again diverges to $+\infty$ and the boundary curvature raises no objection. The admissible range is therefore $\alpha \leq 2$ with $\alpha \neq 1$ – bounded from above by the failure of convexity and punctured at the Shannon point – and $\alpha = 2$ is an endpoint of that range rather than a convenience. We note here in advance that the puncture does *not* survive our simulations reported in §4.9.1: $\alpha \neq 1$ is a property of the unnormalized gate as stated here, and not of the offset-corrected gate we ultimately recommend, whose associator is nonzero at every order.

Intuitively, second-order Rényi entropy measures how many dimensions or population components are doing meaningful work in a representation, making it a natural way to quantify whether composition preserves a distributed structure or collapses toward a single dominant state. More specifically, second-order Rényi entropy measures the diversity of the optimized mixture weights in the formal construction. Yet, a relationship to neural participation ratio is possible only under additional assumptions about how those mixture weights are expressed in population activity; we state these separately in §9.2.

In the meantime, why this choice of entropy measure? A nearby result must be distinguished from this one. Marcolli and Skigin weight the Merge graph by the standard linguistic cost functions (Minimal Search, Resource Restrictions, complexity loss) and ask whether the weighted dynamics converges on connected structures. Because Internal Merge carries zero cost under those functions, every Internal Merge self-loop is a cost-minimal circuit; the associated tropical (min-plus) Perron–Frobenius problem is therefore degenerate, returning a spanning family of eigenvectors indexed by every partition containing a component of three or more leaves rather than a unique eigenvector concentrated on the single-tree states. This does not establish that the standard cost functions fail – the true asymptotics need not coincide with any generator of that eigenspace – but it does establish that they do not, by themselves, certify convergence. What does certify it is a Shannon-entropy term $Sh(P_{\wp})$ on the distribution $P_{\wp}$ of lexical items across workspace components, which vanishes exactly on single trees and is strictly positive otherwise; adding it makes the critical circuit unique and drives the stationary distribution, in the low-temperature limit, to the uniform distribution on completed trees. The term suffices on its own: in the limiting dynamics the standard cost functions contribute nothing, and serve only to restrict Sideward Merge to its minimal form. Note that it is a potential on workspace *states*, not a cost on operations, and so defines a free-energy landscape over the workspace rather than a per-step search cost, and by Stirling's approximation it is, to leading order, $n^{-1}$ times the logarithm of the multinomial coefficient counting how lexical material can be distributed across components – a log-degeneracy rather than a processing cost. That derivational result is a Shannon ($\alpha \to 1$) statement, whereas the compositional gate here uses collision entropy ($\alpha = 2$) – different functionals doing different jobs. The distinction matters because, for the gate as originally stated, $\alpha \to 1$ is the associative limit at which bracketing sensitivity vanishes, so that gate requires $\alpha \neq 1$.

For the offset-corrected gate of §4.9.1 the inference is unavailable – it is non-associative at the Shannon point – though $\alpha$ still matters there for a different reason (see §4.9.1).

We do not claim that the brain implements order α = 2 specifically; a different order would yield a structurally similar but quantitatively different gate, and the order is itself an empirical parameter. Additionally, the temperature $\beta$ must be read consistently across the gate and the dynamics. Because the $\beta \to \infty$ (tropical, min-plus) limit is associative, the non-associativity on which this framework depends is carried entirely by the finite-$\beta$ entropy correction. For a tree $T$ with $n = |L(T)|$ leaves, internal vertices $V_{\text{int}}(T)$, and lexical labels $\{\alpha_\ell\}_{\ell \in L(T)}$, the construction assigns, pointwise over $X$, one mixture parameter to each internal vertex:

$$\lambda_T : X \to [0,1]^{|V_{\text{int}}(T)|},$$

fixed by

$$\lambda_T = \arg\min_{\lambda} F_{T,S,\beta}(\iota(\alpha_1), \dots, \iota(\alpha_n); \lambda),$$

where $F_{T,S,\beta}$ is the free-energy functional accumulated over the tree – the bracketed sum $\mathcal{B}_{T,S,\beta}$ of Marcolli–Berwick, whose minimum value over $\lambda$ is the encoded function itself. In the equivalent leaf-level formulation, write $A = (a_\ell)_{\ell \in L(T)}$ for the mixture-weight vector and

$$\Delta_n := \left\{ A \in \mathbb{R}^n_{\geq 0} : \sum_{\ell=1}^{n} a_\ell = 1 \right\}$$

for the leaf simplex. Thus, the symbols denoting lexical labels are distinct from the coefficients $a_\ell$, which denote their optimized mixture weights. Explicitly, the bracketed sum is a single minimization over the leaf simplex

$$\mathcal{B}_{T,S,\beta}(x_1, \dots, x_n) \;=\; \min_{A \in \Delta_{\#L(T)}} \left\{ \sum_{\ell} a_\ell \, x_\ell - \beta^{-1} S_T(A) \right\},$$

and it agrees with the non-associative bracketing of the binary operations $\oplus_{S,\beta}$ according to the tree $T$. The tree structure enters entirely through $S_T$, which is generated from the binary measure $S$ by a chain rule over the vertices. For the three-leaf tree $T = \{x_1, \{x_2, x_3\}\}$, the bracketing

$$\mathcal{B}_{T,S,\beta}(x_1, x_2, x_3) \;=\; x_1 \oplus_{S,\beta} \left( x_2 \oplus_{S,\beta} x_3 \right)$$

is realized by

$$S_T(A) \;=\; S\left( \frac{a_2}{1 - a_1} \right) + (1 - a_1)\, S(a_1).$$

The two terms are the two vertices: $S(a_1)$ scores the root's split of mass between $x_1$ and the constituent $\{x_2, x_3\}$, and $S\big(a_2/(1 - a_1)\big)$ scores the inner vertex's split of the mass *conditional* on that constituent having been formed, discounted by the mass $(1 - a_1)$ that reaches it. Reversing the bracketing to $\big((x_1 x_2) x_3\big)$ permutes which conditionalization is nested inside which, and this is exactly where non-associativity resides: $S_T$ is a function of the bracketing and not only of $A$. It is also where the failure of extensivity does its work, since extensivity is precisely the identity that would make the two orders of conditionalization agree.

Two consequences follow for the predictions we will make later. First, the leaf weights of §9.2 are not free: $A$ is the path product of the vertex mixtures, so $a_\ell$ is determined by the tree together with the optimized $\Lambda$, and the leaf-mixture distribution whose collision entropy we identify with $\log\mathrm{PR}$ is therefore a tree-recursive object rather than an unstructured weighting. Second, the sign and magnitude of the predicted effective-dimensionality separation between two matched bracketings are computable in advance from this chain rule, for any candidate tree pair, and should be computed rather than left as a direction to be read off the data.

The resulting embedding

$$\Psi_{S,\beta}: T_{SO_0} \to \mathcal{R}_{S,\beta}, \qquad \Psi_{S,\beta}|_{SO_0} = \iota$$

is, under the conditions established by Marcolli–Berwick, a morphism of commutative non-associative magmas

$$\Psi_{S,\beta}\big(\mathcal{M}(A,B)\big) = \Psi_{S,\beta}(A) \ \oplus_{S,\beta} \ \Psi_{S,\beta}(B).$$

This is a concrete realization of the structure-preserving encoding $\mathcal{N}$ at the beginning of §4: the physiological combination operation $G$ that makes the composition square commute is here instantiated as the Rényi gate $\oplus_{S,\beta}$, with $\Psi_{S,\beta}$ as the encoding map.

Two properties make it the right kind of object. First, it is commutative and non-associative by construction, so it respects the magma laws of §3.1 and §4.1 rather than flattening them, as ordinary addition would. Second, its faithfulness – for the gate as originally stated – is exact but conditional. Marcolli and Berwick establish generic injectivity of $\Psi_{S,\beta}$ (their Thm 4.6), not graded decodability: distinct syntactic objects are separated outright, but only inside a high-temperature window whose width shrinks toward zero as the number of leaves grows. Size and depth therefore enter through the admissible $\beta$-window rather than through any loss of information in the idealized construction. Graded degradation of recoverability enters only in a bounded, noisy realization of the gate – a regime their competence-level construction does not model – and it is there that constituent structure becomes progressively harder to recover from $\Psi_{S,\beta}(T)$ as the tree grows in size and depth (a degradation in the size/depth direction, distinct from the temperature trade-off discussed above).

This provides an exact path to translate a core design feature of language into an empirically testable space. A neural composition mechanism can be tested against a graded family of combination models: (i) linear-additive, (ii) rotate-then-sum, (iii) role-filler, (iv) recurrent, and (v) entropy-optimized nonlinear (Rényi). These are not degrees of one quantity; they differ in *which* of the three magma laws they violate, and the comparison must be conjunctive rather than scalar. If the entropy-optimized model explains variance in, say, broadband $\gamma$ activity, spike-field coupling, or population trajectories beyond lexical and surprisal controls, the algebraic hypothesis gains mechanistic plausibility. Conversely, if simple vector addition performs equally well once bracketing is controlled, the strong version of the present model is weakened.

The faithfulness prediction has to be stated carefully, because it does not license the comparison it is often taken to license. Faithfulness is a claim about the gate's own profile – exact recovery of constituency from $\Psi_{S,\beta}(T)$, within a $\beta$-window that narrows with size and depth – not a claim that the gate must 'out-decode' its rivals as depth grows. Indeed, as we show immediately below (§4.9.1), it does not, and it cannot be expected to: a law that assigns its daughters to distinct roles represents strictly more than the magma contains, and in these simulations recovers bracketing more sharply as a result. The comparison that does discriminate is a conjunction of properties rather than a single accuracy axis, and we will return to it below (§4.9.1, §7.2).

In summary, the second-Rényi construction is useful not because the online parser must literally behave commutatively at all stages of processing, but because it supplies one explicit nonlinear composition law capable of preserving non-associative constituent structure in a continuous function space. In plain terms, the system tries to form a phrase only when the candidate combination becomes sufficiently stable, coherent, or compressible, with the general goal being uncertainty-reduction (Murphy, Holmes, et al., 2024). Figure 5 depicts the presently discussed gate (Figure 5A).

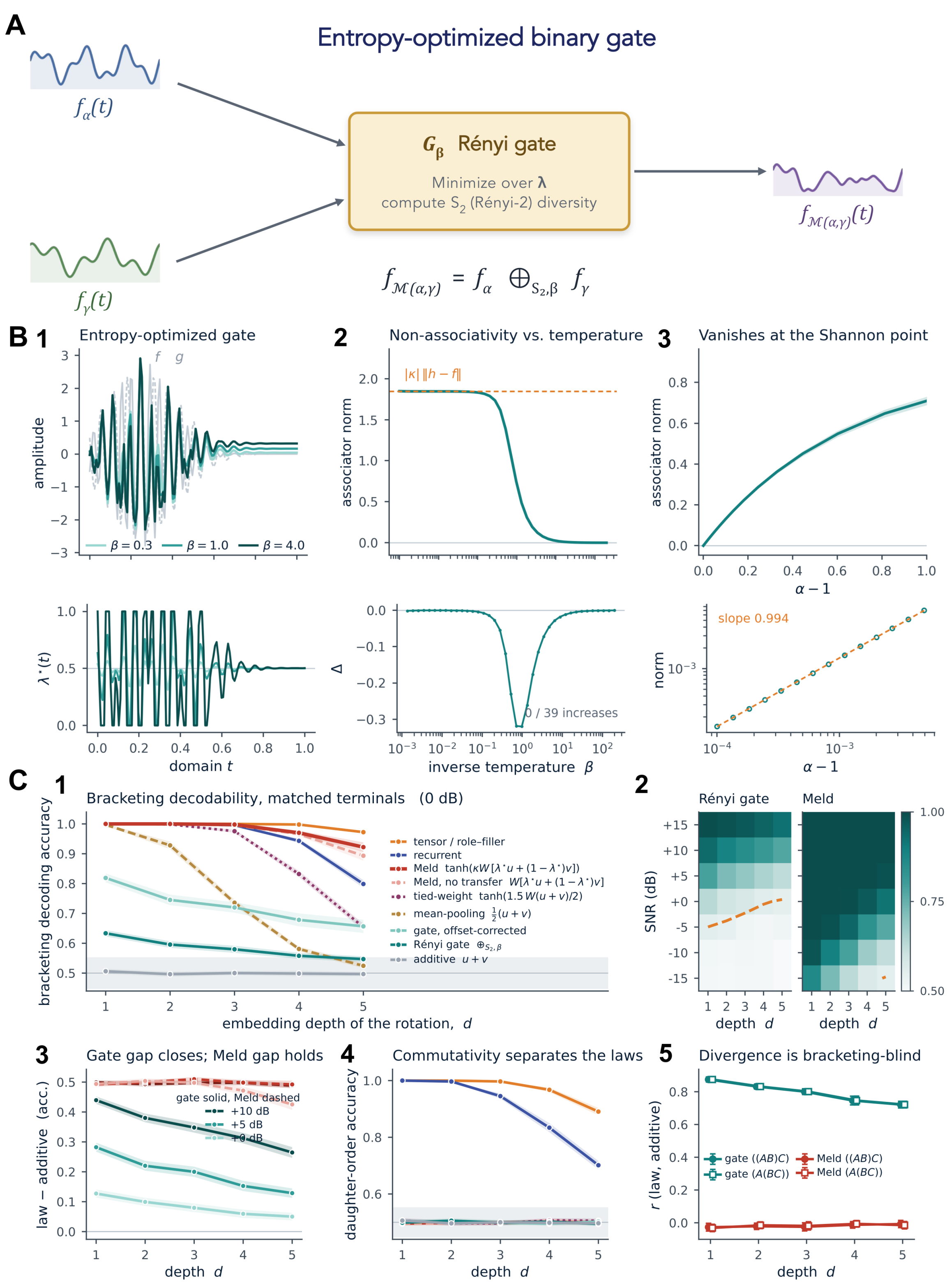

**Figure 5. Representing and simulating composition gates. (A)** Entropy-optimized binary gate. Lexical atoms are encoded as functions $f_\alpha(t), f_\gamma(t)$ on a compact domain (here a time axis). Merge is realized not as vector addition but as the gate $G_\beta$, which combines its two inputs by the entropy-optimized rule $f_{\mathcal{M}(\alpha,\gamma)} = f_\alpha \oplus_{S_2,\beta} f_\gamma$: at each point it minimizes a free-energy trade-off over a mixture parameter $\lambda$, with the second Rényi entropy $S_2$

measuring mixture diversity and $\beta$ setting the temperature. The resulting operation is commutative and non-associative, and preserves tree structure – precisely so in the ideal construction, and up to size- and depth-dependent limits in any bounded neural realization. **(B)** The entropy-optimized gate, its temperature, and its order. (1) Two Gabor atoms $f, g$ on the compact domain (grey) and the gate output $f \oplus_{S_2,\beta} g$ at three inverse temperatures; below, the pointwise optimal mixture $\lambda^*(t)$. As β grows the gate approaches the tropical minimum and $\lambda^*$ saturates at the boundary; as β falls it approaches an entropy-dominated mixture and $\lambda^* \to$ ½. (2) Associator norm $\|(f \oplus g) \oplus h - f \oplus (g \oplus h)\|$ against inverse temperature; mean and 95% CI over 400 random atom triples. Dashed line: the analytic high-temperature plateau $|\kappa|\,\|h - f\|$ with $\kappa = 1 - \frac{3}{2}\lambda^* \approx 0.1159$, $\lambda^*$ the root of $S_2{}'(\lambda) = -\log 2$; simulation matches it to within 0.1%. Lower panel: successive differences, confirming a monotone non-increasing decline (0 of 39 increases). (3) Associator norm against Rényi order at $\beta = 1$, vanishing at the Shannon point $\alpha \to 1$. Lower panel: log–log fit against $\alpha - 1$, slope 0.994, confirming linear scaling. Simulations were run on custom Python scripts. **(C)** Composition laws separate on order, not on depth. (1) Cross-validated bracketing decoding from the composed root representation at 0 dB, for a height-matched three-terminal rotation embedded at depth $d$. Conditions are matched on terminal identity, terminal order, height, node count and leaf-depth multiset, and differ only in the bracketing of the deepest triple. Shaded band: the upper one-sided exact-binomial significance threshold (0.553; 240 test trials). Mean and 95% CI over 60 items. The tensor and recurrent laws recover bracketing more sharply than the gate, and Meld (heavy red) matches them to within a few points at every depth, and the additive law is at chance throughout. The fixed mean-pooling law ½(*u* + *v*) decodes the rotation as well, and is order-blind, and is separated from the gate only on the balanced four-leaf contrast ((*A B*)(*C D*)) vs ((*A C*)(*B D*)), on which a fixed mixture is at chance at every SNR and the gate is above threshold at intermediate *β* (0.827 at *β* = 0.5, 0 dB). (2) Accuracy over the depth × SNR plane for the Rényi gate (left) and Meld (right). Dashed contour marks the significance threshold. The gate's powered region does not extend below about −5 dB at any depth; Meld's reaches −10 dB at every depth. (3) The law-minus-additive difference at three SNRs for the Rényi gate (solid, teal) and Meld (dashed, red). SNR is defined against the variance of the simulated response, not against any recording, and the encoding is oracle. The axis compares the laws with each other; it does not specify an experiment. Because the additive law is at chance everywhere, this is the divergence the present framework predicts should grow with depth; for the gate it declines, at every SNR, from a maximum at the shallowest bracketing contrast, while for Meld it holds at its ceiling across depth. (4) The same analysis with daughter order as the classified variable on an otherwise identical spine. The gate, mean-pooling, the tied-weight unit and Meld are order-blind at every depth; the tensor and recurrent laws are not. Panels (1) and (4) together give the first two axes of the discriminating signature. Bracketing-decodable and order-blind is satisfied by the corrected gate, by fixed mean-pooling (dashed), by the tied-weight saturating unit $\tanh(1.5\,W(u+v)/2)$ (dotted) – the recurrent law with $W_u = W_v$ – and by Meld (heavy red), which combines the corrected gate's mixture with that unit and decodes bracketing to within a few points of the ordered laws at every depth while remaining at chance on order; the laws that fail the conjunction are the ordered ones. Members of the class are separated by the balanced four-leaf contrast and by sensitivity to daughter disagreement, reported in the text (§4.9.1). (5) Correlation between each law's root representation and the additive law's, plotted separately for the two bracketings, for the Rényi gate (teal) and Meld (red). The gate diverges from the additive law with depth, and Meld is uncorrelated with it at every depth (its shared weights rotate the composite out of the additive subspace), but in both cases identically for the two bracketings, so divergence from an additive null carries no grouping information and cannot license a structural interpretation.

To conclude this section, there are two things the language system must do that remain in constant tension, and which roughly comply with the brain's joint goals of integration and segregation. It must keep structure (the difference between ((*A B*) *C*) and (*A* (*B C*))) but eventually commit to one interpretation. The entropy gate discussed here makes that trade-off a single element, with one qualification that §4.9.1 will make precise. At high *β* the gate commits: it takes the minimum, one daughter wins, and the grouping is gone. This is the tropical limit, and it is the only limit at which the gate is associative. At low *β* the gate hedges: it mixes its two inputs, and the grouping survives in the mixture wherever a mixture can carry it – on a three-terminal rotation it does, but $\lambda\star \to$ ½ there, so the gate degenerates into fixed averaging, which is still non-associative but is no longer faithful on structures that share a content-to-depth map. Non-associativity, the formal residue of constituency, is therefore carried by the entropy correction at finite *β* and lost only in the tropical limit. What is lost at the other extreme is the content-dependence of the mixture, and with it faithfulness. The neural consequence is that ‘how strongly the system is committing’ and ‘how much structure it retains’ are not two independent variables, but neither are they one monotone variable. Retained structure is maximal at intermediate temperature and lost at both ends. Any manipulation that pushes a human comprehender toward early commitment (e.g., time pressure, degraded input) should therefore cost recoverable structure steeply, and a manipulation that suppresses commitment altogether should cost it too. This trade-off is a property of the entropy-optimized mixture taken on its own. §4.9.1 shows that passing that mixture through shared saturating synapses (Meld) removes it: structure is then recovered at every temperature, and $\beta$ becomes a parameter of the mixture’s sensitivity to disagreement rather than a dial between hedging and collapse.

*4.9.1. Simulation of composition laws*

In order to simulate some effects of the above composition laws, we represented lexical atoms as Gabor wavelets $\ell_i(t) = a_i(t)\, cos(\omega_i t + \varphi_i(t))$ on a 128-point compact domain (§4.10), with Gaussian envelopes of random centre and width, carrier frequencies drawn uniformly from 6–26 cycles, each normalised to zero mean and unit variance. The gate

$$f \oplus_{S_\alpha,\beta} g = \min_{\lambda \in [0,1]} \left\{ \lambda f + (1 - \lambda) g - \frac{S_\alpha(\lambda)}{\beta} \right\}$$

was evaluated pointwise by grid minimization over 1,201 knots in $\lambda$. At $\alpha = 2$ the binary objective is convex in $\lambda$, so its minimiser is unique and the grid agrees with a stationarity solve to grid resolution: writing $u(\lambda) = \lambda^2 + (1 - \lambda)^2$, a direct computation gives

$$J''(\lambda) = \frac{8\lambda(1 - \lambda)}{\beta\, u(\lambda)^2} \geq 0 \quad on\ [0, 1]$$

which is the one order used for the recommended gate and for every bracketing and composition-law simulation reported here. For $\alpha \neq 2$ we do not assert global convexity – §4.9 establishes only the boundary behaviour at $\lambda \to 0^+$, not convexity on the interior – but the grid does not require it: a scan over 1,201 knots returns the global minimizer over $[0, 1]$ by exhaustion regardless of the shape of $J$, so the $\alpha$ sweep is valid without a convexity claim, and the discontinuity the grid must resolve arises only for $\alpha > 2$ (§4.9).

We implemented a number of rival composition laws (see also §7.2): (1) additive, $u + v$; (2) fixed mean-pooling, ½($u + v$), the offset-corrected gate with $\lambda$ frozen at ½, which controls for the content-dependence of $\lambda\star$; (3) an ordered role-filler law, $R_L u + R_R v$, with fixed random orthogonal role operators scaled by 1/√2, at a per-level attenuation of 0.707 per daughter; (4) a recurrent law, tanh($W_u u + W_v v$); (5) a saturating mean, tanh($g(u + v)/2$)/tanh($g$), $g$ = 2; (6) a tied-weight law, tanh($W(u + v)/2$) with $W$ orthogonal and $g = 1.5$, which is law (4) with $W_u = W_v$; (7) Meld, defined below: tanh($\kappa\ W\ [\lambda\star u + (1 - \lambda\star)v]$) with $\kappa$ = 2, the same $W$, and $\lambda\star$ the corrected gate’s mixture; and (8) the linear tied-weight mixture $W[\lambda^\star u + (1 - \lambda^\star)v]$, i.e., Meld without the saturating transfer ($\kappa = 1$). Unless stated, $\beta$ = 1 and $\alpha$ = 2. One feature of this design should be kept in mind when the laws are compared across depth. Each law attenuates the deepest triple differently as it is carried to the root (by a factor near $\lambda\star$ per level for the corrected gate and for mean-pooling, by 0.707 for the role-filler law, and through a saturating nonlinearity for the recurrent law), while the noise is scaled to the pooled variance of the whole response. Accuracy at depth therefore confounds the algebra with per-level gain, and the cross-law ordering at $d = 5$ is partly a statement about attenuation. We therefore also report, in the released code, the discriminability $d' = \|x_1 - x_2\|/\sigma$ of the two condition means, which fixes nearest-centroid accuracy under isotropic noise and makes the gain explicit (at 0 dB and d = 1: role-filler 16.3, recurrent 15.9, tied-weight unit 14.8, Meld 14.4, linear tied mixture 13.4, mean-pooling 6.4, corrected gate 2.2). The comparison that reflects the algebra is the one at the level where the contrast is formed.

Bracketing contrasts were height-matched three-terminal rotations, $((A\ B)\ C)$ versus $(A\ (B\ C))$, embedded $d - 1$ levels below the root on a right-branching structure ($d = 1 \ldots 5$). The two conditions share height, node count, terminal identity, terminal order and leaf-depth multiset, and differ only in the bracketing of the deepest triple. Order contrasts were constructed identically, with $(A\ B)$ versus $(B\ A)$ at the deepest pair. Sixty items – independent lexicons – were run per cell. Isotropic Gaussian noise was added to the composed root representation at SNRs from −15 to +15 dB, defined against the variance of the response pooled over items, conditions and the domain with the grand mean removed, since a constant offset is not an observable. Decoding used a nearest-centroid classifier – the optimal decoder under isotropic noise – with 120 training and 120 test trials per condition; the per-item significance threshold is 0.553 (one-sided binomial, $p < .05$, 240 test trials). Associator statistics are over 400 random atom triples.

With respect to the reported axis, the signal power against which noise is scaled is a property of the simulated response, not of any recording, and the encoding is an oracle so the accuracies reported here are upper bounds on what a decoder could achieve from these states rather than estimates of what a decoder would achieve from cortical recordings. Consequently, the nominal decibel values should not be read as a specification for an experiment. What can be carried over is the ordering: which law loses the contrast first, how far apart they sit at a given SNR, and how fast each falls off with depth. Turning the axis into a statement about feasibility would mean calibrating it against measured single-trial signal-to-noise in the relevant band and modality, which we have not done.

Simulation confirms the expected profile above, and fixes its endpoints (Figure 5B). The associator declines monotonically in β: zero increases across 39 successive steps spanning $10^{-3} \leq \beta \leq 10^{2.3}$. It reaches zero in the tropical limit as stated. Its high-temperature endpoint, however, is not the value a two-input calculation suggests. Because the inner gate carries forward the constant $-S_2(\lambda^\star)/\beta$, that offset enters the outer gate's linear term multiplied by β and therefore survives the $\beta \to 0$ limit: $\lambda^\star$ converges not to ½ but to the root of $S_2{}'(\lambda) = -S_2(\lambda^\star_{inner})$, which for the two-terminal inner gate of this associator is $S_2(½) = log\ 2$, giving $\lambda^\star \approx 0.5894$. Direct computation then gives

$$\|(f \oplus g) \oplus h - f \oplus (g \oplus h)\| \to |\kappa|\,\|h - f\|, \quad \kappa = 1 - \frac{3}{2}\lambda^\star \approx 0.1159$$

which simulation matches to within 0.1%. Two features of the limit are worth recording. The plateau is $\frac{\|h - f\|}{8.6}$, not $\frac{\|h - f\|}{4}$. And it depends only on the two outer terminals: across 400 triples the high-temperature associator correlates at $r = 1.00$ with $\|h - f\|$ and at $r = 0.04$ with $\|g - f\|$, and so the gate's non-associativity in the exploratory regime is blind to the terminal that the bracketing re-attaches.

The trade-off between structure preservation and commitment has to be measured on the gate we recommend below rather than on the gate as originally stated, because the entropy offset of the original gate separates the two bracketings on its own and diverges as $\beta \to 0$, so a temperature sweep on that gate is confounded by the offset throughout and not only at $\beta = 1$. On the offset-corrected gate, on the same height-matched three-terminal rotation at 0 dB, bracketing decodes at 0.997 at $\beta = 0.1$, 0.819 at $\beta = 1$, 0.532 at $\beta = 3$, and 0.505 at $\beta = 10$, crossing the significance threshold between $\beta = 1$ and $\beta = 3$. Structure preservation and derivational commitment therefore pull against each other over roughly one decade of $\beta$, and the operating temperature is a window rather than a plateau. Manipulations that push the system toward commitment should degrade bracketing discriminability steeply rather than gradually, and the degradation should be measurable well before the gate reaches its tropical limit.

The $(\alpha - 1)$ scaling is confirmed: a log-log fit of the associator against $\alpha - 1$ over $[10^{-4}, 10^{-1}]$ at $\beta = 1$ returns a slope of 0.994 (Figure 5B). The associative limit at $\alpha \to 1$ follows from the gate becoming a log-sum-exp there, and simulation recovers it exactly.

These simulations also expose a problem in the construction as we have stated it. Each application of the gate subtracts the constant $-S_2(\lambda^\star)/\beta$, and a daughter's own offset re-enters its mother's linear term discounted by that daughter's mixture weight. Write the encoded functions as

$$\Psi(A) = \overline{\Psi}(A) - \frac{c_A}{\beta}, \qquad \Psi(B) = \overline{\Psi}(B) - \frac{c_B}{\beta},$$

where $\overline{\Psi}$ denotes the offset-free mixture and $c_A, c_B \geq 0$ are the accumulated entropy offsets; for an atomic input $\alpha$, set $c_\alpha = 0$. The offset inherited by the mother then obeys

$$c_{\mathcal{M}(A,B)} \;=\; S_2(\lambda\star) \;+\; \lambda\star\; c_A \;+\; (1 \;-\; \lambda\star)\; c_B.$$

How the offset grows in depth is therefore a function of tree shape. On a balanced tree the weights at each level sum to one and the offset grows linearly, $\approx dS_2(\lambda\star)/\beta$. On the right-branching spine used in the contrasts reported here, only one daughter of each node is itself composed, so the offset is a geometric series in the discount weight and converges to a bounded value ($\approx 2S_2(\lambda\star)/\beta$ for $\lambda\star$ near ½) rather than growing without bound in d. What holds of every shape is that the offset is depth- and shape-dependent and diverges as $\beta \to 0$. Because the two bracketings carry different constant offsets, a classifier can separate them from that offset alone, so the unnormalised gate's 0.637 overstates its structural decoding. But this does not affect the comparison – the offset-corrected gate (0.819) has no such constant and wins regardless.

The gate is therefore not scale-free under nesting, and the accumulated offset is what drives $\lambda^\star$ away from ½ in the limit computed above. A natural repair here is to redefine the embedding so that it carries forward only the entropy-optimized mixture,

$$\Psi(\mathcal{M}(A,B)) \;=\; \lambda^\star\, \Psi(A) \;+\; (1 \;-\; \lambda^\star)\, \Psi(B),$$

with $\lambda^\star$ the argmin of the full free-energy objective. This gate is commutative, non-associative and scale-preserving, and it is empirically the better structural code: it decodes bracketing at 0.819 against 0.637 at $d = 1$ and 0 dB, and holds above threshold down to −10 dB where the unnormalised gate fails by −5 dB (Figure 5C, panel 1-2). It is not, however, associative at the Shannon point (associator 1.58 at $\beta = 1$, against 0.00 for the unnormalised gate), and its $\beta \to 0$ associator is exactly $\frac{\|h - f\|}{4}$, $\lambda^\star \to$ ½ being restored.

Importantly, the corrected gate is not the only law occupying this corner. Fixed mean-pooling, ½($u$ + $v$), is commutative, non-associative and order-blind, and on the three-terminal rotation it decodes bracketing at 0.997 at $d$ = 1 – above the corrected gate at $\beta$ = 1, since the gate at $\beta \to 0$ is mean-pooling and the gate at finite $\beta$ trades part of that separation for commitment. Bracketing decodable and daughter order at chance is therefore not diagnostic of an entropy gate, and is satisfied by the most ordinary pooling operation in machine learning. What mean-pooling lacks is a content-dependent mixture, and the contrast that exposes this is one on which a fixed mixture cannot move: the balanced four-leaf pair (($A$ $B$)($C$ $D$)) versus (($A$ $C$)($B$ $D$)), matched on terminals, on the leaf-depth multiset and on the content-to-depth map. Under mean-pooling both reduce to ¼($A$ + $B$ + $C$ + $D$) and the pair is indistinguishable at any signal-to-noise ratio, and the additive law likewise. The corrected gate separates them at intermediate temperature – 0.827 at $\beta$ = 0.5 and 0.753 at $\beta$ = 1 at 0 dB, against 0.497 for mean-pooling – and not at either limit, since $\lambda\star \to$ ½ as $\beta \to 0$ restores mean-pooling and the tropical limit is associative. The role-filler and recurrent laws separate the pair as well (1.000), and are again excluded on the order axis.

The discriminating signature is therefore a *conjunction* of three: bracketing decodable on the rotation, bracketing decodable on the balanced four-leaf pair, and daughter order at chance.

Of the laws introduced so far, only the entropy gate satisfies all three, and only within a window of β. It is not the only law that can. A saturating rate unit into which both daughters enter through the same synapses, $\tanh(gW(u+v)/2)$, is commutative because the weights are shared, non-associative because its ½-weighting is (as mean-pooling already is), bounded by saturation, and not idempotent; it decodes the rotation at 1.000, 0.998, 0.976, 0.832 and 0.654 across $d$ = 1…5, the four-leaf pair at 0.966, and daughter order at chance throughout. The recurrent law we simulated fails the order axis because its two weight matrices differ.

The reason it beats the gate on the four-leaf pair is structural. Every pointwise, commutative, translation-equivariant gate has the form $g(x, y) = (x + y)/2 + h(|x - y|)$, and both entropy gates are members, differing only in $h(0)$ ($-\log 2/\beta$ originally, 0 after correction). A balanced diagonal tree of depth $d$ then evaluates to $A + d \cdot h(0)$, so the class is bounded in depth exactly when it is idempotent; and for $h(0) = 0$ the tree function agrees to second order in leaf amplitude on all trees with the same content-to-depth map, so the four-leaf pair is separated only at fourth order (the separation scales as $\varepsilon^4$ for the second-Rényi and the Shannon corrected gates alike). The narrow window is the class rather than the parameter. A saturating law is not equivariant, recovers pairing at third order through the cubic term of its transfer, and also escapes. What it lacks is the gate's sensitivity to daughter disagreement, since a sum-only law $\varphi(u + v)$ cannot see $|u - v|$.

The two escapes combine, and the combination is the operation that we propose here. We call the proposed neural binding operation *Meld*: a commutative, non-associative, content-dependent operation that converts two currently active constituent states into a single reusable composite:

$$Meld(u, v) = \tanh(\kappa W[\lambda \star u + (1 - \lambda \star)v])$$

Meld takes the corrected gate's entropy-optimized mixture – a pointwise soft minimum that rewards agreement between the daughters and penalizes disagreement – and passes it through shared orthogonal weights and a saturating transfer. The daughters are not concatenated and are not assigned fixed roles; they are combined through a mixture that depends on their content into a composite that can itself be a daughter. Meld is commutative, non-associative, bounded, not idempotent, content-dependent in its mixture and disagreement-sensitive by construction.

On the depth ladder, Meld decodes bracketing at 1.000, 1.000, 0.998, 0.970 and 0.922, against 0.972 at $d$ = 5 for the ordered role-filler law and 0.799 for the recurrent law, with daughter order at chance at every depth and the four-leaf pair at 1.000 ($d'$ = 9.1). It has no temperature window: bracketing on the rotation is at 1.000 and the four-leaf pair above 0.99 for every $\beta$ from 0.03 to 30, because as $\beta \to 0$ Meld becomes the tied-weight unit and as $\beta \to \infty$ a saturated tied minimum, both of which carry structure. Neurally it is a *circuit* rather than a formula: two constituent populations converge on a common target through the same synapses, integrate sublinearly (min-like, so that the target responds where both agree) and saturate. Among the laws tested, only Meld and its unsaturated form (law 8) satisfy every condition that the NAP (with its Neural Admissibility Criterion) has produced; both decode structure to within a few points of the inadmissible role-filler law while representing nothing that Merge lacks, and we propose Meld because a cortical target population saturates. In brief, Merge specifies what the computation must accomplish, and Meld is one candidate for how its binding step $B$ (§4.7) could be physically realized (Figure 6).

Meld's three components are not on a par. Two are forced by the algebra: without the soft minimum the law is sum-only, $\varphi(u+v)$, and blind to disagreement; with untied weights, $W_u \neq W_v$, it is the recurrent law and leaks order, and tying them is what commutativity requires of a target fed through synapses. The third is not. The idempotence result above concerns pointwise translation-equivariant gates, and any non-permutation $W$ leaves that class: the linear tied-weight mixture $W[\lambda^\star u + (1-\lambda^\star)v]$ (law 8) is not idempotent, since $G(u,u) = Wu \neq u$, and separates the balanced four-leaf pair at second rather than fourth order. With $h(\delta) \approx -c\,\delta^2,\ c = \beta/8$, the mixture's departure from the mean at small disagreement $\delta$, the pairings differ at second order by (with $\circ$ the pointwise product)

$$c\,W\big\{\,W[(A-D)\circ(B-C)]\ -\ \big(W(A-D)\big)\circ\big(W(B-C)\big)\big\},$$

which vanishes only for a permutation $W$. In simulation, law 8 matches Meld on every axis: bracketing 1.000, 1.000, 0.998, 0.964, 0.899 across $d = 1 \ldots 5$, order at chance, the four-leaf pair at 1.000 ($d' = 8.4$). Removing $W$ instead, $\tanh(\kappa[\lambda^\star u + (1-\lambda^\star)v])$, also leaks no order, but decodes the rotation at 0.843, 0.754, 0.705, 0.677, 0.684 and the four-leaf pair at 0.862 ($d' = 2.6$). The saturating transfer therefore contributes only what a rate transfer contributes – boundedness at any output gain, which a linear mixture lacks – and is retained because cortical targets saturate, not because the algebra selects it. Meld's algebraic content lies in the tied synapses and the soft minimum.

The saturating transfer is the standard model of a neuron's rate response – near-linear for small input, flattening for large – and bounds the composite however deeply structures are nested, at any output gain. Pairing is carried before it is reached: the content-dependent mixture makes the composite depend on how the daughters were paired, and the shared weights lift that dependence out of the pointwise class, which is what separates ((A B)(C D)) from ((A C)(B D)). The mixture supplies sensitivity to the daughters' relation; tying the weights removes daughter order and lifts pairing from fourth to second order; the transfer keeps recursion bounded at any gain. We stress that the transfer need not be tanh: any bounded nonlinearity would be appropriate, and divisive normalization (Carandini & Heeger, 2012) in its place performs equivalently on every contrast and slightly better at depth, since it preserves rather than compresses signal energy; we report tanh as the reference implementation. What comes out is the same kind of object that went in, so the composite can be a daughter in the next application.

Several relevant precursors can help sharpen what is novel about Meld. Tensor-product and holographic reduced representations established that distributed neural-style representations can support recursively constructed bindings (Plate, 2003; Smolensky, 1990), while vector-derived transformation binding (VTB) was developed specifically for neurally implementable, deep symbol-like processing and is itself non-associative, but also non-commutative (Gosmann & Eliasmith, 2019); for further technical discussion, see supplementary material in Murphy (Murphy, 2025). The Assembly Calculus comes still closer to the present goal: its Merge operation creates a new neuronal assembly from existing assemblies and was proposed as a substrate for language and other higher cognitive functions (Papadimitriou et al., 2020), while recent hippocampal-entorhinal models use vector binding to obtain compositional cognitive maps (Kymn et al., 2024). At a finer physiological scale, sublinear dendritic integration has now been shown experimentally and computationally to support feature-binding computations (Cazé et al., 2024; Tang et al., 2025), while divisive normalization provides a well-established nonlinear integration motif with the bounded response properties required here (Ohshiro et al., 2011). Reimann further demonstrates that non-associativity itself can be computationally useful for cognition, there as a means of retaining temporal structure in memory (Reimann, 2025). These precedents therefore make Meld more, rather than less, neurally plausible: essentially every one of its ingredients has an independent neural or cognitive precedent. The move we make here is to explore their conjunction under an independently specified linguistic admissibility constraint. Meld is not merely recursively closed or non-associative; it is a bounded, same-space, content-dependent operation that is simultaneously non-associative and exchange-symmetric, so that it preserves grouping while refusing to preserve daughter order, while its composite remains sensitive to daughter disagreement at fixed summed drive. To our knowledge, no previous neurally interpretable binding law has been derived to satisfy this full conjunction of requirements.

Meld therefore does not introduce a novel physiological primitive; rather, its novelty lies in showing that familiar neural primitives, when jointly constrained by the algebra of the computation they must realize, converge on a previously unarticulated binding law. Figure 6 summarizes Meld as a circuit, places it among the four factors of Merge, sets it against every other composition law tested, and shows its behavior across temperature. The remainder of this section and §7.2 return to each panel in turn.

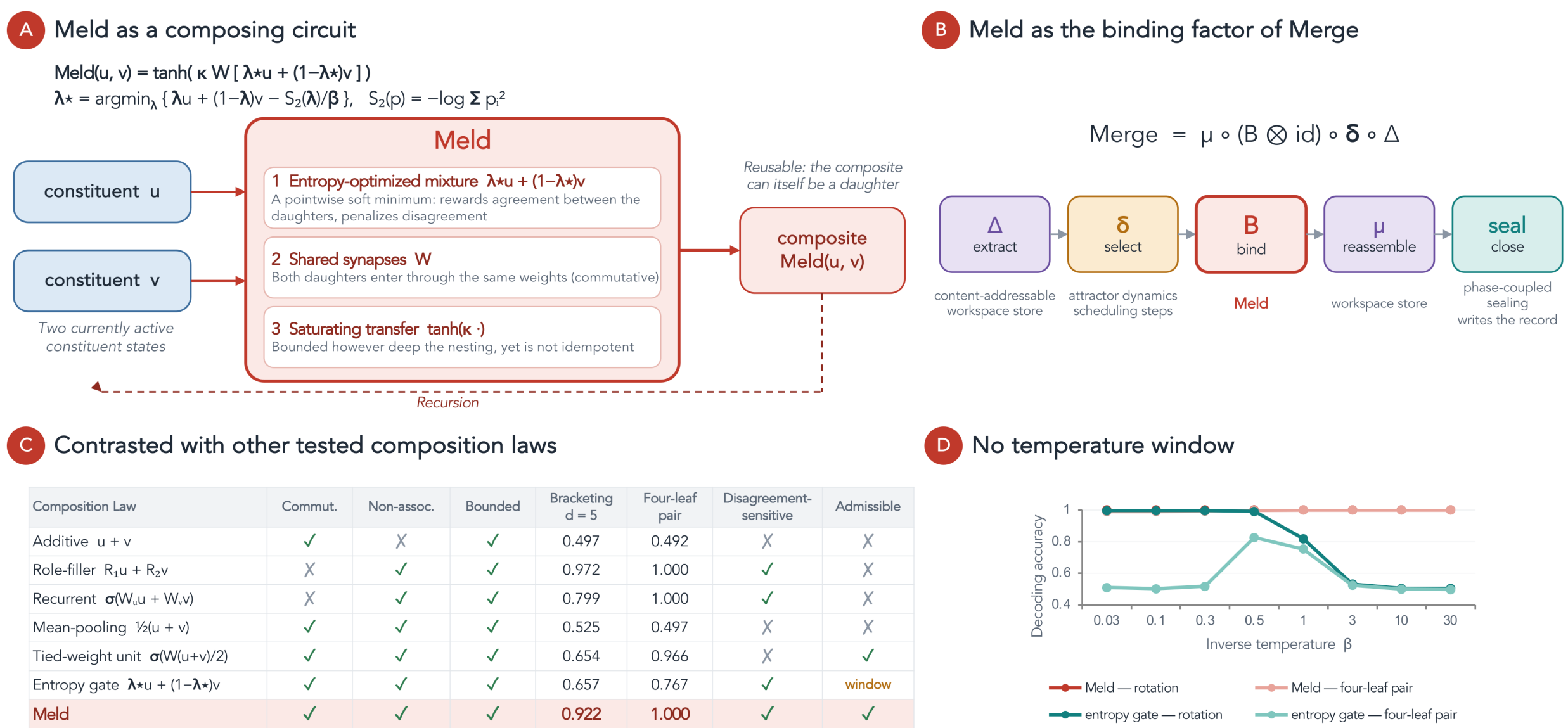

| Composition Law | Commut. | Non-assoc. | Bounded | Bracketing d = 5 | Four-leaf pair | Disagreement-sensitive | Admissible |
|---|---|---|---|---|---|---|---|
| Additive $u + v$ | ✓ | ✗ | ✓ | 0.497 | 0.492 | ✗ | ✗ |
| Role-filler $R_1u + R_2v$ | ✗ | ✓ | ✓ | 0.972 | 1.000 | ✓ | ✗ |
| Recurrent $\sigma(W_u u + W_v v)$ | ✗ | ✓ | ✓ | 0.799 | 1.000 | ✓ | ✗ |
| Mean-pooling $\frac{1}{2}(u + v)$ | ✓ | ✓ | ✓ | 0.525 | 0.497 | ✗ | ✗ |
| Tied-weight unit $\sigma(W(u+v)/2)$ | ✓ | ✓ | ✓ | 0.654 | 0.966 | ✗ | ✓ |
| Entropy gate $\lambda_\star u + (1-\lambda_\star)v$ | ✓ | ✓ | ✓ | 0.657 | 0.767 | ✓ | window |
| Meld | ✓ | ✓ | ✓ | 0.922 | 1.000 | ✓ | ✓ |

**Figure 6. Meld, the proposed neural binding operation.** (A) Merge is the formal operation at the center of theoretical syntax; Meld is the proposed neural realization of its binding step. Two currently active constituent states enter a common target population. The target forms the entropy-optimized mixture $\lambda_\star u + (1 - \lambda_\star)v$ of §4.9.1 – a pointwise soft minimum that rewards agreement between the daughters and penalizes disagreement – through shared synapses $W$, which is what makes the operation commutative, and passes it through a saturating transfer, which bounds it however deeply structures are nested without making it idempotent. The composite is reusable and can itself be a daughter (dashed loop). (B) Where Meld sits within the context of Merge. Merge factorizes as $\mu \circ (B \otimes \mathrm{id}) \circ \delta \circ \Delta$ (§3.2, §4.7); the composition laws compared in this article are candidates for the binding factor $B$ alone, which Meld realizes, while extraction and reassembly are realized by the content-addressable workspace store, selection by the attractor sequencer, and closure by the phase-coupled sealing operation of §5.3. A binding law cannot be tested in isolation: its inputs are produced by Δ and its output is consumed by μ, so the order-blindness test is a statement about the composite the parser passes to the next Merge step. (C) Meld against every other composition law tested (§4.9.1): the algebraic properties each law has, its bracketing accuracy at $d = 5$ and on the balanced four-leaf pair (0 dB, 60 lexicons, nearest-centroid), whether its composite depends on daughter disagreement at fixed daughter sum, and whether it is admissible. The ordered laws decode well because they keep daughter order, which Merge discards; Meld is within a few points of them while representing nothing it lacks. (D) Meld has no temperature window. The entropy mixture on its own preserves grouping only within a band of commitment, collapsing to fixed averaging as $\beta \to 0$ and to an associative minimum as $\beta \to \infty$; Meld's rotation and four-leaf decoding are at or above 0.99 for every $\beta$ from 0.03 to 30, because it degrades to the tied-weight saturating unit at one end and to a saturated tied minimum at the other, both of which carry structure.

On neural data, the four-leaf pair is not available as a matched-string contrast, since $((A\ C)(B\ D))$ has no linearization as the string $A\ B\ C\ D$. What discriminates the gate from mean-pooling there is the readout we will return to in §9.2. Under mean-pooling the leaf mixture is fixed by tree shape alone, $a_\ell = 2^{-\mathrm{depth}(\ell)}$, so any two bracketings with the same leaf-depth multiset – including every contrast used above – carry identical $S_2$ and identical effective dimensionality. Under the gate the mixture depends on the daughters' values, and the two bracketings separate in $D_{\mathrm{eff}}$. The effective-dimensionality prediction we will discuss in §9.2 is therefore not a corollary of the composition comparison but the part of it that a content-independent mixture cannot satisfy. It does not separate the entropy mixture from a saturating sum-only law, whose nonlinearity also makes the leaf mixture bracketing-dependent; that separation is sensitivity of the composite to daughter disagreement at fixed daughter sum, which Meld has and a sum-only unit lacks.

We stress that the corrected gate does not inherit the generic-injectivity theorem established for the original embedding (Marcolli & Berwick, 2026), and it fails injectivity in two identifiable places. First, as $\beta \to 0$ it reduces to mean-pooling and the balanced four-leaf pair above collapses, so faithfulness fails on ordinary four-leaf trees in exactly the limit that §4.9 identified as most structure-preserving. Separation of that pair is restored only by the content-dependence of $\lambda^\star$ at finite $\beta$, and is not monotone in $\beta$. Second, the corrected gate is idempotent, $G(u, u) = u$, so $\Psi(\mathcal{M}(A, A)) = \Psi(A)$ and every object built by repeated self-merge collapses onto its leaf. The gate as originally stated is not idempotent, and it is not so precisely because of the offset, $x \oplus x = x - \log 2/\beta$: the term that makes it unbounded is also what keeps it faithful on the diagonal. Offset-correction therefore yields boundedness at the price of faithfulness, at $\beta \to 0$ and on the diagonal. A consequence of this is that a nonzero associator is necessary for faithfulness but not sufficient – the corrected gate has a nonzero associator at every order and still collapses $\mathcal{M}(A, A)$ – so the associator statistics reported above index non-associativity and nothing more. And such faithfulness as the corrected gate has, it has at intermediate temperature and off the diagonal. We do not claim it generically, and the separations the bracketing simulations exhibit are what they exhibit on the contrasts tested.

The two claims made in this section therefore hold of different gates. On the offset-corrected gate the requirement $\alpha \neq 1$ does not follow, and the Shannon point is not the associative limit. Our recommendation is to adopt the offset-corrected gate and drop the $\alpha \neq 1$ argument, since a construction whose output carries a tree-shape-dependent offset that diverges as $\beta \to 0$ is not a candidate neural code, and $\alpha$ is in any case an empirically fitted parameter (by §9.2, below) rather than a formal requirement.

One critical implication of this section is that the gate as originally stated by Marcolli and Berwick seems not to be a candidate neural code: every application of the gate leaves behind a constant, and the constant depends on how deep and what shape the tree is, and it grows without bound in exactly the high-temperature regime where structure is best preserved. No bounded, noisy population could feasibly carry a quantity with that behavior. The repaired gate (strictly speaking, the offset-normalized candidate gate) carries forward only the entropy-optimized mixture and drops the accumulated offset, which is scale-free under nesting and a better structural code by a wide margin, and is hence what any neural realization should be tested against.

Two kinds of non-compliance have been reported here and they have different standing. The first is finitude: the competence object is infinite, and any bounded realization is faithful only on a bounded region – the $\beta$-window narrowing with leaf count, the depth bound on the phase code, the workspace bound. These are the content of the ‘stated equivalence’ in the Neural Admissibility Criterion of §1. The second is structural, and the corrected gate has two instances. Idempotence collapses $\mathcal{M}(A, A)$, but no licit derivation merges an object with an identical sister: copies arise through Internal Merge and the coproduct, yielding {*X*, {…*X*…}}, so faithfulness is claimed on the off-diagonal magma. The four-leaf collapse is a property of the $\beta \to 0$ limit rather than of any operating point. Inside the window the pair separates, and what is missing is a proof of generic injectivity at finite $\beta$. The claim that survives is that Meld is a candidate neural realization of the binding factor *B* of Merge, admissible on the derivations the grammar generates at every temperature tested and within a stated depth bound. The corrected entropy mixture it contains is admissible on its own only within a temperature window. Within the pointwise translation-equivariant class no gate can perform better on the diagonal, since boundedness there forces idempotence, Meld is bounded without being idempotent, and pays in contraction with depth, which is finitude of the first kind.

The corrected gate should therefore be read as a candidate realization of the local binding factor $B$, not as a stand-alone faithful encoding of the entire workspace-level Merge operator. Full admissibility remains a property of the factorized system.

### *4.10. Phase-synchronization instantiation*

Following directly from the above discussion, the function-space construction acquires another concrete neural reading when the atoms are taken to be oscillatory (Murphy, 2020b). Suppose $\iota : \mathcal{SO}_0 \to C(X, \mathbb{R})$ encodes each lexical atom as a simple sinusoid, or a linear superposition of them

$$\iota_{\alpha_\ell}(t) = A_{\alpha_\ell} \sin\left(\nu_{\alpha_\ell} t + \omega_{\alpha_\ell}\right).$$

Here $A_{\alpha_\ell}$ is the carrier amplitude, $\nu_{\alpha_\ell}$ its angular frequency, $\omega_{\alpha_\ell}$ its lifted phase, and $t$ is time. We use $\mathfrak{M}$ for the proposed synchronization realization of the abstract Merge operation $\mathcal{M}$. The proposal is then to represent the magma operation $\mathfrak{M}$ by cross-frequency phase synchronization, as Marcolli and Berwick discuss (Marcolli & Berwick, 2026). This requires a commensurability condition. Two carriers synchronize in an $n{:}m$ regime when their frequencies stand in a rational relation. Specifically, there exist positive integers $n_{(\alpha_1,\alpha_2)}$ and $n_{(\alpha_2,\alpha_1)}$ that are coprime,

$$\gcd\left(n_{(\alpha_1,\alpha_2)}, n_{(\alpha_2,\alpha_1)}\right) = 1,$$

and satisfy the commensurability condition

$$n_{(\alpha_1,\alpha_2)}\, \nu_{\alpha_1} \;=\; n_{(\alpha_2,\alpha_1)}\, \nu_{\alpha_2},$$

in which case the relevant phase difference is not $\omega_{\alpha_1} - \omega_{\alpha_2}$ but the integer-scaled quantity

$$\Delta\omega_{\alpha_1\alpha_2} \;:=\; n_{(\alpha_1,\alpha_2)}\, \omega_{\alpha_1} - n_{(\alpha_2,\alpha_1)}\, \omega_{\alpha_2}.$$

Merge is then realized by driving this quantity to a common value. The composed object retains both daughters' amplitudes and carrier frequencies and assigns them a single shared phase,

$$\mathfrak{M}\left(\iota_{\alpha_1}, \iota_{\alpha_2}\right) \;=\; A_{\alpha_1} \sin\left(\nu_{\alpha_1} t + \omega_{\{\alpha_1,\alpha_2\}}\right) + A_{\alpha_2} \sin\left(\nu_{\alpha_2} t + \omega_{\{\alpha_1,\alpha_2\}}\right),$$

and the shared phase is fixed by the gate of §4.9 applied to the scaled daughter phases,

$$\begin{aligned} \omega_{\{\alpha_1,\alpha_2\}} \;&:=\; n_{(\alpha_1,\alpha_2)}\, \omega_{\alpha_1} \;\oplus_{\mathrm{S}_2,\beta}\; n_{(\alpha_2,\alpha_1)}\, \omega_{\alpha_2} \\ &= \min_{\lambda \in [0,1]} \left\{ \lambda\, n_{(\alpha_1,\alpha_2)} \omega_{\alpha_1} + (1-\lambda)\, n_{(\alpha_2,\alpha_1)} \omega_{\alpha_2} - \beta^{-1} \mathrm{S}_2(\lambda) \right\}. \end{aligned}$$

Iterating over a tree gives the synchronization model $\Psi_{\mathrm{S}_2,\beta} : \mathcal{SO} \to C(X, \mathbb{R})$,

$$\Psi_{\mathrm{S}_2,\beta}(T) \;=\; \sum_{\ell} A_{\alpha_\ell} \sin\left(\nu_{\alpha_\ell} t + \omega_T(\alpha_1, \ldots, \alpha_n)\right)$$

$$\omega_T(\alpha_1, \dots, \alpha_n) = \min_{w \in \Delta_n} \left\{ \sum_{\ell=1}^{n} w_\ell \, n_\ell \omega_{\alpha_\ell} - \beta^{-1} S_T(w) \right\}.$$

Here $w = (w_\ell)_{\ell=1}^{n} \in \Delta_n$ is the leaf-weight vector, and $n_\ell$ is the positive-integer frequency multiplier assigned to leaf $\ell$ by the relevant commensurability relation, which is commutative and non-associative by the results of §4.9, and for which $T \mapsto \Psi_{S_2,\beta}(T)$ is generically injective. The construction extends in the same manner to amplitude-phase couplings and other synchronization phenomena.

Three observations follow, with direct implications for the algebraically compliant code we develop in §5.

First, the observable is $n{:}m$, not $1{:}1$. The commensurability condition is not a technicality of the formalism – it directly fixes what should be measured. The relevant phase-locking statistic for a construction of this kind is the $n{:}m$ phase-locking value,

$$\mathrm{PLV}_{ij}^{n:m} \;=\; \left| \frac{1}{N} \sum_{k=1}^{N} \exp\left(i[\, n\, \phi_i(t_k) - m\, \phi_j(t_k)\,]\right) \right|,$$

with $n, m$ set by the measured carrier ratio and $\gcd(n, m) = 1$, rather than the $1{:}1$ statistic that is standardly reported. Here, $i$ and $j$ index the two carriers or recording sites, $N$ is the number of observations, $t_k$ is the $k$th observation time, and $\phi_i(t_k)$ and $\phi_j(t_k)$ are the corresponding instantaneous phases. A $1{:}1$ estimate is blind to exactly the regime in which the construction lives. Testing this account therefore requires estimating the carrier ratio per site per trial and computing the locking statistic at that ratio, and a null $1{:}1$ result is not evidence against the model. Note also that the construction is stated over lifted phases: $\omega$ enters the objective linearly and is scaled by integers, so the minimization is over the reals and not over the circle. Any realization must therefore unwrap phase, which is a substantive commitment and is in some tension with the wrap-around bound we will derive in §5.1.

Second, the offset problem of §4.9.1 applies here. $\omega_T$ is defined as the *minimum value* of the free-energy objective, and therefore carries the accumulated constant $-\beta^{-1} S_T(w^\star)$ that we show in §4.9.1 to be tree-shape dependent and to diverge as $\beta \to 0$. In the generic function-space setting this is an implausibility, but here it is a genuine incoherence: the encoded quantity is a phase, whose natural range is bounded, and a phase that drifts without bound as temperature rises is not a phase. The repair we recommend in §4.9.1 transfers directly and is more clearly forced in this instantiation than in the one that motivated it. Carrying forward only the entropy-optimized mixture gives

$$\omega_T(\alpha_1, \dots, \alpha_n) = \sum_{\ell=1}^{n} w_\ell^\star \, n_\ell \omega_{\alpha_\ell}, \qquad w^\star = \operatorname*{arg\,min}_{w \in \Delta_n} \left\{ \sum_{\ell=1}^{n} w_\ell \, n_\ell \omega_{\alpha_\ell} - \beta^{-1} S_T(w) \right\},$$

which is a convex combination of the scaled leaf phases and therefore lies in their convex hull at every temperature, while remaining commutative and non-associative.

Third, this model will soon become a rival to the phase-address code we develop in §5.1. The two proposals make opposite predictions about phase dispersion, and this is worth stating before we progress to the next section because they are easily conflated under the common (and rather vague) heading of 'phase codes for language'. In the synchronization model, composition *is* phase convergence: all leaf carriers of a completed object share a single phase $\omega_T$, and the entire tree is encoded in the value of that one scalar. In the phase-address code we develop in §5.1, composition writes depth into phase *offset*, $\Phi(d) = d \cdot \Delta\varphi$, so that a completed object of depth $d$ occupies a spread of phases and grouping is recovered from the joint code $\langle \Phi(d), \Gamma \rangle$ binding content to address. The discriminating measurement is the circular dispersion of constituent-carrier phases as a function of composition: the synchronization model predicts that it collapses at the point of constituent completion, and the address code predicts that it grows with the number of occupied depths.

We regard the address code as the more plausible of the two as a likely neural realization, for reasons that also bear on §4.9.1. Generic injectivity is not *robustness*: the synchronization model places an entire tree into one real number, so that arbitrarily small perturbations of $\omega_T$ map to distinct trees and there is no redundancy from which a bounded, noisy population could recover the intended one. In contrast, the address code is discrete and redundant, which is why it degrades by aliasing at a bound fixed by the carrier ratio (§6.5.1) rather than by unbounded confusion. This is naturally not an argument against the formal result, which stands, but against reading it directly as a neural hypothesis – it is effectively the same distinction between competence-level faithfulness and bounded realization that we drew above in §4.9, rendered concrete and quantifiable.

The neurophysiological observables available here are otherwise standard (Bruña et al., 2018; Canolty et al., 2010; Safavi et al., 2023; Sauseng & Klimesch, 2008; Vissani et al., 2025), but are now more constrained. For phase-amplitude coupling, we can test whether the amplitude of a high-frequency process associated with lexical or feature content is modulated by a low-frequency phase associated with workspace or grouping state (Murphy, 2025). The formal prediction is not that syntax simply increases PAC over the course of a sentence. Rather, it is that PAC or phase synchronization should be selectively organized by non-associative grouping, interface filter compatibility, and workspace manipulation/extraction. A generic sentence > wordlist PAC effect is therefore insufficient – indeed, it would be expected under many non-compositional, non-hierarchical accounts of language. We refer the reader to prior neurocomputational models (Ding, 2023; Martin, 2020; Meyer, 2018; Murphy, 2015b, 2020b, 2025) and empirical work (Beese et al., 2017; Brennan & Martin, 2020; Coopmans et al., 2022, 2023; Mai et al., 2016; Weissbart & Martin, 2024) for help in narrowing down specific frequency bands and cortical regions of interest.

Due to the centrality of phase dynamics not just in theoretical neurolinguistic models but also in experimental reports of language (Goswami et al., 2014; Keshavarzi et al., 2024; Ten Oever et al., 2024), the next section will focus on this topic and will be dedicated to making explicit how a neural substrate can *physically enforce* the two defining axioms of the syntactic magma – non-associativity and commutativity – converting the resulting picture into a graded, ranked set of falsifiable predictions.

## 5. Phase-Address and Coupling-Based Composition

The algebraic results developed above place non-negotiable demands on any neural code for Merge. The free magma generating syntactic objects is *non-associative*: the bracketings $(\ell_i \oplus \ell_j) \oplus \ell_k$ and $\ell_i \oplus (\ell_j \oplus \ell_k)$ are distinct objects, so their physical realizations must remain distinguishable even when the linear string of terminals is held fixed. In addition, the magma is *commutative*: $\ell_i \oplus \ell_j = \ell_j \oplus \ell_i$, so the physical realization must be invariant to the surface order in which the two daughters are delivered to the workspace. These demands are in tension. A code that uses temporal order, or phase position, as a serial-order marker will arguably satisfy commutativity only by destroying the bracketing information that non-associativity requires (or vice versa). In addition, commutativity seems liable to be in tension with real-time processing of linguistic stimuli via linear strings of speech/text. The central claim of this section is that cross-frequency coupling has the potential to dissolve some of these tensions by carrying depth in a phase-address code, within-set identity in an order-free hash-key code, and grouping in the sequence of composites those two codes write at each closure, and that these functions are physically separable, mechanistically grounded, and experimentally dissociable.

*5.1. Non-associativity as a phase-address code*

Recent work (Brennan & Martin, 2020; Dekydtspotter et al., 2025, 2026; Murphy, 2024, 2025; Qi et al., 2025) has considered the possibility that many structural elements of syntax-semantics are indexed by a slow δ-θ rhythm, with a nested β-band and $\gamma$ envelopes. While δ-θ coupling is by no means the only plausible option (other lower frequencies are patently also plausible), we choose it for convenience to demonstrate the expressive and formal power of certain phase dynamics for capturing elements of syntax. Assuming only the most generic properties of phase dynamics, one option for neural enforcement is that the bracketing depth of a constituent is encoded neither by its power nor by its latency, but by the *phase offset* at which its $\gamma$-indexed linguistic features couple to the matrix δ-θ cycle. Concretely, within a single δ-θ cycle, the β-band troughs that gate successive levels of embedding are offset by an approximately fixed phase increment per level of embedding – a monotone depth-phase map – so that a daughter merged at depth *d* acquires a $\gamma$-coupling phase $\psi(d)$ relative to the mother δ-trough:

$$\Phi(d) = d \cdot \Delta\varphi\,, \quad \Delta\varphi \approx \frac{\pi}{2}\,, \quad \psi(d) \approx \psi_0 + \Phi(d) \quad (mod\ 2\pi).$$

The two bracketings of a three-element string are then physically distinct: in $(\ell_i \oplus \ell_j) \oplus \ell_k$ the inner set occupies depth $d+1$ and the outer attachment occupies depth $d$, whereas in $\ell_i \oplus (\ell_j \oplus \ell_k)$ these depths are reversed. Note that the two bracketings share the *same* multiset of leaf depths ({d, d+1, d+1}); what differs is only which content sits at the shallow address. The disambiguating quantity is therefore not the marginal distribution of coupling phases – which is identical across the two parses, as is broadband γ power – but the *joint* code ⟨Φ(d), Γ⟩ binding each depth phase to the γ-band content Γ coupled at it (developed below in §5.2). Bracketing is recoverable from this phase-content binding, not from phase occupancy alone. Because the assignment is by depth *address* and not by ordinal position, the code is intrinsically non-associative without being order-sensitive. This is the oscillatory correlate of the magma's generation of non-planar binary trees: the phase coordinate is the neural realization of the Gorn-style depth address that distinguishes the two bracketed structures.

One limit of the static code must be noted here. The joint map ⟨*Φ*(*d*), *Γ*⟩ assigns each terminal its depth, and a content-to-depth function does not determine an unordered tree: ((*A B*)(*C D*)) and ((*A C*)(*B D*)) are distinct syntactic objects with the same terminals, the same leaf-depth multiset and the same content-to-depth map, so a static phase-content readout cannot separate them. For parses of a fixed linear string the deficit does not arise – two bracketings of the same string with the same depth map are the same ordered tree, hence the same syntactic object (§7.2) – and the three-terminal contrasts below are of that kind. As a code for Merge, however, the static map is not faithful, and the containment topology of §6.5, which reads dominance rather than depth, is strictly more expressive than it. What restores grouping to the phase code is time. Each sealing event of §5.3 writes a composite tag ⟨*Φ*(*d*), *Γ*⟩ whose content *Γ* is that of the sealed set rather than of a terminal, so the sequence of sealing events records which contents were composed together at each depth, and that sequence does determine the tree. The angular phase address is therefore one coordinate of a dynamic grouping code, not a complete code for grouping. Full constituency is supplied by phase address × composite identity × sealing history, and the observable for grouping is the time-resolved identity of the composite written at each seal rather than the static phase-content map at the end of the constituent. §5.1.1 is stated accordingly.

Importantly, the value of $\Delta\varphi$ need not be stipulated. The finest phase increment the code can resolve is bounded below by the duration of one $\beta$ cycle expressed as a fraction of the $\delta - \theta$ cycle, $\Delta\varphi_{min} = 2\pi \frac{f_{\delta\theta}}{f_\beta}$, since two depths whose gating troughs fall within the same $\beta$ cycle are not separable. The number of resolvable depths is therefore bounded by the frequency ratio itself:

$$d^* = \frac{2\pi}{\Delta\varphi} \leq \frac{f_\beta}{f_{\delta\theta}}.$$

For instance, if we assume $f_\beta \approx 16\,Hz$ and $f_{\delta-\theta} \approx 4\,Hz$ this yields $d^* \approx 4$, recovering $\Delta\varphi \approx \frac{\pi}{2}$ from the carrier frequencies rather than by assumption. The phase code is thus intrinsically a performance-bounded realization of an unboundedly recursive competence operation. This results in a clear prediction: the ratio of peak $\beta$ to peak $\delta - \theta$ frequency should predict the maximum embedding depth at which bracketing remains recoverable. A null result here is difficult to reconcile with an angular phase-address code, since the bound is not a free parameter of the model but a consequence of its carriers.

This also makes a sharp, and to our knowledge novel, prediction that separates a linguistic phase code from the hippocampal theta-sequence and phase-precession codes with which it is otherwise superficially congruent – where phase is canonically read as a position/order marker. In sequence memory, phase has classically been read as an *ordinal* marker; yet Liebe and colleagues report that, in both humans and recurrent networks, the phase of firing does *not* track temporal order (Liebe et al., 2025) – precisely the dissociation a commutative-but-non-associative code requires. Some material predictions follow from this.

#### *5.1.1. Prediction A: Bracketing dissociation*

For globally ambiguous strings that differ only in bracketing – with the cleanest case being coordination scope (e.g., 'You can have coffee and cake or ice cream'), where $[[A\ and\ B]\ or\ C]$ and $[A\ and\ [B\ or\ C]]$ share an identical number of binary combinations – the *joint* distribution of low-frequency coupling phase and the $\gamma$-band content coupled at each phase (i.e., which lexical/feature content occupies the deep vs. shallow depth address) should differ categorically at the disambiguating region, even though the marginal phase-occupancy distribution is matched, while broadband $\gamma$ power, lexical-surprisal-driven θ-$\gamma$ effects, and any sentence > word-list contrast remain matched. This isolates non-associativity proper. For contrasts with four or more terminals in which the content-to-depth maps coincide (§5.1), the prediction transfers from the static map to the sealing sequence: the γ-band content coupled at the deep address at successive commit bursts should identify the composed set at each seal, and the sequence of composite identities should separate the bracketings even where the terminal-level phase-content map does not. The prediction is thus stated on the three-factor code – phase address, composite identity, sealing history – and not on phase address alone.

### *5.1.2. Prediction B: Depth-address versus depth-count*

Across a parametric manipulation of embedding depth ($d = 1 \dots 4$), the β-trough phase gating each new constituent should advance approximately linearly with $d$ (slope ≈ Δφ per level), saturating near the workspace bound (e.g., ~4-5 constituents). This phase-address signal is predicted to be dissociable from the β-*power* ramp that indexes the total number of open nodes via a re-entrant cortico-striato-thalamo-cortical loop (A. G. Lewis et al., 2015, 2016; Matar et al., 2021; Murphy, Hoshi, et al., 2022; Weissbart & Martin, 2024; Zioga et al., 2024). Power tracks *how many* nodes are open; phase tracks *where* in the tree a given node sits. A design that holds open-node count constant while varying the depth address of the critical node (e.g., balanced versus right-branching trees of equal node count) should therefore dissociate a β-power count away from a β-phase address.

## *5.2. Commutativity as an order-free set code*

Once the parser has identified the two daughters of a local composition, their co-active feature bundles may need to be kept distinct without encoding their surface order. A phase-address or hash-key scheme is one plausible way to achieve this: it separates the daughters for downstream decoding while allowing head selection or labeling to be determined by coupling strength, category, or interface compatibility rather than by first-versus-second arrival.

One possible implementation is that the two daughters of a set share the slow δ-θ coupling phase that already encodes their common embedding depth (Φ(d); §5.1), and are separated instead by distinct γ-phase slots nested within that depth window. Each γ slot functions as a temporary hash key: an arbitrary address within the γ cycle that keeps co-active bundles separable for downstream decoders. The relevant symmetry-breaking operation within the local set is head selection. This can be modeled as the daughter whose γ-amplitude bundle is maximally coupled to the slow structural phase:

$$x^{\star} = \underset{x \in \{i,j\}}{\arg\max} \mathrm{MI}\left(\varphi_{\delta\theta}(t); a_{\gamma,x}(t)\right), \qquad \mathrm{Head}\left(\ell_i, \ell_j\right) = \ell_{x^{\star}}.$$

Here, $\varphi_{\delta\theta}(t)$ is the slow structural phase, $a_{\gamma,x}(t)$ is the $\gamma$-band amplitude envelope associated with daughter $\ell_x$, and $\mathrm{MI}$ is the phase-amplitude modulation index which we use in a manner compatible with the Kullback-Leibler form introduced by Tort and colleagues (Martínez-Cancino et al., 2019; Perley & Coleman, 2024; Tort et al., 2010). That $H$ is defined over an unordered pair does not by itself deliver order-independence: $argmax$ over a two-element set is order-free for any function, and the substantive question is whether the modulation index is itself order-biased. In a real-time parser it generically will be, since the first-arriving daughter has had longer to establish coupling. Order-independence therefore requires the further condition that $MI$ be estimated over a window in which both daughters are co-active and their coupling has equilibrated.

This is an empirical assumption rather than a formal consequence: head selection should be late relative to first-daughter onset, by approximately the coupling equilibration time, rather than tracking the arrival of either daughter. Subject to that condition, the implementation preserves the residue of Merge commutativity at the level of the completed local object, even though the real-time parser that constructed that object remains order-sensitive.

The category inherited by the winning bundle can then be treated as a compound address involving both a structural phase coordinate and a local population code. Schematically, the phase coordinate indexes where the object is being maintained within the active workspace, while γ-band population structure carries information about lexical features. The point is not that phrase types occupy fixed universal phase angles, but that phrase identity and structural position should remain jointly decodable from the interaction between local high-frequency activity and slower coordinating dynamics (Murphy, 2025).

### *5.3. Set commitment*

Non-associativity must not only be *written* as a depth address; it must be *protected*, so that an already-grouped set cannot be silently re-bracketed by a later operation. Recent work in the neurobiology of attention has pointed to certain frontoparietal low-frequency dynamics that serve to 'shield' ongoing representational inferences (Bonnefond & Jensen, 2025; Gutteling et al., 2022). In other research, a brief β-band 'commit' burst (Lundqvist et al., 2024) likewise offers a means to suppress the daughters' $\gamma$ carriers, and write the category vector to a read-only frontotemporal label register that subsequent Merge gates may consult but not overwrite. The ROSE model makes use of these dynamics (Murphy, 2024, 2025, 2026a; Uluslu & Murphy, 2026), and here we formalize closure as a sealing operator ᵊ:

$$^{ə}[\ell_i \oplus \ell_j] : \quad a_{\gamma,i}, a_{\gamma,j} \to 0\,, \quad \rho\left(^{ə}[\ell_i \oplus \ell_j]\right) = \langle \Phi(d), \Gamma \rangle,$$

where $\rho$ is the active-workspace readout map: after sealing, $\rho$ returns the composite's category vector and not the daughters' bundles, rendering the set an indivisible unit for the immediately following Merge step.

It is essential that sealing *not* be read as deletion, since that would contradict the workspace requirement of §4.7: the coproduct $\Delta$ needs the accessible subterm $T_v$ itself, not merely its label, and a read-only register holding only $\langle \Phi(d), \Gamma \rangle$ would leave the framework with no object to extract. Sealing therefore relocates rather than destroys. The daughters' structure is transferred from active $\gamma$ carriers into the associative store developed in §6.7 – of which the label register is a sealed instance – where it remains addressable and from which the coproduct can recover it. What sealing suppresses is live access, not access as such: the constituent is unavailable to the next gate by default, and available to $\Delta$ on demand.

This is the correlate of closing Merge – the transition from an actively composed object to a stabilized, labeled composite state that re-enters the coproduct – and is what physically freezes bracketing while leaving within-set order undefined. It also predicts an asymmetry that a purely activation-based account does not: re-opening a sealed constituent should be measurably slower and costlier than reading a still-active one, because it requires an associative-store read rather than a direct $\gamma$ readout.

The sealing operator ə also predicts a specific population-level dynamical signature. If closure corresponds to a set becoming an indivisible unit – a fixed point of the local dynamics that is robust to further input until explicitly re-opened by the coproduct – then constituent completion should coincide with a collapse of across-trial and across-time variability in the sealing population: a drop in Fano factor and in effective dimensionality (*variance quenching*) (Churchland et al., 2010) at the moment of closure, of the kind observed at commitment in other decision and motor systems. The complementary prediction follows from the coproduct: when a sealed constituent is later re-opened for extraction, agreement, or reanalysis (§4.7, §7.1), its population state should *re-expand* – a transient rise in variability and dimensionality as the unit is unpacked.

Closure and re-opening therefore predict opposite-signed excursions of population *variability*: an across-trial variance collapse (Fano factor, noise dimensionality) at set commitment, and a transient re-expansion at coproduct extraction. This noise-side dimensionality is distinct from the signal-side mixture dimensionality of what we will introduce in §9.2 – which tracks how many terminals a constituent mixes ($\log PR$ of the leaf-mixture), not how variable the state is at fixed input – and the two can move oppositely, since a constituent may mix more terminals even as its representation settles onto a more stable fixed point.

Figure 7 summarizes much of what this section has proposed. Each of the predictions in this section follows from taking the formal properties of language as a constraint on the physics of the neural code for language, rather than from the weaker premise that composition is indexed by some general low-frequency modulation.

Phase dynamics enforce the algebra of Merge

*Cross-frequency coupling carries bracketing and within-set identity on two physically separable codes*

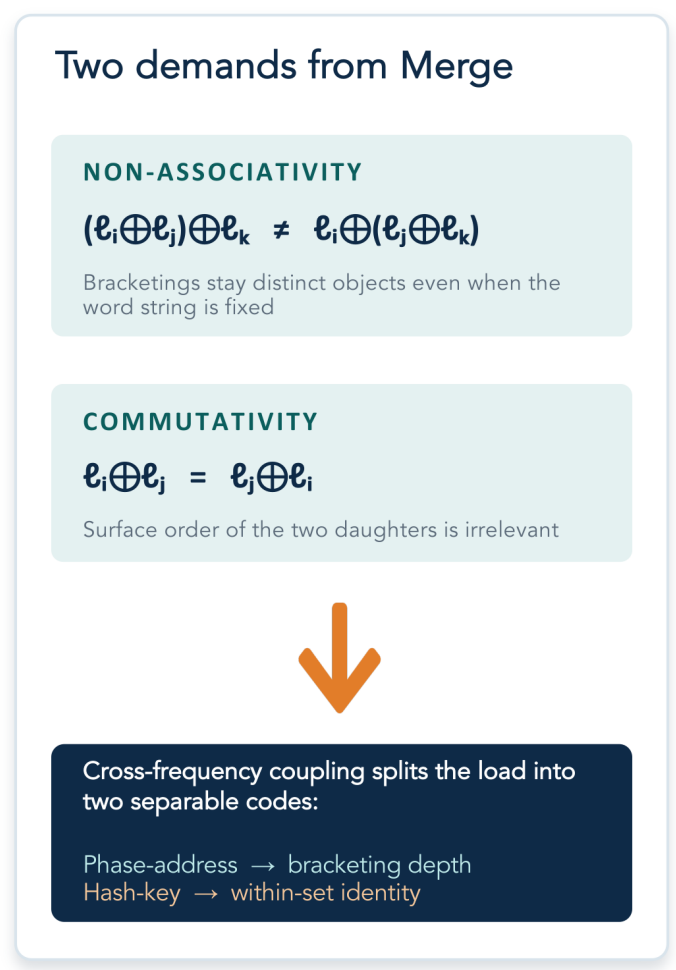

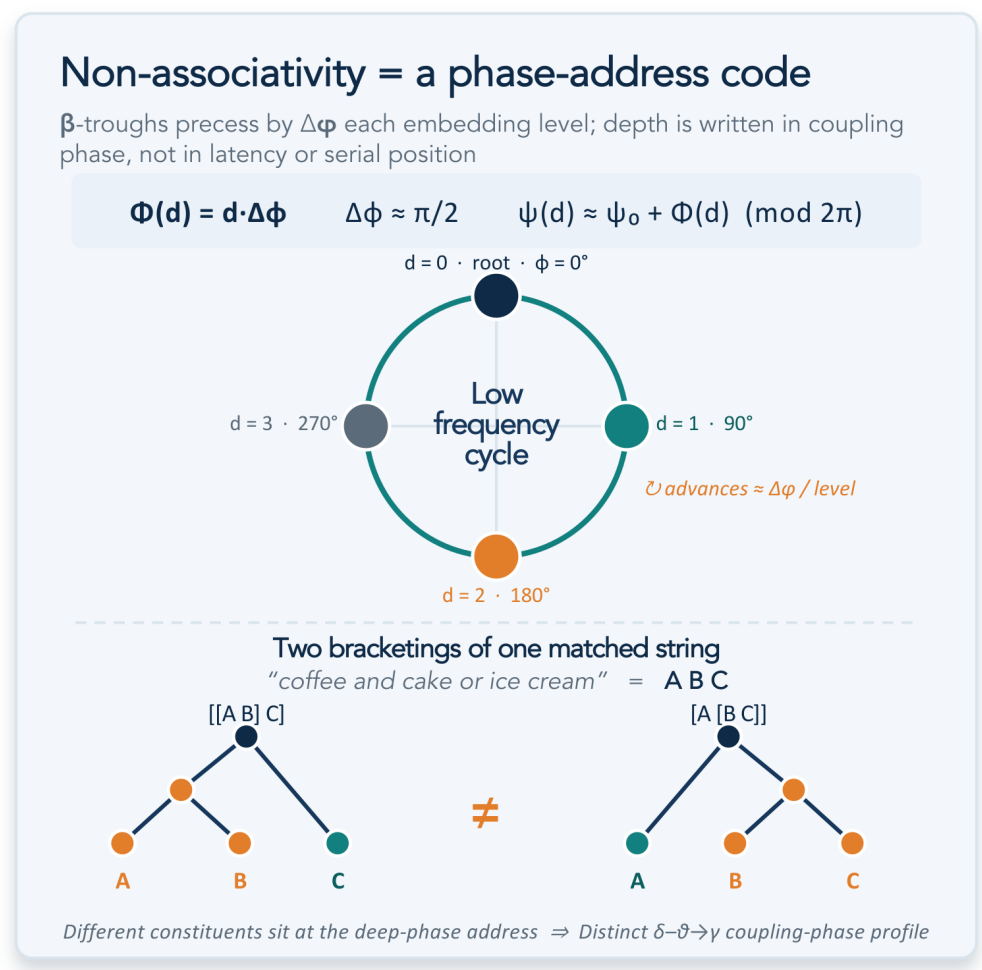

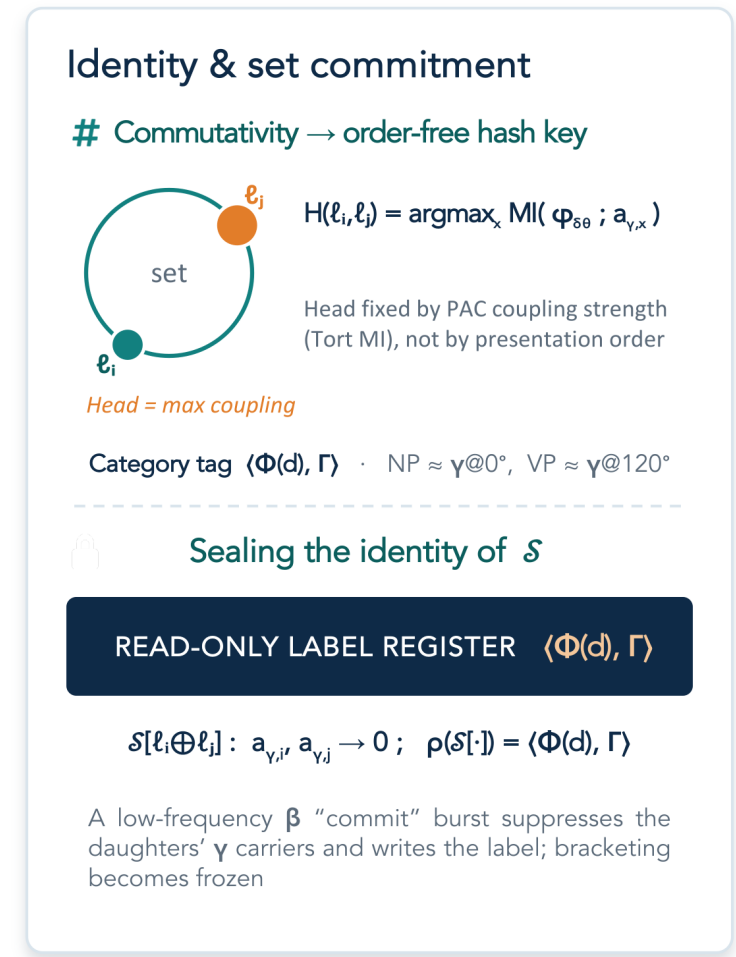

**Figure 7. Cross-frequency coupling enforces the algebraic structure of Merge through two physically separable codes.** The free magma of syntactic objects imposes two demands on any neural code: non-associativity, so that $(\ell_i \oplus \ell_j) \oplus \ell_k$ and $\ell_i \oplus (\ell_j \oplus \ell_k)$ remain distinct even at fixed terminal order, and commutativity, so that the realization is invariant to daughter order (*Left*). These are in tension for any serial-order phase code, and are dissolved by assigning bracketing and within-set identity to distinct codes. *Centre:* bracketing depth is carried by a phase-address code – β-troughs gating successive embedding levels are offset by a fixed increment Δφ per level, so a daughter merged at depth d acquires coupling phase $\Phi(d) = d \cdot \Delta\varphi$ relative to the δ-θ cycle (clock). Because the two bracketings of a matched string place different constituents at the deep-phase address (orange), they yield categorically distinct δ-θ-to-$\gamma$ coupling phase profiles despite identical terminals and identical numbers of binary combinations. *Right:* within a set, each daughter's $\gamma$-feature bundle locks to an arbitrary $\gamma$-phase slot functioning as an order-free hash key, with headedness fixed by maximal $\gamma$ to δ-θ coupling, $H(\ell_i,\ell_j) = \mathrm{argmax}_x$ MI(φ_δθ; a_{γ,x}), and the winning bundle inheriting the category tag $\langle \Phi(d), \Gamma \rangle$; a β-band 'commit' burst then seals the set ($\mathcal{S}$), suppressing the daughters' $\gamma$ carriers and writing the tag to a read-only label register, freezing bracketing while leaving within-set order undefined.

Overall, the implication of this section is best stated as a division of labor: Any single code that uses timing to mark order will destroy grouping, and any code that discards order to preserve grouping loses the information the parser actually receives. Cross-frequency coupling escapes the dilemma by using two coordinates of the same signal for two different jobs: the slow phase at which a burst occurs says how deep in the tree the constituent sits, and the fast slot within that window says only which of the two daughters it is, carrying no information about which arrived first. One consequence follows that no standard sentence-versus-wordlist contrast can ever deliver: for parses of a fixed string, bracketing should be recoverable from which content occupies which phase, and not from the distribution of phases alone, which is identical across the two parses in cases of structural ambiguity. In general, it is recoverable from the sequence of composites written at each seal (§5.1).

## 6. Candidate Neural Mechanisms Across Scales

The previous section developed one concrete instantiation of the core algebraic admissibility conditions. Here we ask, systematically, which neural mechanisms detectable across recording scales could enforce the algebraic properties developed in the preceding sections, and which predictions each one makes. As above, our goal is not to crown one favored mechanism, since this naturally remains an empirical question. Rather, our goal is to specify which mechanisms are admissible under the algebraic constraints and what each entails at different scales.

Two preliminaries, outlined below, will prevent our discussion from being reduced to a list of neural correlates. The first fixes how strong a claim a given piece of evidence can support (§6.1). The second fixes what a mapping from algebra to physiology *must contain* in order to count as a linking hypothesis (Embick & Poeppel, 2015; Poeppel, 2012; Poeppel et al., 2012).

### *6.1. Three levels of neural admissibility*

We distinguish three increasingly strong levels at which a neural signal can be said to be admissible for an algebraically defined operation.

**Weak admissibility**: A neural signal is statistically sensitive to an algebraic property, such as grouping depth or node closure. This is useful for localization and hypothesis generation, but is not yet mechanistic: sensitivity to a property does not show that the signal carries out the operation that the property defines (Ross, 2025).

**Operational admissibility**: The signal changes at the time and scale at which the operation is required, dissociates from lexical and task confounds, and its defining algebraic information is *selectively exploited* by downstream syntactic or semantic processing. At this level the signal behaves as the operation would have to behave, even if its causal role is not yet established.

**Causal admissibility**: Perturbing the signal's source or its coupling pathway selectively disrupts the corresponding operation while sparing matched non-syntactic processes. This is the strongest level and the one at which a mechanism, rather than a correlate, has been identified (Barack et al., 2022; Ross & Bassett, 2024; Siddiqi et al., 2022).

These three levels are close in spirit to Embick and Poeppel's distinction between correlational, integrated, and explanatory neurolinguistics (Embick & Poeppel, 2015). The difference in emphasis is that their triad grades the relation between theories, asking how tightly linguistic and neural vocabularies are aligned, whereas ours grades the evidential status of a signal with respect to a formally specified operation. A result can be 'integrated' in their sense – stated in a shared vocabulary – while remaining only weakly admissible in ours, if the shared vocabulary is a feature annotation rather than an operation. In addition, these distinctions also parallel Krakauer and Ramsey's argument that a neural state carrying information about some property is distinct from that information being exploited in the functional role attributed to the state (Krakauer & Ramsey, 2026); for present purposes, operational admissibility therefore requires that the defining algebraic distinction itself matter to downstream computation.

Much current evidence in the literature reaches only weak admissibility (Brennan et al., 2016; Brennan & Martin, 2020; Cai et al., 2026; Caucheteux et al., 2023; Chen et al., 2025; Hale et al., 2022; Jamali et al., 2024; Lyu et al., 2025; Stanojević et al., 2023; Weissbart & Martin, 2024). Some intracranial and stimulation work arguably gets closer to operational or causal admissibility for circumscribed general components of language (Forseth et al., 2020; McCarty et al., 2023; Woolnough, Thomas, et al., 2026; Woolnough & Tandon, 2024), such as lexical access (Forseth et al., 2018), but even this work leaves open the question of specific mechanisms for properties of language that are more precise than general composition demands. The purpose of the present section is to specify how these stronger levels can be reached for algebraically defined syntax-semantics.

### *6.2. What a linking hypothesis must contain*

A linking hypothesis must contain four components: an algebraic source object, an algorithmic transition that approximates it under real-time constraints, a physiological realization, and a falsification condition. Without the first component, neural data are only localized correlates.

Without the last, the algebra becomes purely ornamental and unfalsifiable. The mechanisms surveyed below are evaluated against this standard.

Neural recording scales themselves do not form a simple nested hierarchy of one underlying signal. Single units, multi-unit activity, local field potentials, high-gamma power, low-frequency phase, and inter-areal connectivity have overlapping but non-identical biophysical generators (Buzsáki, 2019; Buzsáki et al., 2012; Buzsáki & Wang, 2012; Lei et al., 2026; Ryun et al., 2025). A unit may carry lexical or conceptual selectivity while the local field reflects synaptic input, local coordination, or population timing; broadband gamma may index local population firing; low-frequency phase may organize excitability and long-range timing; directed measures may reveal state-dependent communication between regions.

This complex landscape offers many theoretical opportunities, because language's formal operations differ in what they demand. Lexical identity and semantic features can plausibly be carried by local units and ensembles; binary composition may require transient local binding together with field coordination; non-associative depth and workspace retrieval may require timing codes or recurrent state dynamics; colored filtering may require interaction between local semantic role information and wider control systems. The algebra therefore predicts *scale dependence* (Figure 8).

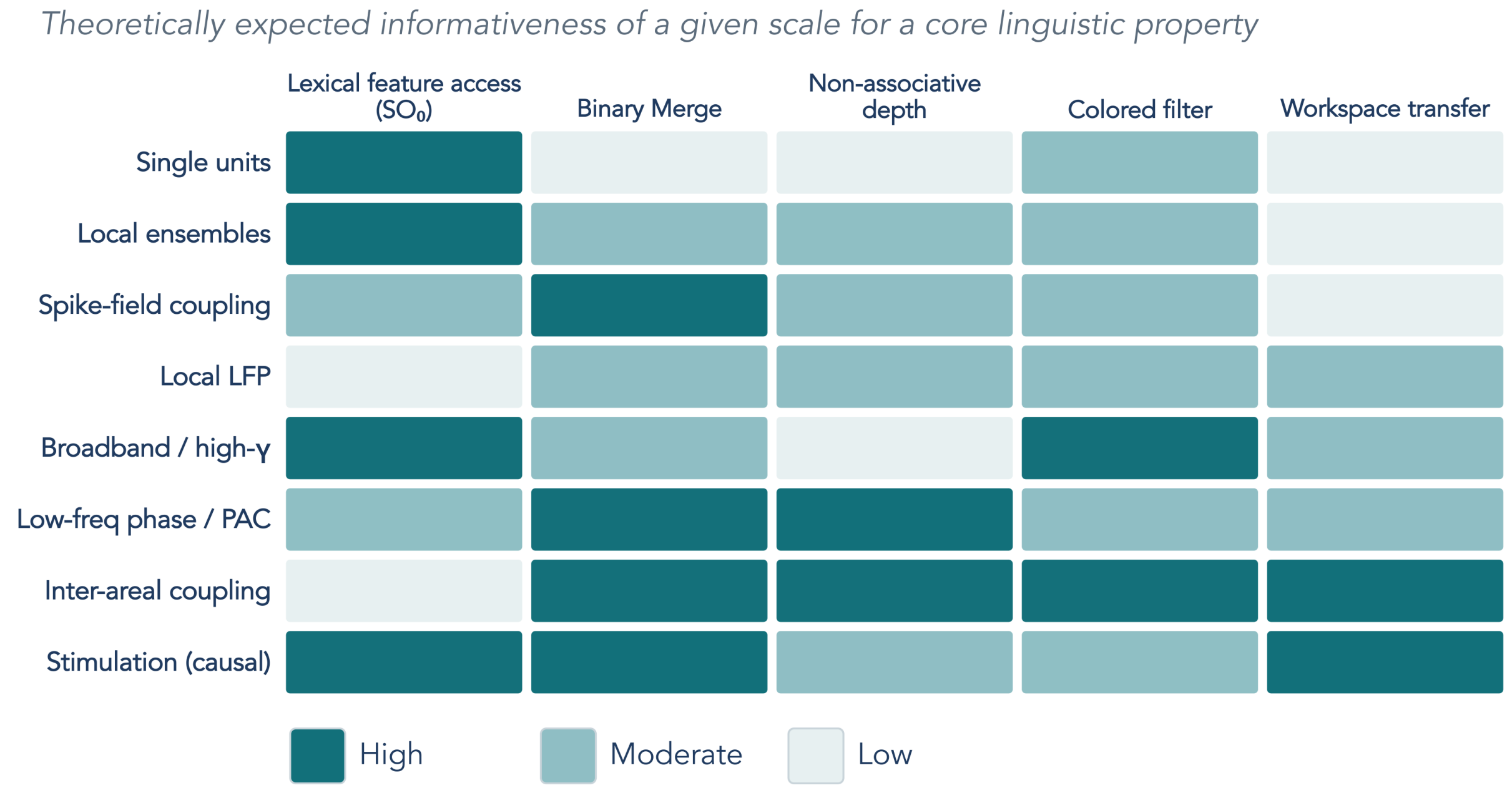

**Figure 8. Neural recording scales that could hypothetically reveal core linguistic features isolated by modern mathematical linguistics.** Expected informativeness of each recording scale (rows) for each algebraic operation (columns), graded high, moderate, or low. The shading is a theoretical prediction surface derived from the mechanisms developed in the main text.

### *6.3. Units and local ensembles: indices of combinable lexical features*

Single units can encode semantic, lexical, phonological, or task-relevant features with considerable specificity (Dijksterhuis et al., 2024; Jamali et al., 2024; Yan et al., 2026). Such codes are plausible substrates for the atomic inventory of lexical features, $SO_0$, from which linguistic objects are built. Units and small ensembles may carry information about lexical identity, conceptual category, animacy, concreteness, or morphosyntactic features. A unit-level code alone, however, is unlikely to implement higher-order syntax unless it participates in a coordination scheme that assigns structural role and grouping (Chen et al., 2026; Lakretz et al., 2026; Qi et al., 2025). The critical unit-level signature for syntax is therefore not likely to be a unit that fires for Merge. Rather, it is a unit or ensemble whose spiking becomes phase-locked (Murphy, 2020b), gain-modulated (Martin, 2020; Murphy, 2020a), or subspace-aligned (Johnston et al., 2024; Semedo et al., 2019; Tafazoli et al., 2026) when its feature is recruited into a particular structural role. Indeed, no coherent single-unit computation corresponds to a dedicated 'Merge neuron', because Merge is a relation between two objects rather than a property of any one item.

The faithfulness conditions of the function-space construction can help refine one prediction at this scale into an explicit – and somewhat unexpected – commitment about representational geometry. Faithful reconstruction of constituency in that construction begins with lexical features whose representations are injective and approximately linearly independent. This is in tension with distributed lexico-semantic embeddings, in which semantic similarity is typically expressed by linear proximity (Katlowitz et al., 2026). The brain may therefore maintain partially separable representational geometries for lexico-semantic similarity and for constituency-preserving syntactic features (Figure 9).

Concretely, let $z_i$ be the neural representation of a lexical item. We predict a decomposition,

$$z_i = z_i^{syn} + z_i^{sem} + \varepsilon_i,$$

into components in two subspaces, in which the constituency-supporting subspace has lower mutual coherence than the semantic-similarity subspace. Here $\varepsilon_i$ is the residual component not captured by the two identified subspaces. The decomposition must be identified operationally or the prediction is vacuous, since absent a constraint one can move mass between the two components at will. We therefore define the subspaces by what they do: $V_{syn}$ is the subspace maximizing held-out decoding accuracy for bracketing under matched terminals, and $V_{sem}$ is the subspace maximizing correlation with semantic-similarity ratings or distributional embeddings, each estimated on independent data. The atoms' projections into each then have well-defined mutual coherence. For $q \in \{\mathrm{syn}, \mathrm{sem}\}$, define

$$\mu_q := \max_{i \neq j} \frac{\left|\left\langle z_i^q, z_j^q \right\rangle\right|}{\left\|z_i^q\right\| \left\|z_j^q\right\|},$$

so that the prediction is

$$\mu_{\mathrm{syn}} < \mu_{\mathrm{sem}}.$$

This makes the comparison one between two identified quantities rather than between two labels. Note that we do not require $V_{syn}$ and $V_{sem}$ to be orthogonal; the claim concerns the coherence of the atoms within each, not the relation between the subspaces. Because the maximal coherence of *n* atoms in a *k*-dimensional subspace is bounded below by the Welch bound, which falls with *k*, the comparison is meaningful only between subspaces of matched dimension, or after rank-matching by projection – and since the maximum is set by a single worst pair, the average coherence (or the Babel function at the relevant sparsity) is the more stable statistic and should be reported alongside it.

Hence, the subspace that best reconstructs bracketing should be more nearly orthogonal, or otherwise more disentangled, than the subspace that best predicts semantic-similarity ratings or distributional embeddings. A failure to recover any such factorization would weaken the present function-space route as a linking hypothesis, since it would indicate that the brain does not separate the geometry that makes atoms re-combinable from the geometry that makes them semantically similar; this would consequently support many of the theoretical presuppositions of recent LLM-brain alignment research concerning distributional semantics (Harris, 1968; Turney & Pantel, 2010).

This coherence asymmetry has a natural mechanistic identity. Associative-memory networks whose connectivity is a sparse expander graph store patterns as a low-coherence, near-orthogonal code, precisely because near-orthogonality is what lets an error-correcting network separate and recover overlapping stored states (Chaudhuri & Fiete, 2019). The constituency-preserving subspace is therefore what one expects of a code optimized for recombination and robust extraction – sparse and near-orthogonal – whereas the semantic-similarity subspace is a dense, correlated code in which proximity encodes meaning. On this reading the decomposition outlined above is not merely a geometric conjecture but the signature of two coexisting memory regimes in one population, and the low coherence of the syntactic subspace is explained rather than stipulated: recombination under a content-addressable read-out requires it. Higher-order (simplicial) variants of these networks, which store relational rather than only pairwise structure and thereby raise capacity (Burns & Fukai, 2023), are the natural candidates for a store that must hold typed, bracketed objects rather than unstructured patterns.

## Two geometries for lexical atoms

*Faithful constituency needs near-independent atoms, whereas semantic similarity needs graded proximity. The brain may keep both, in separable subspaces.*

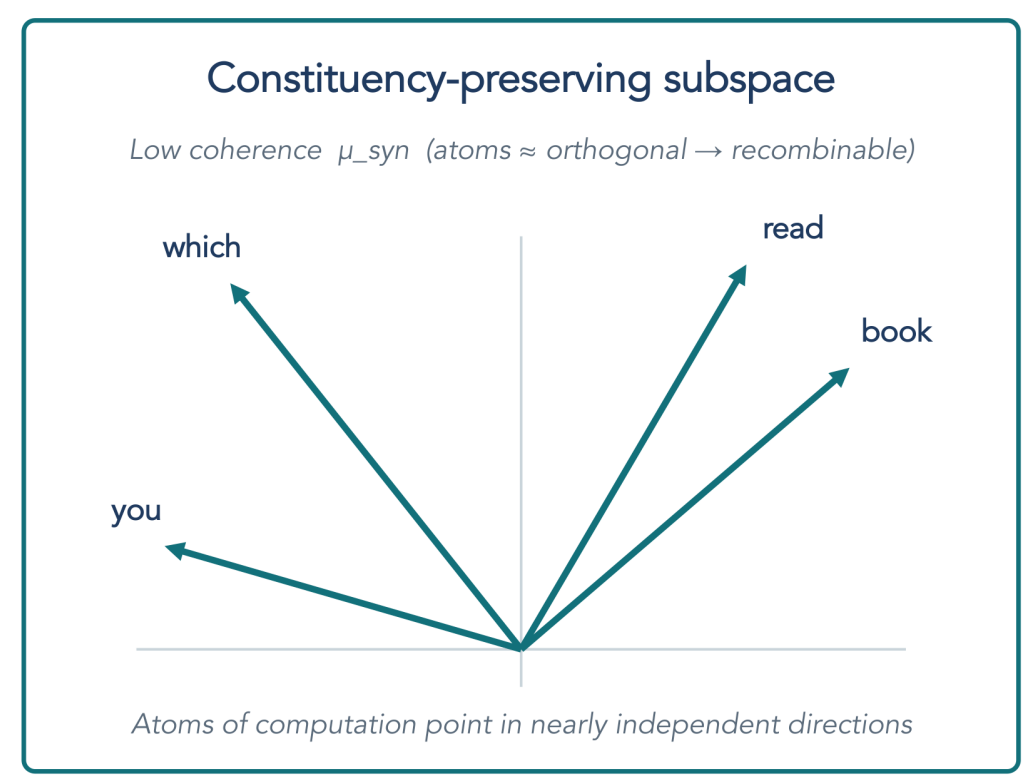

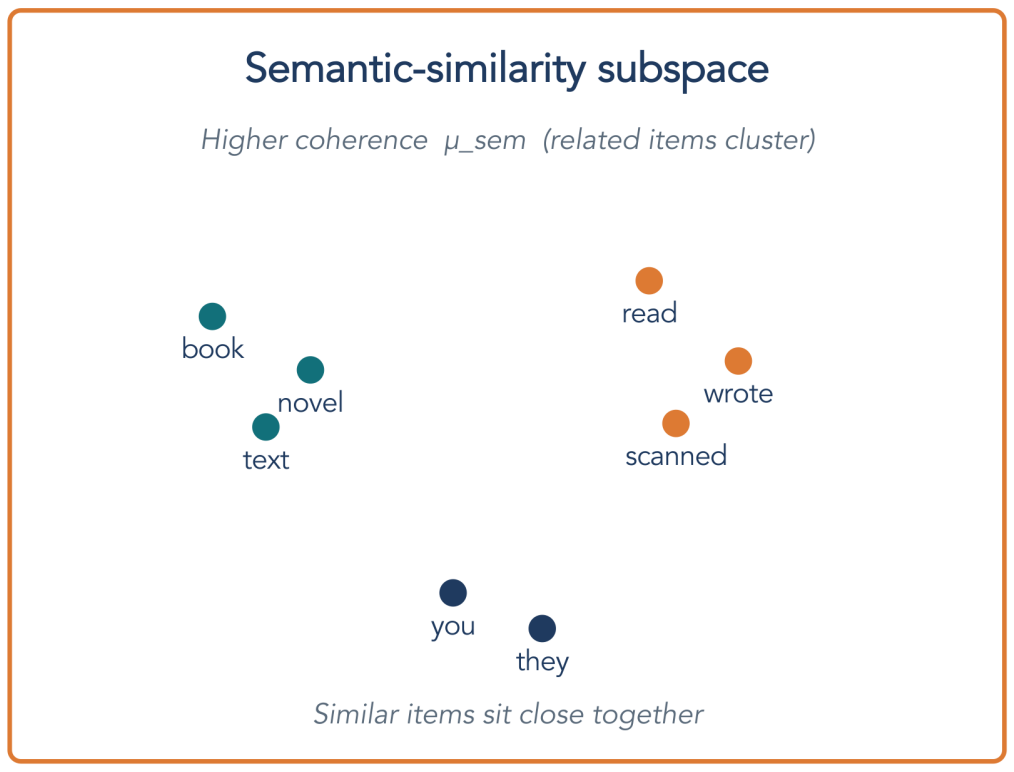

Prediction: $z_i = z_i^{syn} + z_i^{sem} + \varepsilon_i$ with μ_syn < μ_sem

*The subspace that best reconstructs bracketing should be more disentangled than the one that predicts semantic similarity*

## Four composition laws for linguistic structure-building

*Given neural codes u, v of two objects, different composition laws make different algebraic commitments*

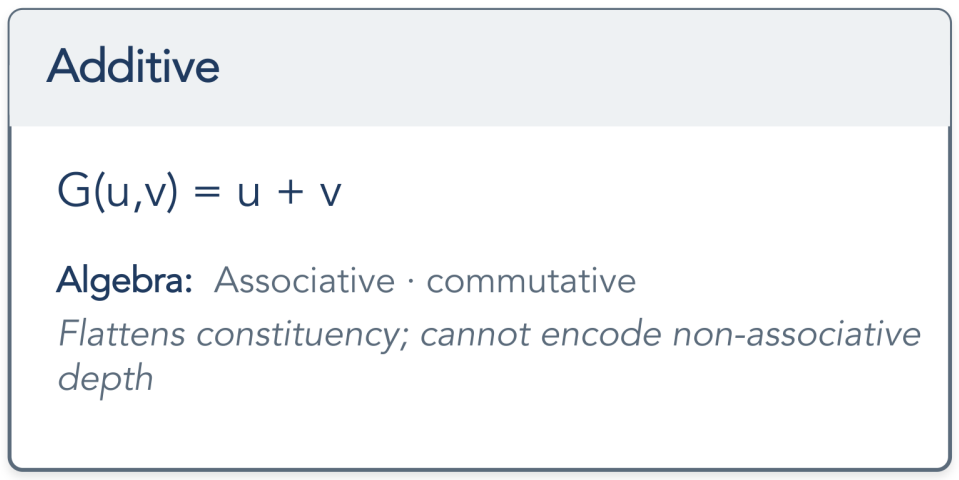

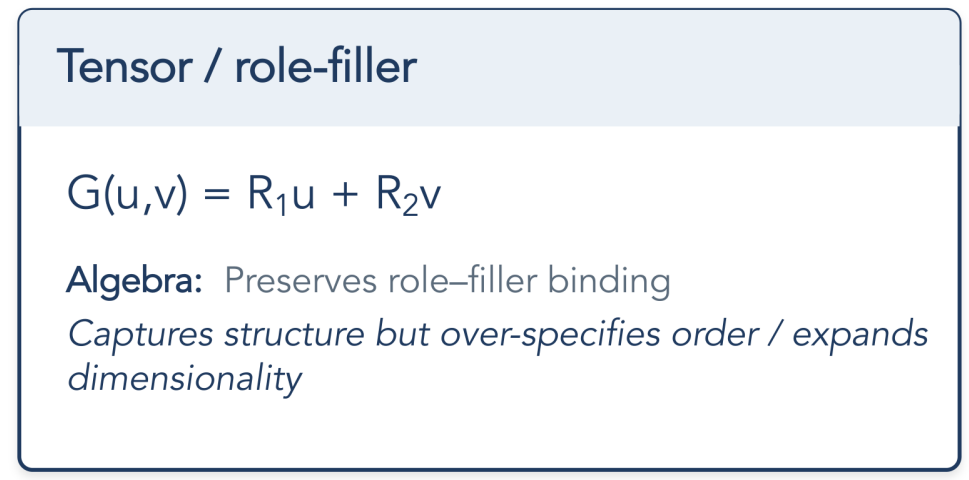

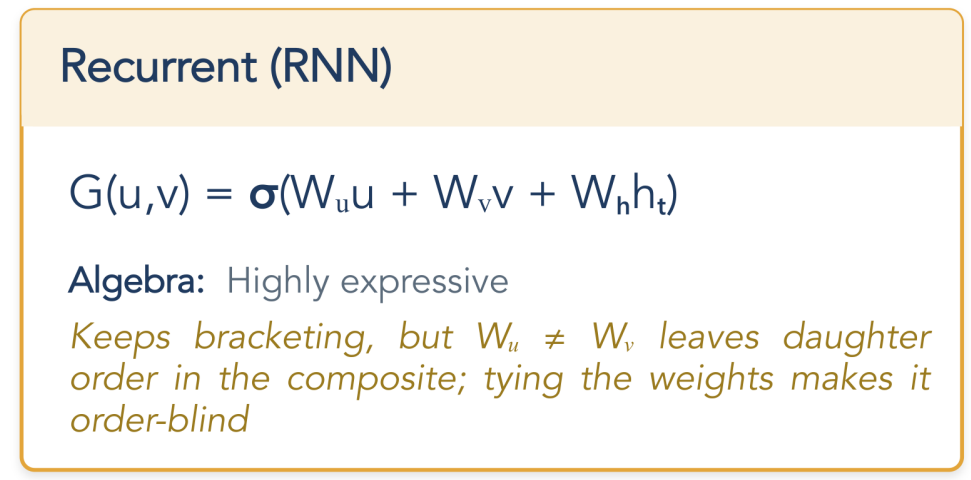

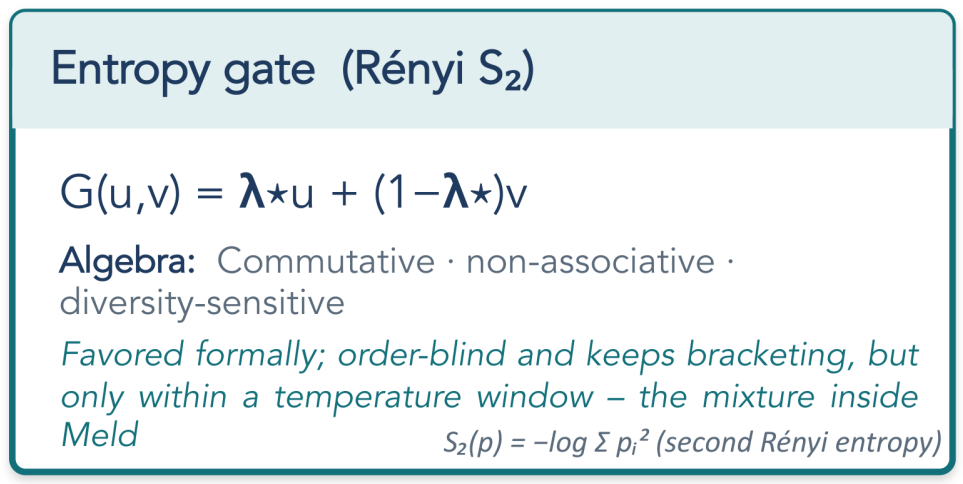

**Figure 9. Properties of neural geometry and composition relevant for experimental assessment.** Top: Candidate geometries for constituency-preserving and semantic-similarity representations. The constituency-preserving subspace (left) holds lexical atoms in nearly independent directions (low mutual coherence $\mu_{\text{syn}}$), so that they remain recombinable, whereas the semantic-similarity subspace (right) places related items close together (higher coherence $\mu_{\text{sem}}$). The prediction is a decomposition $z_i = z_i^{\text{syn}} + z_i^{\text{sem}} + \varepsilon_i$ with $\mu_{\text{syn}} < \mu_{\text{sem}}$: the subspace that best reconstructs bracketing should be more disentangled than the one that best predicts semantic similarity. Bottom: Four composition laws. Given neural encodings $u$ and $v$ of two objects, rival composition laws make different algebraic commitments and are separable under matched controls. The additive law is associative and flattens constituency; the role-filler law preserves role-filler binding but over-specifies order (note that we simulate the norm-preserving variant, which does so without expanding dimensionality); the recurrent law preserves bracketing but, with separate weights for its two daughters, leaves their order in the composite; and the second-Rényi entropy gate is a specific commutative, non-associative, diversity-sensitive operation, whose mixture is the one Meld carries. A fixed mean-pooling law shares the gate's commutativity, non-associativity and order-blindness but has a content-independent mixture, and is separated from the gate only by contrasts that a fixed mixture cannot resolve (§4.9.1). The laws are not ranked on one axis. Bracketing-decodable and blind to daughter order is a class, containing the entropy gate, fixed mean-pooling, any saturating unit into which both daughters enter through the same weights, and Meld (§4.9.1), which combines the entropy mixture with such a unit; the ordered laws fall outside it. Members of the class are separated by the balanced four-leaf contrast and by sensitivity to daughter disagreement, not by bracketing accuracy (§4.9.1, §7.2).

In summary, there is an inherent tension between two things a lexicon must be able to deliver. To register that ‘dog’ and ‘cat’ are similar, we require their representations to be close together. To combine them into structures and pull them apart again without corrupting either, we require near-independence. These are different geometries, and one subspace cannot deliver both, and thus the prediction is that the brain keeps both.

### *6.4. Spike-field coupling: binding local items into structured roles*

Spike-field coupling is a natural candidate for connecting local item codes to field-level structure (Vissani et al., 2025). If spikes encode lexical or semantic features and the phase of the local field supplies a temporal coordinate system, then spike timing relative to phase can encode when a local item is available for binding or retrieval. This implements a role-filler relation without requiring every structural role to have a dedicated local neuron (Van Bree, 2024b). Algebraically, the appeal here is that coupling can preserve local identity while embedding it in a larger coordination regime. Non-associativity can be approximated if different groupings impose different nested timing histories, and colored filtering can be approximated if phase alignment or coupling strength differs for compatible versus incompatible role assignments.

Notably, phase is not the only coordinate the field supplies: human single neurons can be tuned to the instantaneous frequency of the local field, and frequency-tuned and phase-tuned cells form largely non-overlapping populations (Jourahmad et al., 2026). For the NAP, this furnishes a second, approximately independent address – a candidate frequency-address complementing the phase-address of §5.1 – along which structural role, lexical features, or grouping depth could be encoded.

More broadly, for cross-scale inference – bridging signals detected at the spiking and LFP levels – no single metric is sufficient. A robust workflow typically combines spike-phase coupling (Canolty et al., 2010; Jacobs et al., 2007; Vinck et al., 2012; Vissani et al., 2025), spike-field coherence, standard time-frequency analyses, and some combination of latent-variable methods (e.g., GPFA, LFADS) (Pandarinath et al., 2018; Yu et al., 2009). More recent methods include Joint SSMT, a Bayesian state-space framework that jointly infers LFP spectrograms and spike-field coupling strength (Zheng et al., 2026).

In humans, spike-phase coupling has been linked to speech sound accuracy (Vissani et al., 2025), memory formation (Rutishauser et al., 2010), spatial navigation (Guth et al., 2025; Watrous et al., 2018), and temporal-coding schemes (Jacobs et al., 2007). Recent work shows that high-frequency LFP oscillations decode attentional states better than or on par with spikes when multiple electrodes are used (Prakash et al., 2021, 2024). This directly contrasts with sensory (Kanth & Ray, 2020) and motor decoding (Hwang & Andersen, 2013), where multichannel spikes outperform multichannel LFP. Relatedly, LFPs are useful in revealing information about perceptual decision-making that is not readily apparent in spiking activity (Hou et al., 2026). While low-frequency phase is informative for decoding behavioral outcomes (Busch

& VanRullen, 2010), whether decoding outcomes based on LFP features are as informative as decoding based on spikes remains less well addressed (Prakash et al., 2024).

Following much of the logic we have followed thus far, lexical features may be locally readable in spikes, high gamma activity, and LFPs; compositional and discourse-level semantics may depend on field-level coordination that organizes when and where those local feature codes can be expressed. According to the *spatial computing* framework (Chen et al., 2026), lower frequencies are the 'stencil' constraining where the 'paint' of spikes can occur. Properties of the LFP and lower frequency components may thereby act as *control* fields that sculpt spiking activity, while gamma/spiking carries item-specific *content*, as the ROSE model also proposes (Murphy, 2024, 2025, 2026a). Recent work on endogenous ephaptic coupling gives this control-field proposal stronger mechanistic grounding: self-generated fields decorrelate population activity, expand effective dimensionality and improve decoding, whereas the same field imposed exogenously has the opposite effect (Banaie Boroujeni & Kastner, 2026). Although the contribution of this channel in cortex remains to be established, these results demonstrate that endogenous fields can actively reorganize population geometry rather than merely reflect spiking activity.

Within this context, broadband high-frequency $\gamma$ activity is commonly read as an index of local population activation. It is a strong candidate marker of lexical retrieval and semantic feature activation, but likely not, in itself, a marker of algebraic properties of syntax. Its value for the present framework is instead *mediating and diagnostic*: when combined with multi-scale recordings, it can help distinguish effects expressed at finer scales (spiking, spike-LFP coupling) from effects expressed at broader scales (inter-areal coherence, directed connectivity).

Lastly, we recently documented a sentence-sensitive alpha instantaneous-frequency ramp during reading (Woolnough, Murphy, et al., 2026). This may index a low-frequency control variable that progressively reconfigures communication between language and default-mode networks as a structured representation becomes available to conceptual interpretation. Within the present framework, this would constitute not the local code for linguistic composition itself (especially given the lack of Jabberwocky effects), but a frequency-sensitive workspace/interface signal that gates distributed gamma or spiking content and coordinates its transfer across cortical systems.

### *6.5. Low-frequency phase and non-associative grouping*

Low-frequency phase and cross-frequency coupling have been argued to be reliable signatures for encoding hierarchical grouping (Kazanina & Tavano, 2023; Weissbart & Martin, 2024). Phase can provide temporal windows for grouping, $\gamma$ amplitude can reflect local activation within those windows, and nested oscillatory organization can preserve grouping history. A simple additive firing-rate code struggles with non-associativity, but nested phase-amplitude structure can, in principle, distinguish $\mathcal{M}(\mathcal{M}(A,B),C)$ from $\mathcal{M}\big(A,\mathcal{M}(B,C)\big)$, exactly as the depth-address code of §5 proposes. Recent evidence from MEG provides convergent support for treating low-frequency phase as a structural coordinate. Martorell and colleagues found that sentence-frequency phase angle shifted with reversed Spanish/Basque structure independently of lexical content, while frequency indexed coarser sentence timescale and phase differentiated finer-grained internal structure (Martorell et al., 2026). This does not yet establish the stronger claim that phase encodes non-associative bracketing or depth (Murphy, 2025), but it makes that hypothesis considerably more targeted and plausible.

One way to make claims about "nested oscillatory organization" more precise is to read the tree's *dominance order* directly off the topology of cross-frequency coupling rather than off any single phase angle. If each constituent's content is carried by a high-frequency (γ) burst, and the low-frequency phase that modulates that burst indexes the constituent that contains it, then the directed relation 'phase of X modulates amplitude of Y' induces a partial order over the active constituents. The prediction is that the partial order induced by the directed PAC graph should be isomorphic to the dominance relation of the syntactic tree, and equivalently that its transitive reduction should be isomorphic to immediate dominance, up to the depth bound of §5.1. Under this reading, $\mathcal{M}(\mathcal{M}(A,B),C)$ and $\mathcal{M}(A,\mathcal{M}(B,C))$ differ not only in the phase angle assigned to a constituent but in *which burst is nested within which*; a containment topology rather than a scalar.

Notice that the containment topology developed here and the angular phase-address of §5.1 are not two implementations at the same level. The containment code carries the dominance order of the tree. The static angular code carries a depth function, which determines the tree for parses of a fixed string but not in general (§5.1), and is one coordinate of the dynamic grouping code (phase address × composite identity × sealing history) that carries dominance. However, the two predict different depth-scaling, which makes them dissociable rather than merely complementary. The angular code, $\Phi(d) = d\cdot\Delta\varphi$ with $\Delta\varphi \approx \pi/2$, has a wrap-around depth $d^* = 2\pi/\Delta\varphi \approx 4$, at which the phase recycles, $\Phi(d^*) \equiv \Phi(0)$. Depths separated by $d^*$ then share a coupling phase and become confusable. Consequently, below the bound the angular code does not degrade with depth at all: a phase address is an angle, and no angle is harder to estimate than another, so wherever the carrier ratio is adequate to the ladder, depth recovery is flat. Above the bound the code does not decline gracefully either; it aliases, producing categorical confusions between depths $d$ and $d + d^*$ and a depth profile that rises into the unaliased interior of the cycle and falls away on either side of it, a shape no capacity-limited mechanism produces. The containment code reads dominance off the transitive reduction of the directed PAC graph, a topological rather than angular relation, and is not intrinsically capped; but it must estimate one edge per level, so its errors accumulate along the chain and it predicts smooth, monotone, SNR-limited degradation from the first level. A parametric depth manipulation ($d = 1\ldots5$, with matched terminals and transition statistics) therefore adjudicates between them on the shape of the depth profile rather than its height: flat decoding within the measured carrier ratio, with non-monotone aliasing at lag $d^*$ beyond it, supports the angular phase-address, whereas monotone decline supports the containment code. Below (§6.5.1), we confirm both profiles under matched terminals and 1/f noise. The two need not be mutually exclusive, since different bands or regions may realize different codes, but the experiment separates them, and the non-monotone aliasing at $d^*$ is the unique signature of the angular scheme.

Some caveats are required here. We do not predict that every syntactic tree maps to a *unique oscillatory tree*, since biological rhythms are variable and task-dependent. We predict instead that when stimuli force different hierarchical groupings with matched lexical content and comparable predictability, low-frequency phase alignment and PAC should change at the moment the grouping becomes necessary, and that the direction of cross-frequency interaction should reflect whether a local representation is being recruited into a higher-order structure, or whether the parsing state is constraining local interpretation.

*6.5.1. Simulation of angular phase address vs. containment code*

To simulate some of the dynamics entertained thus far, sixteen channels were synthesized at 500 Hz over 4-second trials. Seven contents occupied seven orthonormal spatial patterns; every trial contained all seven, and only the assignment of content to chain position varied, so terminals were matched across depths. The probe content was fixed and its depth $d \in \{1, \dots, 6\}$ decoded; chance is $\frac{1}{6}$. Background was 1/f amplitude noise, calibrated band by band so that the signal-to-noise ratio within each analysis band (1-10Hz and 65-95Hz) equaled the nominal value (150 trials per depth per condition).

Under the angular model, a δ-θ control field at $f_{\delta\theta} = 4\ Hz$ carried a nested β carrier; the constituent at depth $d$ emitted an 80Hz γ burst (Gaussian envelope, 0.4 β cycles) at the β trough falling at phase $\Phi(d) = d \cdot \Delta\varphi\ (mod\ 2\pi)$, so the addresses are the troughs and their number is the carrier ratio. The δ-θ phase was taken from the channel mean band-passed to 2-6 Hz, the probe's γ envelope from its spatial filter, and the address estimated as the circular mean of δ-θ phase weighted by that envelope, $arg\ \Sigma\ env(t)\ e^{i\varphi(t)}$, then assigned to the nearest $d \cdot \Delta\varphi$, with ties between aliased depths broken at random.

Under the containment model, each content carried an intrinsic low-frequency carrier at a distinct frequency (2.5-8.5 Hz in 1Hz steps) with a per-trial random phase, and the γ amplitude of the node at chain position $k$ was modulated by the carriers of *all* its ancestors with weight decaying geometrically in distance (0.9, ratio 0.5), so the PAC graph depicted realizes dominance and its transitive reduction realizes immediate dominance. The read-out is the Tort modulation index (18 phase bins) between the low-frequency phase of every content and the γ amplitude of every other gave a directed 7 × 7 matrix; this was thresholded (0.01, calibrated once and held fixed across SNRs), oriented by the dominant direction, closed transitively and reduced, and the probe's depth taken as the longest path to it in the reduction. When the graph left the probe unreachable, the trial was assigned a uniformly random depth rather than a default, so chance is unbiased. The common-carrier control replaced the seven intrinsic frequencies with a single 4 Hz carrier, leaving everything else unchanged.

Simulating both codes against matched terminals, parametric depth $d = 1 \dots 6$ and 1/f noise confirms that they dissociate. When the carrier ratio is smaller than the range of the ladder, the angular code produces exactly the predicted fingerprint. At $\frac{f_\beta}{f_{\delta\theta}} = 4$ and 0 dB γ-band SNR, depths 3 and 4 – the two that retain unique addresses – are decoded at ceiling (1.00, 1.00), while depths 1, 2, 5 and 6 fall to chance-between-two (0.43, 0.45, 0.49, 0.51), and the confusion mass concentrates at lag $d^*$: 0.57 and 0.51 for the pair (1, 5), 0.55 and 0.49 for (2, 6). Accuracy against depth is not merely non-monotone but non-monotone in a specific way – it rises into the unaliased interior of the cycle and falls on either side of it – which is not a profile any capacity-limited code produces.

When the ratio is adequate, however, the angular code does not decline with depth at all. At $\frac{f_\beta}{f_{\delta\theta}} = 7$, accuracy is 1.00 at every depth from 1 to 6 (Figure 10D). This confirms the flat profile anticipated in §6.5, and it is the more diagnostic of the two: a phase address is an angle, and as noted one angle is no harder to estimate than another, so an adequately carried angular code has no depth cost at all. The containment code, by contrast, must estimate one edge per level, and its errors accumulate along the chain: accuracy declines monotonically – 1.00, 0.97, 0.87, 0.69, 0.51, 0.45 across $d = 1 \ldots 6$ at the same SNR – with an aliasing index indistinguishable from zero at every SNR tested.

As shown, simulating the containment code with all constituents carrying a common low-frequency carrier returns depth decoding at chance – 0.162 against a chance level of $\frac{1}{6} = 0.167$ – at high SNR. More fundamentally, a sum of same-frequency sinusoids is a single sinusoid:

$$\Sigma_j m_j \, cos(2\pi \, f \, t \, + \, \psi_j) \; = \; R \, cos(2\pi \, f \, t \, + \, \Psi).$$

Here, $R \geq 0$ and $\Psi$ are the resultant amplitude and phase, defined by the phasor sum

$$Re^{\mathrm{i}\Psi} := \sum_j m_j \, e^{\mathrm{i}\psi_j}.$$

A node whose γ amplitude is modulated by several ancestors at a common frequency therefore carries a modulation algebraically indistinguishable from modulation by a single ancestor at the resultant phase.

What the containment code requires, then, is that constituents be separated on some coordinate other than phase, so that coupling can be attributed source by source. The frequency address of §6.4 is one possible candidate, furnished by the observation (noted above) that human single neurons are tuned to the instantaneous frequency of the local field and that frequency-tuned and phase-tuned cells form largely non-overlapping populations. Giving each constituent an intrinsic low-frequency carrier at a distinct frequency – here 2.5 to 8.5 Hz in 1 Hz steps – makes the PAC graph identifiable and recovers the monotone depth profile reported above. The containment code of this section is therefore not an alternative to the frequency address of §6.4; it *presupposes* it. If constituents are not frequency-separated, the containment reading of cross-frequency coupling is unavailable, and the angular reading is the only one on offer.

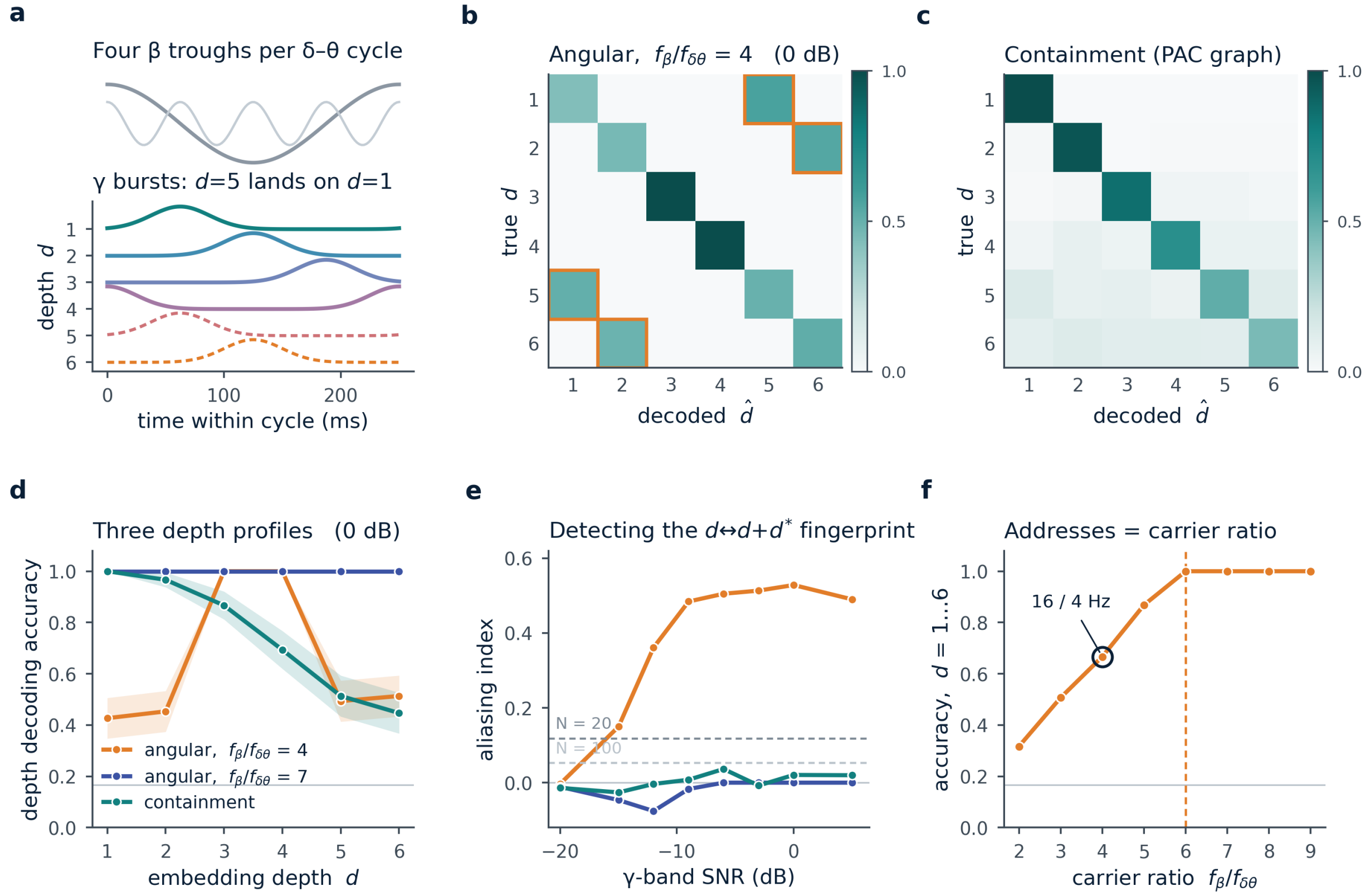

**Figure 10. Simulation results indicating that the carrier ratio, not Δφ, decides whether depth is addressable.** (**a**) One δ-θ cycle of a synthesized trial. Top: the δ-θ control field (grey) with its nested β carrier; markers give the four β troughs, which are the available phase addresses at $\frac{f_\beta}{f_{\delta\theta}} = 4$. Bottom: the γ envelope of the constituent at each depth $d = 1 \ldots 6$. Depths 5 and 6 (dashed) fall on the troughs of depths 1 and 2. (**b**) Confusion matrix for the angular read-out at $\frac{f_\beta}{f_{\delta\theta}} = 4$ and 0 dB γ-band SNR, 150 trials per depth, matched terminals. Depths 3 and 4 retain unique addresses; the remaining mass concentrates in the four ringed cells at lag $d^* = 4$. (**c**) Confusion matrix for the containment read-out at the same SNR: a graded spill to neighboring depths and no lag-$d^*$ structure. (**d**) Accuracy against depth for the three codes. The angular code is non-monotone at an inadequate carrier ratio and flat at an adequate one; the containment code declines monotonically. Bands: 95% CI. (**e**) Aliasing index – confusion mass at lag $d^*$ minus the mean off-diagonal mass elsewhere – against γ-band SNR. Dashed lines: one-sided detection thresholds for 20 and 100 trials per depth. (**f**) Depth accuracy against the carrier ratio at high SNR, decoding $d = 1 \ldots 6$. Accuracy reaches ceiling exactly at a ratio of 6, the number of depths to be separated; the circled point is the worked example of §5.1 ($f_\beta \approx 16\,Hz$, $f_{\delta\theta} \approx 4\,Hz$), which supports only four addresses.

To conclude, these simulation results serve to replace a plausible intuition with a sharper and different empirical target. The original intuition was that a phase code should degrade with depth. It does not: a phase address is an angle, and one angle is no harder to estimate than another; given enough β cycles per δ-θ cycle to give every depth its own address, depth decoding is flat all the way down. What limits an angular code is not depth but the carrier ratio it is built from. When the depth ladder outruns that ratio, the code does not fade – it aliases, confusing depth d with depth d + d*, and accuracy rises into the unaliased interior of the cycle and falls on either side of it. That non-monotone profile is not something any capacity-limited mechanism produces, and makes the angular hypothesis worthy of further exploration.

The containment code fails in the opposite way. It must estimate one edge per level and its errors accumulate along the chain, so it declines smoothly and monotonically, with no aliasing at any signal-to-noise ratio tested. The two codes are therefore separable in a single parametric depth manipulation, and the discriminating measurement is the shape of the depth profile rather than its height. Reading dominance off the coupling graph works only if constituents are separated on some coordinate other than phase, because modulations at a common frequency sum to a single modulation that is algebraically indistinguishable from one ancestor acting alone. Handing every constituent its own low-frequency carrier restores identifiability and recovers the monotone profile. If constituents are not frequency-separated, then a fairly general claim such as 'nested oscillations carry hierarchy' is not a hypothesis one can state, let alone test, and the angular reading is the only plausible candidate remaining.

Recent intracranial evidence also bears directly on the admissibility of an angular code. As mentioned above, we recently reported a progressive ramping of instantaneous frequency in the alpha range across both language and default-mode sites during sentence reading, absent for Jabberwocky sentences and indistinguishable there from pseudoword lists (Woolnough, Murphy, et al., 2026). The measure therefore indexes semantic composition rather than structure-building as such, and constitutes a discriminating null: it separates the interface channel from the algebra rather than converging on it. Two consequences appear to follow: First, the phase offset of gamma to the low-frequency carrier is invariant across word position and sentence structure while the frequency ramps, so angular position remains a stable coordinate even as the rate of phase advance changes – the ramp is, in the present terms, output-null with respect to the gamma readout. Second, a monotone frequency ramp integrates to a quadratic phase trajectory, so any angular code must be defined on instantaneous rather than nominal phase. A stationary carrier is not available as an implementation.

*6.6. Inter-areal coupling and directed dynamics: workspace coordination*

Workspaces are unlikely to be localized to a single site (Murphy et al., 2026; Woolnough et al., 2023). They require coordination among distributed cortical sites, depending on experimental task demands. Inter-areal coupling is therefore not an optional addition to local composition; it may be the mechanism by which a constructed object becomes available to subsequent interpretation, memory, decision, and externalization. The total workspace transition operator developed earlier, $K = \sqcup \circ (B \otimes \mathrm{id}) \circ \Pi^{(2)} \circ \Delta$, decomposes a Merge step into coproduct extraction, gated binary composition, and product reassembly. The directed-dynamics prediction is that these sub-operations should be realized as distinguishable directed interactions rather than as undifferentiated connectivity increases.

Partial directed coherence, autoregressive hidden Markov models (arHMMs), state-space methods, and cross-region phase synchronization can all provide tests of these ideas. Distinct directed motifs may underwrite different representational transitions: for example, posterior-to-frontal flow during selection of a lexical item into structure, frontal-to-temporal flow during top-down phrase-structure labeling or theta-role constraint, temporal-to-parietal flow during semantic integration, and recurrent frontotemporal loops during workspace retrieval or reanalysis.

For example, arHMMs provide an empirical framework for testing whether language processing proceeds through discrete or metastable network states, each governed by a distinct pattern of local and inter-areal dynamics (Saravani et al., 2019). Such models have identified reproducible single-trial sequences of distributed cortical states during word production without assuming that the relevant processes occur at identical times across trials (Forseth et al., 2021). Applied to syntactic processing, the latent states could correspond provisionally to workspace configurations or operations such as candidate selection, composition, filtering, commitment, retrieval, and reanalysis, while the state-specific autoregressive coefficients characterize the directed cortical interactions supporting each regime. The critical NAP prediction here is not merely that several hidden states can be fitted, but that their transition graph and internal dynamics should respect independently specified relations among linguistic workspaces – for example, distinguishing alternative bracketings, and permitting access to previously constructed objects. Nevertheless, an unconstrained arHMM remains descriptive: its latent states cannot be identified with formal workspace states unless their representational content, admissible transitions, and generalization across lexical materials are independently validated.

These empirical state-space results also motivate a more specific class of transient dynamical realizations. Forseth and colleagues recovered reproducible single-trial sequences of distributed network states during picture naming, with each state defined by a distinct pattern of directed interregional interactions (Forseth et al., 2021). Those analyses do not directly establish stable heteroclinic channels or winnerless competition specifically, but they show that human language production can be organized as constrained transitions among metastable network regimes. This is the empirical phenotype for which heteroclinic dynamics was developed: stable heteroclinic channels generate reproducible sequences of metastable states that remain robust to noise while sensitive to informative perturbations (Meyer-Ortmanns, 2023; Rabinovich et al., 2008, 2012; Rabinovich & Varona, 2011). Under the NAP, however, phase-space dynamics is not yet natural language syntax. A metastable trajectory becomes a candidate realization of syntax only if its states and transition relations preserve the independently specified invariants of the linguistic workspace. This is the issue we now turn to.

### *6.7. Attractor-network realizations across scales*

The mechanisms surveyed above can be given more explicit dynamical form by entertaining several classes of graph-structured recurrent dynamics rather than committing to a single attractor architecture. First, combinatorial threshold-linear networks (CTLNs) are inhibition-dominated recurrent networks defined from a directed graph, in which discrete graph motifs predict the network's static and dynamic attractors (including limit cycles that generate ordered sequences) through parameter-independent graph rules (Curto et al., 2019, 2024; Parmelee, Moore, et al., 2022)[2]. Second, winnerless-competition dynamics (WLC) and stable heteroclinic channels (SHCs) provide a complementary realization in which the relevant computational object is a reproducible trajectory through metastable states connected by unstable directions (Meyer-Ortmanns, 2023; Rabinovich et al., 2001, 2008; Rabinovich & Varona, 2011, 2018). Third, bipartite expander Hopfield networks are content-addressable memories in which sparse expander connectivity yields exponentially many robust, self-correcting attractors (Chaudhuri & Fiete, 2019). The main construction was introduced by Chaudhuri and Fiete, and its striking theoretical property is that a network containing N-scale numbers of neurons can possess exponentially many robust stable states, approximately $2^{\alpha N}$, while correcting errors affecting a finite fraction of the input neurons.

---

[2] I thank Matilde Marcolli for pointing me to the relevance of combinatorial threshold-linear networks (CTLNs).

These architectures are therefore complementary with respect to the factorization of Merge developed above: the expander/Hopfield construction is naturally a model of workspace storage and retrieval, whereas CTLN or SHC/WLC dynamics are candidate realizations of the derivational sequencer (Figure 11).

For example, CTLNs supply a plausible dynamical vocabulary for our present claim that real-time language processing should be represented as a structured trajectory through workspace states. Let $W_t$ be the syntactic workspace after processing input up to time $t$, and let

$$\mathcal{N}(W_t) = x_t$$

be its neural realization. A parser would then be implemented not by convergence onto one sentence-level attractor, but by a sequence

$$A_{W_0} \to A_{W_1} \to A_{W_2} \to \cdots \to A_{W_n},$$

where each $A_{W_t}$is a stable or metastable population state corresponding to the currently available syntactic objects. Critically, the CTLN model must demonstrate faithful bracketing separation (e.g., via different relative firing-rate patterns, locations on a manifold, or oscillatory phases attached to the same active neurons), not merely different transient word responses.

Still, CTLNs do not automatically solve all algebraic properties of language. Consider phrasal closure. A standard CTLN selects among activity patterns supported by a fixed graph; it does not literally create a new node or subnetwork every time two constituents are merged. To model closure, a resulting attractor must become an input object of the same representational type as its components. A successful model therefore needs something like

$$G(A_X, A_Y) = A_{\mathcal{M}(X,Y)},$$

where the new state can participate in a subsequent operation of the same kind.

SHC/WLC dynamics are especially attractive for an incremental parser in one respect that CTLNs do not emphasize as strongly: metastability and departure from the current state are the primitive dynamical objects. A partial derivation must be stable enough to survive noise and maintain its currently available objects, yet labile enough for a newly arriving word, contextual cue, or retrieval event to redirect the trajectory. Stable heteroclinic channels were developed precisely to reconcile these demands of reproducibility and sensitivity. CTLNs can also generate transient and persistent sequences, and their complementary advantage is unusually strong graph-to-dynamics control: graph motifs place parameter-independent constraints on the network's attractor repertoire. We therefore do not take SHCs to replace CTLNs. Rather, both are candidate realizations of the same NAP requirement: graph-structured transient dynamics that schedule operations over a workspace, and their relative adequacy should be decided by whether they realize the independently specified workspace transition graph.

For an SHC realization, the formal workspace-transition graph can be mapped directly onto a heteroclinic network. Let

$$x \in \mathcal{W} \mapsto s_x, \qquad (x \to y) \in E_K \mapsto \gamma_{xy} \subset W^u(s_x) \cap W^s(s_y).$$

Here each workspace state $x$ is associated with a metastable saddle $s_x$, while an admissible transition x → y is realized by a heteroclinic connection $\gamma_{xy}$ from the unstable manifold of $s_x$ into the stable manifold of $s_y$. This makes the scheduling hypothesis more explicit than the generic claim that parsing traverses 'states': the algebra fixes which workspace states and transitions are admissible, while the dynamical system supplies dwell times, sensitivity to perturbation, and the probability or reliability of taking each permitted exit. The mapping also makes clear why phase-space dynamics is not yet syntax under the NAP: an arbitrary heteroclinic network is only a cognitive sequencer. It becomes a *syntactic* sequencer when its metastable states and connections are equivariant with the formal workspace graph.

This realization yields a further prediction about the local geometry of derivational choice. The branching entropy H(x) defined in §4.8 is not identical to Rabinovich and colleagues' information-flow capacity, nor to any single eigenvalue statistic. Nevertheless, if outgoing derivational alternatives are implemented as dynamically available escape directions from $s_x$, workspaces with richer branching should possess a correspondingly richer or less strongly dominated unstable eigenspectrum. One simple observable is the dimension of the unstable manifold,

$$\mathrm{dim}E^u(s_x) = \left|\{j\colon \mathrm{Re}\lambda_j(s_x) > 0\}\right|.$$

More generally, the relative positive or least-stable eigenvalues should constrain the accessibility and dwell-time distribution of competing exits. Thus, after controlling for state occupancy and generic difficulty, the NAP predicts a systematic relation between formal branching H(x) and the locally estimated stability geometry of the corresponding neural state. This is a stronger claim than merely finding more state switching in syntactically difficult conditions.

This prediction should be read together with the stability requirement on heteroclinic channels. In the canonical construction, a sequence of metastable states is reproducible only if each saddle has a one-dimensional unstable manifold and a saddle value greater than one,

$$\nu_i \;=\; \frac{|Re\;\lambda_s^{(i)}|}{\lambda_u^{(i)}} \;>\; 1,$$

where $\lambda_u^{(i)}$ is the unstable eigenvalue of the $i$-th saddle and $\lambda_s^{(i)}$ its stable eigenvalue of smallest modulus (Afraimovich et al., 2004; Rabinovich et al., 2008). A saddle with $dimE^u(s_x) \;\geq\; 2$ – which is how Rabinovich and colleagues model a decision node – purchases branching at the expense of the reproducibility of the channel that passes through it. The branching prediction above and the dynamical-stability bound $d_{sequencer}$ we will introduce immediately below are therefore one claim seen from two sides: derivational choice is realized as a local loss of channel stability, and the number and depth of such nodes a trajectory can traverse while remaining recoverable is what bounds it.

A terminological distinction is important here. "Heteroclinic binding" in the prior dynamical-systems literature refers primarily to the coordination of parallel sensory or intrinsic information streams into coupled metastable sequences. That is a useful mechanism for multisensory cross-stream coordination, but it is not the binding operation at the core of linguistic Merge: coordinating two evolving streams does not by itself create a new recursively reusable syntactic object, preserve non-associative constituency, or enforce the order-blindness required of the completed

Merge state. In the present architecture, SHCs are therefore a candidate realization of scheduling and coordination, whereas Meld remains the candidate realization of the binary composition gate.

Finally, an unbounded competence operation can admit several independently bounded physical realizations, with different limits arising from different physiological constraints. The phase-address code supplies a carrier-ratio bound; the workspace store supplies a memory-capacity bound; and a CTLN or SHC sequencer supplies a dynamical-stability bound on how long or how richly branching a trajectory can remain reliably recoverable (Bick & Rabinovich, 2009, 2010). Writing each $d$ below for the maximum depth supported by that subsystem, the observable behavioral ceiling should therefore satisfy

$$d_{\text{effective}} \leq \min\left(d_{\text{phase}}, d_{\text{workspace}}, d_{\text{sequencer}}\right).$$

This decomposition may be experimentally useful because the three failure modes need not covary. Phase aliasing, loss of workspace addressability, and instability of the derivational trajectory can each impose a performance ceiling even when the formal competence operation is unchanged.

Meanwhile, bipartite expander Hopfield networks can feasibly address a central puzzle at the heart of the language sciences: How can a finite biological network support an enormous repertoire of distinct, robust, recombinable linguistic states? (or, in classical terms, the “infinite use of finite means”). The answer provided by the expander construction is to define a compact family of local constraints whose simultaneous satisfaction licenses a combinatorially large set of global states (Burns & Fukai, 2023; Chaudhuri & Fiete, 2019; Yampolskaya & Mehta, 2026). But under the NAP as we are developing it here, two neural states are not adequate syntactic representations merely because they uniquely distinguish two trees; their geometry and dynamics must support the relevant operations. In an expander interpretation, a ‘codeword’ for $[[A\ B]\ C]$ must make $[A\ B]$ addressable if later computation requires it. Expander networks are therefore strongest when construed as *structured factor codes*, not opaque labels.

Nevertheless, bipartite expander Hopfield networks remain an idealized, largely equilibrium-based model: their cortical realization is unestablished, their constraints are not developmentally learned, their nearest-codeword repairs may ignore contextual interpretation, and standard attractor cleanup does not itself implement coproduct extraction or re-Merge.

As such, given their respective strengths and weaknesses, it may be necessary to jointly invoke these models in a scale-dependent fashion. For example, at *fast timescales*, expander-like dynamics can stabilize and correct the current candidate object. Let $\tilde{c}_t$ denote the noisy or provisional code presented to the associative store, and let $c_{W_t}$ denote the corrected codeword representing workspace $W_t$. Below, the first arrow is error correction within the current workspace state:

$$\tilde{c}_t \xrightarrow{\text{constraint correction}} c_{W_t}.$$

At a *slower timescale*, graph-structured competitive dynamics (e.g., CTLN or SHC/WLC dynamics) schedule the next workspace operation:

$$c_{W_t} \xrightarrow{\text{select / Merge / label / extract}} \tilde{c}_{t+1} \xrightarrow{\text{stabilize}} c_{W_{t+1}}.$$

The total parser becomes:

$$c_{W_0} \to c_{W_1} \to c_{W_2} \to \cdots,$$

with each state robustly stabilized but each transition sensitive to input and context (Figure 11C).

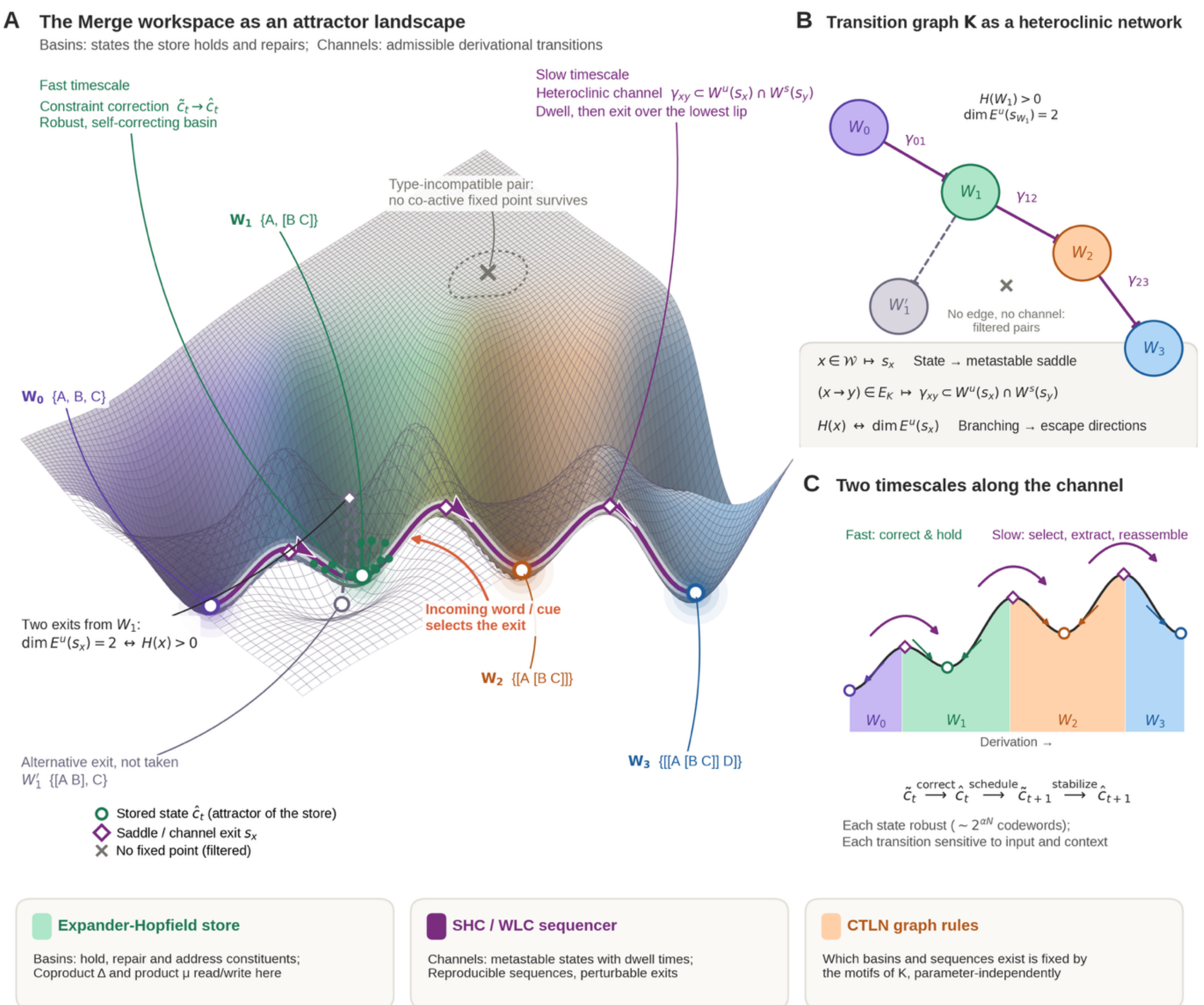

**Figure 11. The Merge workspace as an attractor landscape.** (A) Each basin is a state $\hat{c}_t$ of the content-addressable store (stable at the fast timescale). At the derivational timescale it behaves as a metastable saddle $s_x$ whose lowest lips are its unstable directions $\gamma_{xy}$. The trajectory $W_0 \to W_1 \to W_2 \to W_3$ is a heteroclinic channel: reproducible under noise, redirectable by input. $W_1$ has two dynamically available exits ($dimE^u(s_x) = 2 \leftrightarrow H(x) > 0$); the incoming word selects between the channel to $W_2$ and the admissible alternative $W_1{}' = \{[A\ B], C\}$. Noisy codes $\tilde{c}_t$ converge on the $W_1$ floor (constraint correction within the store). A type-incompatible pair has no surviving co-active fixed point (dashed ring, ×), the graph-rule realization of the colored-operad filters (§4.5). (B) The formal transition graph $K$ and its equivariant mapping onto a heteroclinic network: states to saddles, admissible transitions to channels, branching entropy to the dimension of the unstable manifold. (C) The energy profile along the channel separates the two timescales: fast correction and holding within a basin (expander-Hopfield store, $\sim 2^{\alpha N}$ robust codewords) and slow scheduling over lips (CTLN or SHC/WLC sequencer). The binding gate $B$ (Meld, §4.9.1) and the seal (§5.3) act at the transition and are not part of the landscape.

The complementarity between CTLNs and bipartite expander Hopfield networks aligns with the decomposition of the workspace transition operator, K (§4.7-4.8). The coproduct Δ and the product μ (i.e., extraction of an accessible subterm, and reassembly of the forest) together with maintenance of the workspace itself are the operations for which a high-capacity, self-correcting associative memory is the appropriate substrate; the read-only label register of §5.3 is a sealed instance of the same store. Consider the transition dynamics: K read as a directed graph over workspace states, with the local branching entropy H(x) rising as an incoming word opens attachment sites and falling as filters close them. This is a regime that graph-structured transient dynamics can naturally occupy, with graph rules determining the fixed points of the whole from those of component subgraphs. The one operation neither construction supplies is the binary gate *B*, the non-associative combination itself, which remains the composition gate of §4.9 and §4.9.1; i.e., the corrected Rényi mixture with Meld built around it.

The admissible substrate is therefore *a division of labor*: an associative store to hold and retrieve constituents (i.e., bipartite expander Hopfield networks), graph-structured transient dynamics to sequence the derivation (i.e., a CTLN or stable heteroclinic channel), and a nonlinear gate to bind them (i.e., Meld, §4.9.1).

Two of the algebraic properties we discussed above immediately acquire concrete dynamical signatures under this reading. The exocentric head-selection case (§4.6), whose hypermagma output is set-valued, is a multistability-and-competition problem of the kind CTLNs natively model: two candidate labelings correspond to competing attractors, and head selection is symmetry-breaking into one of them under lexical or interface bias – predicting a metastable dwell followed by a stable state rather than a generic difficulty response. The colored-operad filters (§4.4-4.5) correspond to the permitted/forbidden co-activation structure of such networks – established for symmetric threshold-linear networks by Hahnloser and colleagues (Hahnloser et al., 2003) and carried over to the non-symmetric CTLN case by the fixed-point supports $FP(G)$ of Curto and colleagues (Curto et al., 2019): incompatible type combinations are configurations with no surviving co-active fixed point, and the pullback prediction $I_{filter} \neq 0$ becomes the claim that combined filters eliminate fixed points super-additively rather than additively, offering a graph-rule realization of the fiber-product structure of §4.5.

In summary, the reason to entertain multiple attractor architectures is well-supported within the psycholinguistic profile of compositional meaning. Written out over workspaces, Merge has multiple regimes: hold the constituents already built so they remain accessible for further manipulation/interpretation; decide which operation to perform next; bind two objects into one. A content-addressable store may be the right substrate for the first, because it must be robust to noise and must hold combinatorially many distinct states. A graph-structured attractor system may be the right substrate for the second, because it must sequence. Importantly, *neither* can supply the third (i.e., the unique properties of linguistic binding), which remains the role of the composition gate – Meld (§4.9.1). One obvious consequence is a prediction about how the language system breaks in different ways; e.g., a system can have an excellent memory for constituents and an orderly derivational schedule and still fail to compose in ways that lead to conceptual interpretations. Hence, any theory that names only one such mechanism or neural measurement ‘syntax’/‘semantics’ has no way to satisfactorily express that.

## 7. Preliminary Predictions

In the previous sections, we have outlined what we refer to as a Neural Admissibility Program (NAP) for the neuroscience of language. NAP does not treat formal linguistic structure only as a source of stimulus contrasts, nor does it identify composition with any single neural signature. Rather, it treats formal properties as admissibility constraints. Yet, the most informative experimental programs in the cognitive neurosciences use constrained experiments to identify operations, and naturalistic experiments to test whether the same operations generalize to live use cases. In order to move NAP further towards testability, the prediction families below build directly on the bracketing-dissociation and depth-address predictions already stated above (e.g., neural activity at specific scales should distinguish bracketings at the point where the grouping becomes relevant, even when lexical content and serial order are matched as far as possible). At the most general level, locally clustered or interdigitated unit codes can hypothetically support a scale-general code, whereas dependence on population/field codes at higher levels of linguistic complexity would support scale-dependent, distributed-state accounts in which higher-order meaning is not recoverable from isolated unit firing alone.

### *7.1. Workspace extraction and reactivation*

The Hopf-algebraic workspace view predicts that constructed constituents remain addressable, because the coproduct $\Delta(T) = \sum_v F_v \otimes T/F_v$ makes substructures available for extraction. Future experiments should therefore test whether a phrase can be decoded or reactivated when it is later retrieved for agreement or semantic interpretation. Delayed reactivation should be interpreted as a formal requirement for extracting an accessible term. Direct cortical signatures would include renewed high-frequency activity at sites that encode the original constituent, spike-field coupling that changes when the constituent is retrieved into a new structural role, and directed frontotemporal interactions during structural reanalysis (e.g., syntactic disambiguation).

Elaborating on these issues, recent work suggests that cross-frequency coupling may also be spatially organized as a traveling phase field rather than instantiated independently at each cortical location (Chen et al., 2026; Davis et al., 2020; Woolnough et al., 2022). Under this extension, local high-frequency bursts carry constituent content, while the direction and velocity of a lower-frequency wave determine the order in which cortical populations become excitable and exchange that content. Coproduct extraction, composition and reassembly should therefore be associated with distinct phase-gradient configurations rather than with a generic increase in PAC. For example, posterior-to-frontal waves may accompany selection and projection, whereas recurrent or reversing waves may accompany reanalysis and workspace reopening.

### *7.2. Composition laws as explicit neural models*

The function-space account is useful precisely because it makes rival neural models explicit and comparable on the same data. Given neural encodings $u$ and $v$ of two lexical or phrasal objects, a number of composition laws can be compared (Figure 9),

$$G_{\mathrm{add}}(u, v) = u + v, \qquad G_M(u, v) = M(u + v), \qquad G_{\mathrm{rf}}(u, v) = R_L u + R_R v,$$

$$G_{\mathrm{RNN}}(u,v) = \sigma(W_u u + W_v v + W_h h_t), \qquad G_{S_2}(u,v) = u \oplus_{S_2,\beta} v,$$

where the entropy gate uses the second Rényi entropy $S_2(p) = -\log \sum_i p_i^2$ established in §4.9. Here, $M$ is a learned map applied after additive combination; $R_L$ and $R_R$ are fixed or learned role operators for the two daughters; $\sigma$ is the recurrent nonlinearity; $W_u$, $W_v$, and $W_h$ are weight matrices of compatible dimensions; and $h_t$ is the recurrent state available at the current composition step. All five laws are understood to map into the same comparison space. We note that the above form of $G_{S_2}$ represents the *unnormalized form*. The corrected gate that we discussed previously is as follows,

$$G_{S_2}(u,v) = \lambda^\star u + (1-\lambda^\star)v,$$

and the law that we recommend, Meld, passes that mixture through shared saturating synapses: *Meld*(*u*, *v*) = tanh(*κW*[*λ⋆u* + (1 − *λ⋆*)*v*]) (§4.9.1).

The additive law is usually too weak because it is associative and therefore cannot preserve non-associative constituency (Murphy, 2020a); the recurrent law preserves bracketing but, with separate weights for the two daughters, leaves their order in the composite (§4.9.1); the ordered role-filler law can preserve ordered role-filler structure, especially when roles are explicitly encoded, but it tends either to over-specify order or to expand dimensionality. Meld is privileged here because it is a specific commutative, non-associative, disagreement-sensitive operation that can be tested under explicit controls, and the entropy mixture inside it is the part that carries the content-dependence.

Comparing composition laws also supplies a direct, data-level test of the equivariance requirement of §4 – the demand that the encoding square N∘M≈G∘(N×N) commute. We can fit a *composition decoder* Ĝ that predicts the neural state of a composed object from the neural states of its daughters, Ĝ(N(A), N(B)) ≈ N($\mathcal{M}$(A,B)), using any of the candidate laws above as the functional form of Ĝ. Equivariance then makes a graded, cross-condition prediction: a decoder fit on one set of daughter pairs should generalize to held-out pairs, and should generalize better under the correct grouping than under a mismatched grouping of the same terminals. On matched-terminal bracketing contrasts (§5.1.1), the transfer accuracy of the composition decoder should therefore separate the two parses, whereas an encoding that merely correlates with linguistic structure without implementing it will show no such grouping-specific transfer. This turns the commuting square into a concrete RSA/cross-decoding protocol that can be run on neural data. Lastly, we note that an associative substrate is not ruled out merely because its primitive operation is associative. It is ruled out only if the complete system fails to preserve, reconstruct, or causally deploy the relevant non-associative distinctions.

The simulations we reported in §4.9.1 illustrate this comparison. On bracketing, the role-filler law separates the two bracketings far more sharply than the entropy gate at the level where the contrast is formed (*d* = 1), and degrades less with depth (0.972 at *d* = 5). The second fact is largely per-level gain – each law attenuates the deepest triple differently as it is carried to the root, and the noise is scaled to the whole response – and should not be read as a statement about the algebra. The first fact is not gain, and as §4.9 anticipated, neither is decisive. The role-filler law retains a distinction the magma quotients away, and composed along a path its role operators tag each terminal with its full address in the tree rather than merely its depth. Given matched terminal order, depth-tagging is bracketing (see below), so the ordered code over-determines the target and its advantage on this axis is purchased by representing what the competence object does not contain. We make no claim here of information-theoretic dominance: both laws map $\mathbb{R}^d \times \mathbb{R}^d \rightarrow \mathbb{R}^d$, neither encoding is faithful to its inputs, and neither is a sufficient statistic for the other; the ordering reported here is an empirical fact about these encodings at matched SNR. What matters is that a law can win the bracketing race while being inadmissible before we even evaluate success. A model comparison conducted on bracketing alone therefore selects against the gate on grounds that are independent of what the brain does.

What separates the laws is *commutativity*, and it does so categorically rather than in a graded manner. Repeating the analysis with daughter order as the classified variable on an otherwise identical spine, the gate and its offset-corrected variant are at chance at every depth (0.494-0.506), while the role-filler and recurrent laws reach 1.000 and 1.000 at $d = 1$ and remain far above chance at $d = 5$ (0.891 and 0.701). The gate's null is not a floor effect: the same axis, on the same stimuli, at the same SNR, rejects both rivals at every depth tested. The two axes are therefore not two measurements of one quantity at different sensitivities. Of course, on neural data this two-axis test must be run inside some kind of pre-registered subspace to be falsifiable: one might fix the constituency-preserving subspace $V_{syn}$ on held-out data by bracketing decoding (§6.3), freeze it, and only then test daughter order inside it.

There is a further reason not to place too much weight on bracketing decodability as such. For ordered binary trees the leaf-depth sequence determines the tree. Under matched terminal order, bracketing decodability is therefore *equivalent* to depth-tagging decodability, and is achieved by any code that tags each terminal with its depth and nothing more. For parses of a fixed string, then, the static depth-address code of §5.1 is not a weaker hypothesis than a grouping code, since the string supplies the order and the depth map then fixes the tree. As a code for the competence object of Merge it is weaker (§5.1): where two syntactic objects share a content-to-depth map, as the balanced four-leaf pair does, the static map is silent and the sealing sequence of §5.3 marks the distinction. The phase-address proposal is well posed on the first reading, and its grouping content on the second is time-resolved.

This concern about neural decodability is not merely methodological: Posani and colleagues report that across 43 cortical regions, once non-independent conditions are merged, near-maximal linear separability is the rule rather than the exception, such that above-chance decoding of a given dichotomy is close to uninformative about what a region computes (Posani et al., 2026) – potentially including, in the present case, a dichotomy corresponding to an a priori inadmissible composition law.

In summary, one might intuitively assume that the natural thing to do with rival composition laws is simply to orchestrate a competition and then back the winner. Yet, the role-filler law wins on bracketing because it retains a distinction the grammar discards – which daughter occupied which slot – and its role operators, composed along a path, tag each terminal with its address in the tree, so bracketing falls out for free. A law that recovers grouping by representing more than the competence object contains has not thereby become a better model of it. The comparison that more helpfully discriminates is a conjunction. Of the laws that carry any grouping information, the entropy gate, fixed mean-pooling, the tied-weight laws, saturating or linear, and Meld remain blind to daughter order. The role-filler and recurrent laws are rejected on the second axis at every depth tested, on the same stimuli and at the same signal-to-noise ratio; mean-pooling is separated from the entropy mixture by the balanced four-leaf contrast and by the content-dependence of its mixture (§4.9.1, §9.2), a sum-only saturating unit is separated from it by disagreement sensitivity, and Meld – which has both and decodes bracketing to within a few points of the role-filler law at every depth – is the member of the class we propose (§4.9.1). Hence, any neural code that recovers both grouping and order is representing at least some feature that natural language does not contain – which disqualifies it only if that feature is never subsequently discarded – while a code that recovers neither is not representing language at all.

The same reasoning applies to the other axis. A non-commutative gate whose output is symmetrized later also yields an order-blind composite, so the order null does not on its own identify the gate. What separates them is whether order is ever jointly decodable with bracketing in the state the next Merge step consumes, and whether it leaks back under load, since symmetrization can be done badly and commutativity cannot. The null also needs a sensitivity check: $V_{syn}$ is fit with terminal order matched, so order should first be shown decodable in that frozen subspace using a contrast that varies it.

### *7.3. Naturalistic structure beyond feature fitting*

For analyses of naturalistic language data, in addition to coding each word with a scalar node count, we should also fit models that distinguish punctuated moments of linguistic operations: composition, closure, headedness, semantic role establishment, locality boundaries, and workspace transition. Lexical, semantic, acoustic, prosodic, surprisal, dependency, and task variables enter as controls, not as *substitutes for structure-building hypotheses* (Slaats & Martin, 2025). One option here is to align parser-derived workspace states with neural state-space models, modeling transitions among latent states corresponding to building, filtering, maintaining, retrieving, and closing.

Relatedly, the existing single-cell literature on syntax (Cai et al., 2026; Lakretz et al., 2026) that often uses naturalistic stimuli does not yet demonstrate that single units encode higher-order syntactic structure in the sense that would matter for a mechanistic theory. What it demonstrates is that firing rates are weakly separable by parser-derived labels that are themselves collinear with derivational position, articulatory planning, and memory load, and that this separability is small in effect size, and confounded by the demands of production (in the case of Cai and colleagues). These results are consistent with the prediction made here that single units supply lexico-semantic and categorial atoms while the higher-order algebraic operations of syntax are enforced by coordinated, multiscale dynamics. The appropriate response is not to deny that such neurons carry syntactically relevant information, but to highlight that carrying information about a syntactic constituent is not the same as implementing the operation by which the constituent was constructed.

### *7.4. Alternative formalisms*

The algebraic formalism presented here is not the only useful mathematics for neural mechanism discovery. Tree-adjoining formalisms offer explicit derivational objects and parsing algorithms (Momma, 2023); categorial grammar makes the syntax-semantics mapping explicit (Grefenstette & Sadrzadeh, 2015); tensor-product and vector-symbolic models address role-filler binding (Dehaene et al., 2022; Kanerva, 2009); sheaf-theoretic and categorical constructions model local-to-global consistency (Hansen & Ghrist, 2019); dynamical-systems and control-theoretic models capture state transitions and perturbation responses (Breakspear, 2017); and probabilistic and neural-network grammars connect structure to processing difficulty (Levy, 2008). The reason to foreground the algebraic framework we have chosen here is that it unifies, under one operation and a single space of objects, several requirements that are usually treated separately. This breadth is what makes it especially suitable for constraining neurobiological implementation, because the same formal vocabulary generates predictions at every scale considered in §6.

### *7.5. Falsification*

Our framework would be weakened if algebraically defined contrasts never outperform lexical, semantic, prosodic, dependency, or surprisal controls across sufficiently-powered datasets. It would be weakened if grouping history cannot be recovered from any neural scale once confounds are controlled. It would be weakened if semantic role, locality, and headedness violations always collapse into a single nonspecific difficulty signal. It would be weakened if spike-field, low-frequency, and inter-areal measures add no explanatory or causal value beyond local firing rate and high-frequency power.

While some of our proposed admissibility conditions can admittedly be satisfied by various compositional systems (e.g., operations should dissociate), the predictions that distinguish our implementation of the NAP from weaker hierarchical baselines are:

(i) the pullback interaction $I_{filter} \neq 0$, with its sign as a further, realization-specific prediction: an additive-latent account read out through a compressive nonlinearity can deliver only $I_{filter} \leq 0$ (§4.5), so a reliably super-additive interaction excludes that family, whereas an expansive readout is not excluded by anything in §4.5 and can produce $I_{filter} > 0$ without a pullback. The positive sign is therefore the prediction of the attractor realization of §6.7 specifically, and is diagnostic of the pullback only where the readout's curvature has been estimated independently, for instance from the single-violation dose-response;

(ii) the depth-aliasing signature of an angular phase-address code – categorical confusions between depths $d$ and $d + d^*$, and non-monotone rather than graded failure past the wrap-around bound, which is testable only where the depth ladder exceeds the measured carrier ratio $d^* = f_\beta / f_{\delta\theta}$ (§6.5);

(iii) the composition signature of §7.2 – bracketing decodable and daughter order at chance, on matched terminals, which no additive, role-filler, or recurrent law satisfies, completed by a bracketing-dependent leaf mixture (§9.2), which fixed mean-pooling does not deliver, and by sensitivity of the composite to daughter disagreement at fixed daughter sum (§4.9.1), which a saturating sum-only unit does not deliver and Meld does – together with the parametric prediction that the size of the bracketing effect scales with the distance between the outer terminals and is invariant to the re-attached one (§4.9.1), with the order null read as a claim at every latency, and against a demonstrated positive on the order axis;

(iv) the mutual-coherence geometry decomposition $\mu_{syn} < \mu_{sem}$ (§6.3);

(v) the phase-address-vs-depth-count dissociation (§5.1.2);

(vi) a nonzero neural associator recovered by decoder transfer, read as a within-depth contrast between matched bracketings at a fixed embedding depth rather than as a divergence from an additive null that grows with depth – the latter being, as §4.9.1 shows, bracketing-blind and therefore uninformative (§7.2, §4.9.1).

(vii) the phase-dispersion contrast between the two phase codes – a collapse of circular dispersion across constituent-carrier phases at constituent completion, as the Marcolli-Berwick synchronization model requires, versus growth of dispersion with the number of occupied depth addresses, as our §5.1 code requires – measured on a parametric depth manipulation with matched terminals, and estimated at the $n{:}m$ locking ratio measured per site rather than at $1{:}1$ (§4.10).

For Meld specifically, this proposed binding law makes three predictions that direct cortical recordings can test. All three require the constituent codes to be estimated separately from the composite: each word is presented alone and its population vector – e.g., broadband γ across a language network region, or unit firing rates – is taken as its constituent state, so that for any two-word phrase the daughters *u*, *v* and the composite are three measured vectors in one space. Bracketing contrasts use height- and terminal-matched pairs such as ((A B) C) versus (A (B C)); order contrasts use (A B) versus (B A). First, within a constituency-preserving subspace fixed from the single-word responses (§6.3), bracketing should decode from the composite and daughter order should be at chance; order decoding there rules out every commutative law, Meld included. Second, the leaf-mixture readout that we will introduce in §9.2 should show bracketing-dependent effective dimensionality, which fixed pooling cannot produce. Third – and among the order-blind laws only Meld predicts this – the composite should depend on the disagreement between the daughters at fixed daughter sum: from the single-word set, select word pairs whose summed vectors are matched in the subspace but whose pointwise difference is not, and test whether their composites differ. A sum-only circuit predicts no difference; Meld predicts one, and at the single-unit level predicts its exact form – target neurons receiving both constituents integrate sublinearly, tracking the smaller input where they disagree, and saturate. Spike-LFP coupling then locates the stages: constituents co-active in a common γ window at the operation (i.e., the 'O' level of ROSE (Murphy, 2025)), and the composite acquiring the δ-θ phase address of the seal (S).

On a similar note, how might we respond to the charge that we are merely cherry-picking our favored neural signatures? We must ensure that a neural signature is admitted as implementing an algebraic operation only to the extent that it tracks that operation's *defining contrast* – the distinction the algebra requires be preserved (bracketing for non-associativity, extraction for the coproduct, etc.) – under matched lexical, surprisal, and other controls. A signature that varies only with a coarse, monotone, or generic quantity (sentence vs. wordlist, open-node count, etc.) is demoted to correlate or consequence, irrespective of its frequency band, anatomical source, or recording scale. For example, a phase or PAC effect that indexes only sentence vs. wordlist (§4.10) would be demoted by exactly the criterion that demotes the broadband $\gamma$ ramp (see §9.2).

Lastly, while we have criticized LLM-driven approaches for providing little direct evidence about underlying algorithms or mechanisms, we acknowledge that our approach in §4 to finding a physiological G for which $N \circ M \approx G \circ (N \times N)$ faces a similar problem: a commuting square shows the code preserves the relevant distinctions, not that the system computes Merge rather than some other operation preserving them. The criterion is therefore necessary, not sufficient, for claims about implementing Merge, but the alternatives it admits fail in instructively different ways; see also related discussion by Krakauer and Ramsey (Krakauer & Ramsey, 2026). An associative encoding with sufficient bookkeeping (e.g., a monoid over a bracket-augmented alphabet, as in a Dyck or stack encoding) can recover bracketing, yet only by representing structure indirectly over strings plus auxiliary symbols; it is excluded not by the square but by the prior requirement that the map be stated over the explanatorily correct objects – workspaces of binary-branching sets. Meanwhile, §6.1 predicts that a perturbation of the depth-addressing dynamics of §5 should degrade genuinely non-associative, structure-dependent processing while leaving a system that merely maintains an associative bracket-tally comparatively intact – a dissociation no descriptive commuting square can adjudicate.

### *7.6. Cross-scale consistency: One computation, multiple realizations*

A final prediction follows from the NAP's central commitment that a *single* algebraic object underlies signatures at every recording scale, even though the framework predicts scale-*dependence* in how strongly each operation is expressed. If the grouping recovered from single-unit and ensemble geometry (§6.3), from spike-field coupling (§6.4), from low-frequency phase and PAC (§6.5), and from directed inter-areal dynamics (§6.6) are all realizations of the same underlying derivation, then on matched-terminal bracketing contrasts they should *agree on the same grouping on a trial-by-trial basis*. The parse decoded at one scale should predict the parse decoded at another, above what shared stimulus and task variables explain. Cross-scale consistency is thus the integrative test that distinguishes the NAP from a looser view on which each scale merely correlates with linguistic structure in its own right; we suggest it as a unique discriminating experiment the program affords, and a natural organizing target for multi-scale (unit, LFP, inter-areal) recordings.

## 8. The Algorithmic Middle Layer: Incorporating Real-Time Parsing

The present algebraic formalism is *inherently atemporal*: syntactic objects are elements of a free commutative non-associative magma, and linear order is treated as a quotient imposed only at externalization. An intervening *algorithmic* layer – the Marrian level between the computational characterization of §§3-4 and the implementational hypotheses of §§5-6 – is therefore required to state which operations the system is attempting at each moment. Minimalist and left-corner parsing (Berwick & Stabler, 2019; Pasternak & Graf, 2021; Stanojević & Stabler, 2018), cue-based retrieval (McElree, 2000; Vasishth et al., 2019), probabilistic prediction (Brennan & Hale, 2019; Hale, 2001), and state-space dynamics (Buonomano & Maass, 2009) are all candidates for this layer. In this section, we briefly develop this layer as a real-time realization of the Hopf-algebra Markov chain of §4.8.

Parsing is not a literal inverse of externalization, but an inference process over the preimage of the externalization map: the parser selects among structures compatible with the unfolding signal under grammatical, contextual, prosodic, and memory constraints. We adopt here a *qualified* one-system view (Lewis & Phillips, 2015): the parser deploys the grammar's operations and representations directly, but the order in which operations are scheduled, the expectations it holds open, and its memory regime are not fixed by the competence algebra. The strong one-system thesis, on which the grammar simply *is* the parser, claims more than our commitments license: an order-free commutative magma cannot by itself be an order-driven incremental parser, because the parser's defining property – its use of linear precedence – is precisely the information the magma quotients away. What we defend is *transparency of operations and representations* together with an explicit, non-trivial scheduling-and-memory layer; the near-identity of grammatical and parsing operations – Berwick and Weinberg's type transparency (Berwick & Weinberg, 1984; Stabler, 2013) – is treated here as a substantive, empirically consequential hypothesis. The commutativity requirement of §3.1 is located at this layer: the parser may carry linear precedence in its working state, but the composite it hands to the next Merge step must not, and it is that hand-off (not the parser's input) on which the order-blindness test of §7.2 is defined.

### *8.1. Separable Predictions for Parsing*

During language comprehension, held predictions and stored features are nothing other than the *addressable, maintained workspace objects* of §4.7 and §6.6. Prediction is thus not a secondary faculty adjoined onto Merge; it is the sustained maintenance of an open workspace object whose selectional requirement is not yet discharged. This predicts a *temporal and physiological dissociation* between a predictive/maintenance signature and a compositional signature, with structural reanalysis/disambiguation realized as re-entry into a prior workspace state via the coproduct (§4.7). Recent MEG decoding already provides convergent evidence: syntactic features indexing structure held open for future completion are selectively sharpened by local prior knowledge and are strongest at word onset, whereas features indexing the resolution of syntactic relations show no such predictive benefit and are strongest at word offset (Iaia & Tavano, 2026).

Consider a minimal worked example: the structure 'The red boat sank', contrasted with a word list, a Jabberwocky variant, and a syntactically well-formed but semantically anomalous sentence. The parser traverses this structure as follows:

*the* : project DP; hold an NP-complement expectation $E_1$ (no Merge)
*red* : extend the predicted NP; predict its N head (no Merge)
*boat* : M(*red*, *boat*); yield NP, then M(*the*, NP); yield DP [$E_1$ discharged] (2 Merges)
*sank* : M(DP, *sank*); yield TP; θ: Theme = subject DP; transfer (integrate)

Under the particular left-corner schedule adopted here – or under a top-down implementation with the same relevant completion points (Chesi, 2023) – each Merge is discharged at left-corner completion. The held predictions and maintained constituents this requires are precisely the addressable workspace objects of §4.7 and §6.6; the parser adds a control regime and resource bounds, not new representational content.

Each step makes a separable prediction (Figure 12). Lexical access should engage local high-frequency and unit/ensemble codes for lexico-semantic features. The held expectations at 'the' and 'red' predict a *maintenance* signature – sustained low-frequency and spike-field coordination of an open object – rather than a composition response. The arrival of 'boat' predicts a *composition burst* as the two pending Merges discharge at left-corner completion, yielding reusable phrase-level states because the magma requires that ‘red’ has been grouped with ‘boat’ rather than summed into a bag of features. Label and head selection via PAC is weak here, since ‘boat’ projects endocentrically. Colored filtering predicts that the contrasts *dissociate* rather than collapse into a single difficulty signal: the semantic anomaly spares grouping and labeling while disrupting interpretive transfer, and combined filter violations interact non-additively (the pullback term of §4.5). The subject DP must then be *maintained* across the verb, predicting continued low-frequency and spike-field coordination until theta-role discharge, with reactivation via the coproduct if later material forces retrieval or reanalysis. Transfer predicts directed frontotemporal coordination.

The algorithmic middle layer
1 · Linear input (time →)
TP
DP
NP
The D
red A
boat N
sank V
Real-time schedule: input order ≠ derivation order (time →)
The $t_1$
red $t_2$
boat $t_3$
sank $t_4$
Predict DP Expect NP
Expect N (Complete NP)
○ held
○ held
M(red, boat) → NP
M(the, NP) → DP
2 Merges apply
M(DP, sank) → TP
θ-filter ✓
Transfer
Bottom-up Merge discharges at left-corner completion; predictions are held across the input. The §4.8 chain runs over parser states, not word order.
2 · Workspace operation log
Lexical access atoms of $SO_0$ enter workspace
Merge M(red, boat) → object, close
Label / head head = boat (N projects)
Merge M(the, [red boat]) → DP
Merge M(DP, sank) → TP
Colored filter θ: Theme → DP ✓ (operad check)
Transfer → Semantic interpretation
Access Merge Label Filter Transfer
Maintain: coproduct keeps [DP] addressable while sank integrates
3 · Predicted neural signature (after lexical, prosodic, surprisal and dependency controls)
Lexical access Local unit / ensemble & high-γ codes for category and semantic features
Composition Bracketing decodable; reusable, reactivatable phrase-level state; entropy gate > additive with depth (§4.10)
Maintenance Low-frequency phase + spike-field coupling hold the object addressable (§4.7)
Transfer Directed frontotemporal coupling + closure dynamics, not one ‘ramp’ (§7.6)

**Figure 12. The algorithmic middle layer applied to a worked example.** The real-time realization of the Hopf-algebra Markov chain of §4.8. Each operation licenses a distinct, separable empirical prediction. Because Merge is bottom-up while input arrives left-to-right, the determiner and adjective license only held predictions (open NP/DP expectations); the two DP-internal Merges discharge together at left-corner completion on ‘boat’.

The signature of Merge, on this view, is therefore not one regressor, region, or ramping motif, but a *structured trajectory through operations over a workspace* – and it is this trajectory, rather than any single annotated feature, that converts a syntactic description into a testable linking hypothesis.

Consequently, commutativity is not inherited by the real-time parser as temporal order-insensitivity. The parser must use linear precedence, prosody, and distributional cues to infer structure from an externalized string. What survives from the commutative magma is instead a *representational equivalence condition on the completed workspace object*. Once the parser has identified two objects as the daughters of a Merge operation, their abstract syntactic relation should not be reducible to the order in which they were encountered. Commutativity therefore migrates from an online transition constraint to a quotient/equivalence constraint on the composite the parser hands to the next Merge step (§3.1, §7.2).

An important precursor to this approach is the work of beim Graben and Potthast, which maps the discrete states of a left-corner parser onto continuous neural dynamics (Beim Graben & Potthast, 2012). Phrase-structure trees are encoded using filler-role tensor products and embedded in a Fock-space representation, while winnerless competition generates continuous trajectories between parser states that are subsequently realized in an Amari neural field. This provides an explicit demonstration of how symbolic parsing states can, in principle, be embedded in a continuous dynamical system. For our purposes, the NAP shifts the explanatory target: rather than beginning with a particular parser and constructing dynamics that reproduce its prescribed state sequence, the NAP asks which neural dynamics are admissible implementations of the formal properties that linguistic computation itself requires. Candidate mechanisms must therefore generate discriminating neural predictions rather than merely furnish a dynamical encoding of a predefined parse.

### *8.2. Summary of Linking Hypotheses*

We summarize here the core linking hypotheses that have presently been developed (Figure 13). Each formal property of language fixes an invariant that a candidate neural mechanism must preserve, and therefore maps onto a specific, falsifiable neural signature. A mechanism counts as an implementation only when this encoding is *equivariant* (the composition square commutes, $N \circ M \approx G \circ (N \times N)$), *faithful* (distinct bracketings and labels remain decodable), and *falsifiable* (the contrast is not explained by surprisal or lexical confounds alone).

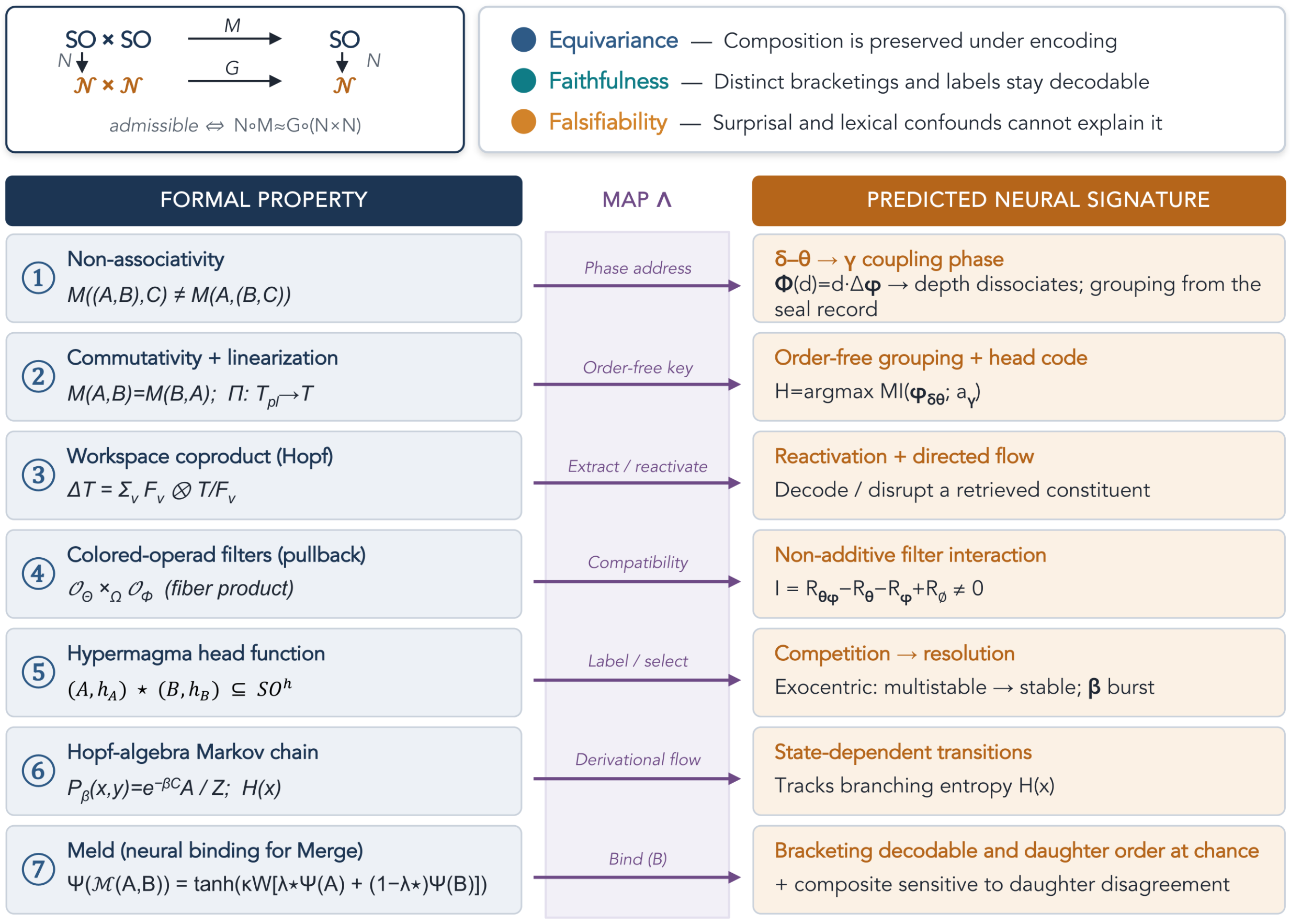

**Figure 13. Summary of linking hypotheses.** Each formal property of Merge imposes an invariant that maps, through an explicit linking map Λ, onto a predicted neural signature, subject to three conditions – equivariance, faithfulness, and falsifiability.

## 9. Further Candidate Mechanisms and Open Questions

### *9.1. Antipode-inspired cancellation as a candidate neural signature of Internal Merge*

One immediate direction to build off what we have established so far would be to explore whether the cancellation structure implicit in Hopf-algebraic models of Merge has a real-time neural counterpart. Antipode cancellation may thus constitute a neural signature of Internal Merge (e.g., involved in structures such as 'Which book did John say he read?'). Somehow, the brain must integrate 'which book' as the object of 'read' while preventing it from being interpreted as a second independent object. A plausible hypothesis is that real-time parsing approximates this competence-level operation through local quotienting updates. When a displaced constituent is retrieved at a gap, its lexico-semantic features should be reactivated, as many psycholinguistic models assume – but its residual structural position should *also* be suppressed, orthogonalized, or phase-opposed as an independent interpretive object. In this sense, Internal Merge would require not only memory for a filler, but a controlled cancellation operation that prevents copied material from being counted twice. Movement dependencies should produce paired signatures of reactivation and cancellation: renewed evidence for the displaced object, together with a temporally aligned reduction or opponent coding of the lower-copy position as an independent semantic contributor. Failures of this balance may help explain interference, reconstruction errors, or illicit double-counting of thematic roles. Standard cue-based retrieval models can explain why 'which book' is reactivated at 'read' (Lewis & Vasishth, 2005; Vasishth et al., 2019; Wagers et al., 2009), but this reactivation does not explain the requirement that the same constituent can be structurally present in more than one position while contributing only once to the relevant semantic dependency.

Collins provides a useful objection to the Marcolli-Chomsky-Berwick implementation of Internal Merge, arguing that it cannot capture reconstruction, since the lower occurrence survives only as a labeled leaf and is invisible to c-command at the interface (Collins, 2026). Within a content-addressable workspace, this objection dissolves into an interesting prediction: a label is an address, retrieval by address returns the full constituent, and binding at the lower position is computed by reactivation rather than by tree geometry. Reconstruction effects should therefore coincide with content reactivation at the gap site, as in filled-gap and antecedent-priming effects, which distinguishes a store that returns content from one that returns only an address.

### *9.2. Effective dimensionality and the order of the entropy gate*

Another follow-up from what we have established above could sharpen the one parameter our function-space gate leaves unmotivated: the Rényi order $\alpha$. The collision entropy in the gate is exactly the log participation ratio of the leaf-mixture, $S_2(A) = -\log \sum_\ell a_\ell^2 = \log D_{\text{eff}}(A)$, so the gate's diversity term is the *effective dimensionality* of the composing population; the same $d_{\text{eff}} = (\sum_k \sigma_k)^2 / \sum_k \sigma_k^2$ estimated routinely from population covariance (Gao et al., 2017). We conjecture that the composing population's variance spectrum realizes the leaf-mixture distribution, so that $\log D_{\text{eff}}$ estimates $S_2$. Two conditions are required for this. First, the per-atom variances $\sigma_\ell$ must be attributable to single atoms. This is not a claim that the covariance eigenvectors align with the atom directions – low mutual coherence bounds cross-talk between atoms but does not diagonalize the covariance in the atom basis – and the estimator below does not rely on any such alignment. It projects onto independently identified atom directions $e_\ell$ and never eigendecomposes. Low coherence (§6.3) enters only by bounding the cross-talk between those projections. Second – and this is a separate assumption about the population's second-order statistics, not a consequence of orthogonality – the signal variance along each atom's direction must scale linearly with that atom's mixture weight $a_\ell$ (linear in the weight, not squared), so that the per-atom variance spectrum inherits the leaf-mixture distribution and $D_{\text{eff}}$ = $(\sum_k \sigma_k)^2 / \sum_k \sigma_k^2$ collapses to $1/\Sigma_\ell a_\ell^2 = exp(S_2)$.

Linearity is not generic, and the gate we recommend does not supply it. The offset-corrected gate of §4.9.1 carries forward the deterministic superposition $\Psi(T) = \Sigma_\ell \, a_\ell \, \Psi(\ell)$, so along an atom direction $e_\ell$ the composed state has amplitude $a_\ell$ and therefore power $a_\ell^2$. The single-trial squared projection onto $e_\ell$ is then exactly $a_\ell^2$. Define the normalized squared-amplitude distribution by

$$\hat{p}_\ell := \frac{a_\ell^2}{\sum_j a_j^2},$$

and, for $q > 0$, $q \neq 1$, define the Rényi entropy of the leaf-weight vector $A = (a_\ell)_{\ell=1}^n$ as

$$S_q(A) := \frac{1}{1-q} \log \left( \sum_{\ell=1}^n a_\ell^q \right).$$

Thus, $S_2(A)$and $S_4(A)$ are respectively the order-two and order-four Rényi entropies of the same leaf-weight vector, and

$$\mathrm{PR}(\hat{p}) := \frac{1}{\sum_{\ell=1}^n \hat{p}_\ell^2} = \frac{\left(\sum_{\ell=1}^n a_\ell^2\right)^2}{\sum_{\ell=1}^n a_\ell^4} = \exp\ \left(3S_4(A) - 2S_2(A)\right),$$

not $\exp\left(S_2(A)\right)$. This is the fixed deterministic superposition named as the failure case above, and it is precisely the gate that the rest of the paper adopted: the identification $D_{\text{eff}} = exp(S_2)$ does not follow from the gate, and the discrepancy is not negligible in the regime that matters. The two readouts coincide only when the leaf mixture is uniform and diverge monotonically as it skews – for a two-leaf mixture $a = (0.9, 0.1)$, $exp(S_2) = 1.22$ against $PR(\hat{p}) = 1.02$ – and the bracketing contrast lives on the skewed side of that gap. Recovering $exp(S_2)$ requires the amplitude to scale as $\sqrt{a_\ell}$ rather than $a_\ell$, i.e., an amplitude code,

$$x = \Sigma_\ell \sqrt{a_\ell} \xi_\ell e_\ell,$$

with $e_\ell$ the atom directions and $\xi_\ell$ uncorrelated and of unit variance, so that the signal power along $e_\ell$ is $a_\ell$ and the across-repeat variance $\sigma_\ell = a_\ell$.

This offers an additional linking hypothesis about how the gate's mixture weights are expressed in population activity, not a property of the gate output. A gating code in which atoms are included with probability $a_\ell$ satisfies neither reading: it gives per-atom power $a_\ell(1 - a_\ell)$ rather than $a_\ell$.

The gating case is instructive here: for a two-leaf mixture $a = (p, 1 - p)$ it returns $\sigma = (p(1 - p), p(1 - p))$ and hence

$$D_{eff} = \frac{(\sum_k \sigma_k)^2}{\sum_k \sigma_k^2} = 2 \quad for\ every\ p, \quad whereas \quad exp(S_2) = \frac{1}{p^2 + (1 - p)^2}$$

varies from 1 to 2, so the two coincide only at $p = \frac{1}{2}$. We therefore state the amplitude reading as the premise the identification requires, which is checkable by regressing per-direction signal variance on decoded atom weight across the contrast set.

The estimator here must be per-trial, not cross-stimulus. $S_2(A\star)$ is a property of one tree's leaf mixture, and cannot be read off a covariance taken across trees; moreover, the mixture weights are simplex-constrained, $\sum_\ell a_\ell = 1$, which forces negative covariance between them across any ensemble, so the cross-stimulus covariance is not diagonal in the atom basis and its eigenvectors are not the atom directions. We therefore estimate the participation ratio within a condition, from the across-repeat signal variance along each independently identified atom direction: $\sigma_\ell$ is the variance of the projection onto $e_\ell$ over repeated presentations of the same tree, $\hat{p}_\ell = \sigma_\ell \,/\, \Sigma_k\, \sigma_k$, and the readout is $PR(\hat{p})$. Under the amplitude code above this recovers $\sigma_\ell = a_\ell$ and hence $D_{eff} = exp(S_2)$; the single-trial squared projection does not, since for the deterministic gate it is $a_\ell^2$ and for the amplitude code it is the noisy $a_\ell\, \xi_\ell^2$, and averaging that noise out is exactly what the across-repeat variance does.

This is a signal-side quantity, distinct from the trial-to-trial covariance at fixed input that defines the noise-side dimensionality of §5.3. In the dense, correlated semantic-similarity subspace neither condition holds and $D_{\text{eff}}$ conflates atom identity with mixture weight, so the estimator must be computed within the constituency-preserving subspace. The same population may then show opposite-signed excursions on the two axes – a signal-side rise in $D_{\text{eff}}$ as a constituent mixes more terminals, alongside the noise-side variance quench of §5.3 at closure.

Under the amplitude code, and only under it, the gate's second-order diversity term is directly observable as the signal-covariance participation ratio, so for matched-terminal, matched-statistics bracketing contrasts the effective dimensionality of the composing population should separate the two parses at the disambiguating region, with sign and magnitude fixed by the tree-recursive chain rule. Absent the amplitude code, the deterministic gate is still read out as a bracketing-dependent function of the same leaf mixture, $exp(3S_4 - 2S_2)$, so the qualitative separation of the two parses survives while the identification with $S_2$ does not; the associative null (vector addition) predicts a bracketing-blind dimensionality on either reading, and fixed mean-pooling predicts a shape-fixed one – $a_\ell = 2^{-\text{depth}(\ell)}$, hence identical $D_{\text{eff}}$ for any two bracketings with the same leaf-depth multiset, which every contrast in §4.9.1 has. A bracketing-dependent excursion of $D_{\text{eff}}$ on such a contrast is therefore the neural-data discriminator between the gate and a content-independent mixture.

A reliably nonzero, bracketing-dependent excursion of $D_{\text{eff}}(\lambda^\star)$ rules out the associative limit for the unnormalised gate of §4.9, whose associator vanishes at the Shannon point and scales linearly in $\alpha - 1$. It does not do so for the offset-corrected gate recommended in §4.9.1, whose associator is 1.58 at $\alpha \to 1$: that gate is bracketing-dependent at every order, and the $\alpha \neq 1$ inference is unavailable for it. The order is instead bounded from above, at $\alpha \leq 2$, by the convexity argument of §4.9, which applies to both gates because both minimise the same objective in $\lambda$.

The stronger claim – recovering the value of the brain's order $\hat{\alpha}$ – does not follow from the participation ratio alone, because $D_{\text{eff}}$ is intrinsically the $\alpha = 2$ diversity: reading it presupposes the very order one is trying to estimate. $\hat{\alpha}$ is instead an ordinary model-comparison parameter, estimated by fitting gate variants across a range of orders to held-out neural data (equivalently, by comparing the Rényi profile of the covariance eigenspectrum across orders against the $S_\alpha$-optimal mixtures), rather than by a single second-moment readout. We therefore treat $\alpha = 2$ as an analytically convenient default and $\alpha'$ as an empirically fitted quantity rather than one the participation ratio delivers directly. No $\alpha \neq 1$ test attaches to the corrected mixture or to Meld: that test belongs to the original thermodynamic-semiring operation (§4.9), and for the corrected mixture the order matters only through the higher-order separation of matched-depth-map trees (§4.9.1).

This approach would recast the dimensionality dynamics already reported in the literature during sentence integration – interpreted there as integration 'ramps' (Desbordes et al., 2023; Fedorenko et al., 2016; Woolnough et al., 2023) – as a structure-preserving signature, and a depth ladder $d = 1 \ldots 4$ then supplies the held-out data against which gate variants of differing order are compared. Because $S_2$ depends on $A$ only through $\sum_\ell a_\ell^2$, the 'Minimal Search' inversion left open by Marcolli and Berwick may be carried by the second-order (covariance) statistics of the composing population, potentially suggesting a subspace/eigen-readout for accessible-term extraction in place of analytic antipode inversion.

### *9.3. Explaining the interface mappings*

A major limitation of the present account is that we do not yet possess a comparably explicit model of externalization (the mapping from hierarchical syntactic structures to systems of articulation) or internalization (the mapping to conceptual systems). This problem is likely to be substantially more difficult than formalizing narrow syntax, since externalization incorporates language-specific word order, morphology, prosody, articulatory planning, perceptual constraints, and interacting sources of noise and variation. Nevertheless, such a model will ultimately be essential for explaining cross-linguistic differences in real-time comprehension and production. The discussion in §8 identifies some necessary ingredients – particularly the scheduling of structure-building operations, the use of linear precedence, and the reconstruction of workspace states from an externalized signal – but remains deliberately incomplete.

Both sensorimotor externalization and conceptual interpretation may be characterized, at least at a minimal level, in primarily *topological terms*: Articulatory, laryngeal, respiratory, and gestural instructions unfold as continuous trajectories through spaces of bodily configurations (Tourville & Guenther, 2011), while conceptual interpretation may involve navigation through structured spaces defined by proximity, continuity, and accessibility relations (Behrens et al., 2018; Constantinescu et al., 2016; Viganò et al., 2021). On this view, the interfaces need not reproduce the full Hopf-algebraic Markov dynamics required to construct, extract, and recombine syntactic objects. Rather, syntax may supply the principal combinatorial machinery, while the sensorimotor and conceptual systems transform its outputs into trajectories through their respective state spaces.

### *9.4. From admissibility to reconstruction: the inside-out complement*

The program developed here runs predominantly in one direction: from formal invariants to the neural dynamics those invariants render admissible. A complementary and ultimately co-equal arrow runs the other way, from the architecture of the cortical substrate to the space of linguistic computations that architecture can cheaply afford. This is the inside-out logic of systems neuroscience (Buzsáki, 2019), and it has a clear precedent in high-level vision, where the macro- and micro-anatomical layout of ventral temporal cortex furnishes a nested spatial hierarchy that acts as the neural infrastructure for the representational hierarchy of visual categories it supports (Grill-Spector & Weiner, 2014). Higher cognitive functions, on this view, are not implemented on an indifferent canvas but recruit, and are shaped by, the organization the brain already has. Applied to language, an independent characterization of the connectivity, laminar and oscillatory repertoire, and population geometry of the frontotemporal network would supply boundary conditions on how a Merge-based system could have come to operate within that pre-existing organization, and hence on which admissibility conditions of §§3-6 are biologically affordable rather than merely formally available.

The relationship here is, appropriately, recursive: structural facts about the substrate constrain inferences about the form language must take to be implementable in it, those inferences sharpen the experiments that probe how the substrate realizes that form, and the results refine our characterization of the substrate in turn. Naturally, this remains, for now, highly aspirational, since the inverse problem is radically under-determined and surface-statistical re-descriptions can masquerade as structure (Qian et al., 2024); it is effectively the equivariance, faithfulness, and falsifiability criteria of the NAP (Figure 13) that would allow such a reconstruction to be validated as genuinely linguistic rather than an arbitrary re-coordinatization of neural data.

### *9.5. The workspace transition system as an attractor network*

Another natural next step is to realize the Hopf-algebra Markov chain over workspaces (§4.8) explicitly within a family of graph-structured transient dynamical systems, rather than identifying it in advance with a single architecture. CTLNs remain attractive because their graph rules tightly constrain fixed points and sequential attractors, allowing the formal transition graph to be related directly to network architecture (Parmelee, Alvarez, et al., 2022). Stable heteroclinic channels and winnerless competition offer a complementary realization in which workspace states are metastable saddles and permitted operations are dynamically available transitions between them (Rabinovich et al., 2008). The latter may be especially natural for real-time language because robustness and sensitivity to informative perturbation are built into the same transient object, while CTLNs provide stronger graph-theoretic control of the attractor repertoire. The NAP need not decide between these architectures a priori: both should be required to approximate the same workspace-transition system, and can therefore be compared by their ability to recover its admissible edges, dwell-time structure, branching entropy, and perturbation responses. Under an SHC realization, for example, one testable linking hypothesis is that states with greater formal derivational branching should exhibit richer unstable local geometry, making the eigenspectrum of the corresponding metastable state a candidate neural correlate of the local branching entropy $H(x)$. The fixed-point structure of CTLNs is moreover captured by a simplicial complex via nerve theorems (Santander et al., 2021) which, together with the higher-order Hopfield store noted in §6.3, supplies a topological description of the workspace's attractor landscape that could ultimately feed the 'inside-out' reconstruction program (§9.4): the connectivity and attractor repertoire of the frontotemporal language network would then bound which admissibility conditions are biologically affordable rather than merely formally available.

### *9.6. Neural interoperability: first- and second-order admissibility*

Factorizing Merge creates a further problem of what we might think of as representational *interoperability*: how can operations implemented in different neural codes exchange computational objects without destroying the invariants those objects carry? On the account developed here, workspace storage and retrieval, derivational scheduling, binary binding, and sealing need not inhabit the same representational format or operate at the same neurophysiological scale (an intuition behind the ROSE model). A constituent may at one moment be maintained as a stable state of a content-addressable memory, at another be selected within graph-structured transient dynamics, at another provide an input to the population-level Meld operation, and subsequently acquire an oscillatory phase address that records its grouping history.

The explanatory target is therefore not simply a collection of individually plausible neural codes. Those codes must remain mutually usable. We refer to the system of admissible transformations that solves this problem as the *neural interoperability* architecture (Figure 14). The neural implementation of composition may consequently reside neither in a single code nor in the sum of several codes, but in the architecture that governs admissible transformations between them.

This distinction immediately suggests two levels of neural admissibility. What we will call *first-order admissibility* concerns whether an individual neural mechanism faithfully realizes the formal operation assigned to it. If a formal operation $M_i$ acts on objects in $S_i$, an encoding $\mathcal{N}_i: S_i \to X_i$ and physiological operation $G_i$ are first-order admissible when the corresponding diagram commutes, approximately,

$$\mathcal{N}_i \circ M_i \approx G_i \circ \mathcal{N}_i.$$

This is the form of admissibility developed throughout the present article, and which forms the backbone of the NAP: a workspace code must preserve addressability, a derivational sequencer must preserve the admissible transition graph, Meld must preserve the invariants of the binding operation, and the sealing code must preserve grouping information. *Second-order admissibility*, by contrast, concerns the transformations *between* such realizations. Let $X_i$ and $X_j$ be two neural state spaces realizing successive components of the computation, and let $I_{i \to j}$ specify the formal information that must survive their interface. A neural transformation

$$T_{i \to j}: X_i \to X_j$$

is second-order admissible only if

$$T_{i \to j} \circ \mathcal{N}_i \approx \mathcal{N}_j \circ I_{i \to j}.$$

Thus, first-order admissibility constrains the codes; second-order admissibility constrains their interfaces. A mechanism could satisfy its own local admissibility conditions and nevertheless fail as part of the language system if its output cannot be transformed into a state that the next mechanism can use while preserving the relevant computational object.

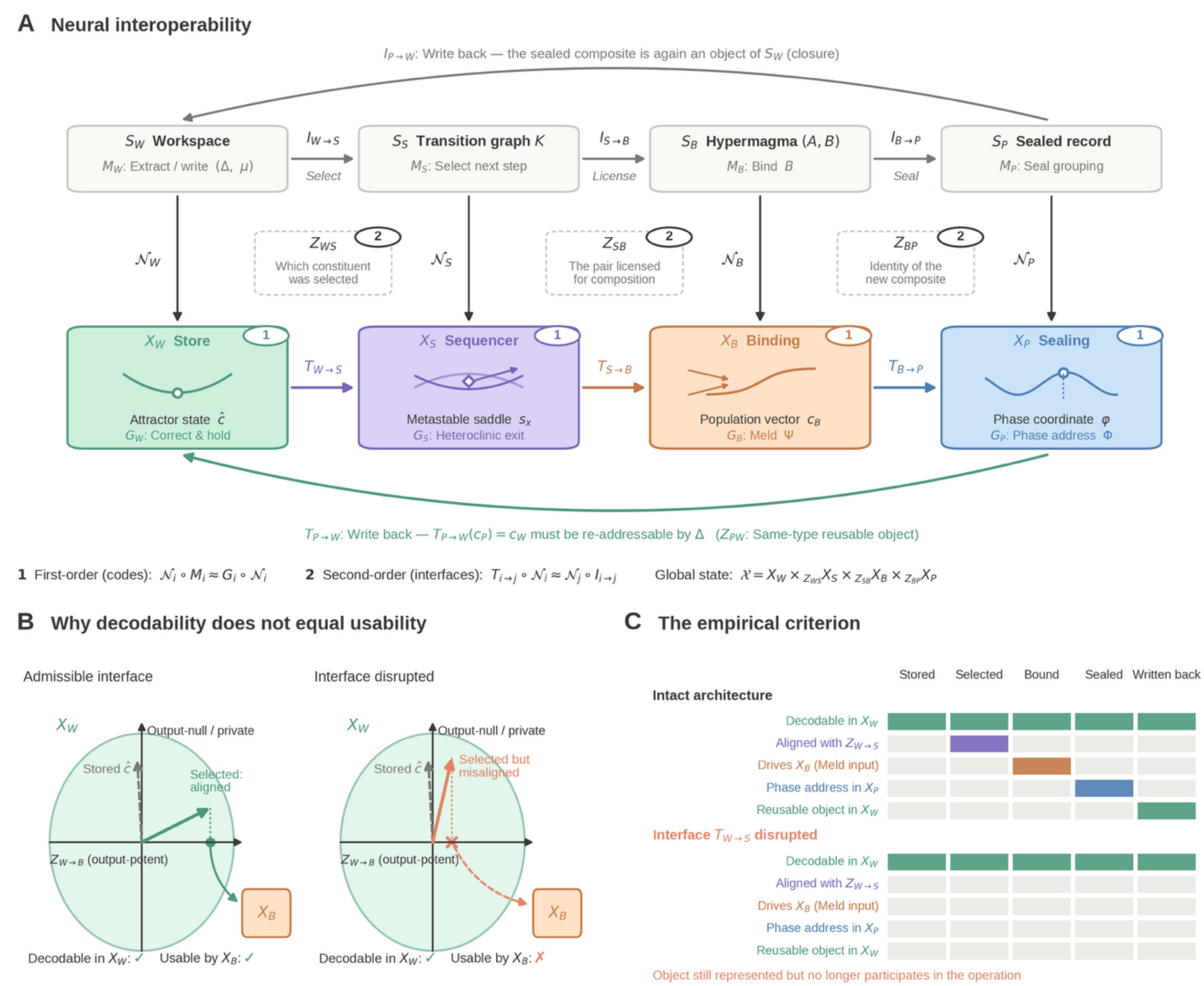

**Figure 14. An architecture for neural interoperability.** (A) The four factors of Merge as a commuting ladder: the formal layer $S_W \to S_S \to S_B \to S_P$ and the neural layer $X_W \to X_S \to X_B \to X_P$, joined by the encodings $\mathcal{N}_i$. First-order admissibility (①) is the condition on each code, $\mathcal{N}_i \circ M_i \approx G_i \circ \mathcal{N}_i$; second-order admissibility (②) is the condition on each interface, $T_{i\to j} \circ \mathcal{N}_i \approx \mathcal{N}_j \circ I_{i\to j}$. The dashed boxes $Z_{ij}$ hold the information on which neighboring mechanisms must agree; the two outer arcs are the write-back that closes the cycle (neural closure). The codes need not share a geometry, only a formally adequate interface, which is why the global state is the constrained product $\mathcal{X}$ rather than the Cartesian one. (B) Why decodability does not equal usability. The same constituent $\hat{c}$ is decodable from $X_W$ in both cases; only when its projection onto the shared subspace $Z_{W\to B}$ carries the invariant does $X_B$ receive the right object. One possible test is 'intersection information' – what is encoded in $X_i$ and read out by $X_j$ – rather than decodability (Panzeri et al., 2017). (C) The empirical criterion: decodable while stored, aligned with $Z_{W\to S}$ at selection, driving $X_B$ during binding, phase-addressed at sealing, re-addressable once written back. Disrupting a single interface leaves the first signature intact and removes all the others.

This issue is already implicit in the factorization of Merge developed above. Indeed, we have already noted how any binding law cannot be tested independently of the operations surrounding it. The same point applies more generally. The workspace store and the derivational sequencer may employ attractor-like representations, Meld may act over a population vector through a nonlinear rate-level transformation, and sealing may be expressed through low-frequency phase organization *à la* ROSE. None of these representations need to share a common geometry, but the selected constituent in the workspace must still be *that constituent* when it becomes an input to Meld.

The resulting problem is therefore stronger than multiscale coordination. Simultaneous activity at several scales is not sufficient, nor is cross-frequency coupling by itself. Neural interoperability requires that the state passed between scales preserve the invariants required by the downstream computation. Some existing results provide plausible ingredients for such an architecture. Inter-areal interactions can be confined to low-dimensional communication subspaces, such that only selected dimensions of one population influence another (Binish et al., 2026; Semedo et al., 2019). Such a subspace has now been identified directly in human intracranial recordings: activity confined to a prefrontal-to-motor communication subspace relays behaviorally relevant information at the single-trial level and predicts context-dependent action more strongly than either region alone (Binish et al., 2026). Related work distinguishes output-potent from output-null dimensions, allowing information to remain internally represented while being temporarily prevented from driving a downstream circuit (Kaufman et al., 2014). Oscillatory synchronization can dynamically regulate which pathways are effective (Fries, 2015), while cross-frequency coupling provides one possible bridge between slower control variables and faster local population computations (Canolty & Knight, 2010). Whether these two families of mechanism – rhythmic coordination and alignment to a communication subspace – are one process or two is an explicitly open question in the population-coding literature (Young et al., 2025). Recurrent systems can likewise reuse and switch among shared dynamical motifs under different contextual conditions (Driscoll et al., 2024). The present proposal places a stronger requirement on all such mechanisms: the dimensions made communicative at a given moment must be the dimensions that preserve the formally relevant object.

A possible mechanistic cycle follows directly. A syntactic object may remain stored in a content-addressable workspace while occupying dimensions that are output-null with respect to the binding population. Graph-structured transient dynamics then select the next admissible operation and place the relevant constituent dimensions into a downstream-potent communication subspace. The two selected daughter representations thereby become jointly available to the Meld circuit, whose nonlinear dynamics produce the composite. A sealing event records the resulting grouping state and may simultaneously alter which dimensions of that composite are downstream-potent, permitting the newly formed object to be written back into the workspace as an addressable state eligible for later extraction and composition. On this view, a linguistic derivation consists not only of transitions among workspace states but also of transitions among coding regimes: information is successively stored, selected, rendered communicable, transformed, sealed, and made available for reuse.

This also sharpens the neural meaning of closure. As is well-understood, in algebraic terms closure requires that the output of Merge be an object of the same formal type as its inputs. Biologically, however, the output of the binding operation need not initially inhabit the same neural code as the stored object that later participates in another Merge step. Neural closure therefore requires a cycle such as

$$X_W \xrightarrow{T_{W \to B}} X_B \xrightarrow{T_{B \to W}} X_W,$$

where $X_W$ denotes the workspace code and $X_B$ the binding-population code. If $c_B$ is the composite produced by Meld, then its write-back

$$T_{B\to W}(c_B) = c_W$$

must preserve exactly those properties that allow $c_W$ to function as the same syntactic object in subsequent computation. In this sense, neural closure is partly an interoperability condition: recursive reuse requires not only a closed local binding law, but successful passage of its output through the representational interfaces that return it to the workspace.

A useful mathematical description of the full architecture may therefore be richer than a Cartesian product of independent neural codes. Let

$$X_W,\ X_Q,\ X_B,\ X_\phi$$

denote, respectively, workspace memory, derivational sequencing, binding, and phase/sealing state spaces. The unconstrained product

$$X_W \times X_Q \times X_B \times X_\phi$$

permits arbitrary combinations of states, most of which would not correspond to a coherent derivational configuration. A more appropriate global object is a constrained product,

$$\mathcal{X}_{\text{interop}} = X_W \times_{Z_{WQ}} X_Q \times_{Z_{QB}} X_B \times_{Z_{B\phi}} X_\phi,$$

where each $Z_{ij}$ contains the information on which two neighboring mechanisms must agree. For example, the identity of the selected constituent, the pair currently licensed for composition, the identity of the newly produced composite, or the grouping event being sealed. The neural realizations on either side may be geometrically very different (an attractor state, a metastable saddle, a population vector, or an oscillatory phase coordinate), but they belong to the same global computational state only when their projections onto the shared interface agree.

This formulation leads to a different empirical criterion for multiscale composition. The strongest evidence for interoperability would not be that the same syntactic variable is independently decodable from several neural signals. It would be that information represented in one regime becomes selectively available to the next in the form required by the downstream operation. For example, constituent identity might remain decodable in the workspace while temporarily occupying output-null dimensions, become aligned with a downstream communication subspace at selection, influence the Meld population during binding, and then reappear in a sealed workspace state. Disrupting the interface should therefore produce a particularly diagnostic failure: the relevant representation could remain locally decodable while losing its computational usability downstream.

Fortunately, there is a ready-made statistic for this. Intersection information (Panzeri et al., 2017) quantifies the information about a variable that is both encoded in a neural response and read out to drive behavior. Substituting the state of the next code for behavior gives a direct measure of second-order admissibility,

$$II(z;\ X_i, X_j),$$

the information about the interface invariant $z$ that is present in $X_i$ and actually used by $X_j$.

Such experiments would extend the logic of the NAP from local mechanisms to global computational architectures. At the first level, one identifies an invariant and asks which neural operation preserves it (first-order admissibility). At the second, one asks whether that invariant survives the transformation by which one admissible neural implementation hands its result to another (second-order admissibility). Neural composition would then be established not by discovering *the* neural code for Merge, but by demonstrating a chain of formally faithful transformations through which heterogeneous neural codes jointly implement the computation. The relevant mechanism may therefore lie just as much in the admissible transformations between codes as in the discrete codes themselves.

### *9.7. Open questions: Reusability, addressability, inter-regional communication*

While the present discussion has focused on a core set of compositional syntactic-semantic features of language, a number of open questions remain. For example, how does a newly composed object become a reusable object of the same type? How do inactive constituents remain addressable? How are representations selectively exported between cortical areas?

Beyond CTLNs, expander Hopfield networks and phase-amplitude coupling, several contemporary circuit and population frameworks offer complementary realizations of the NAP that might aid in answering these and other questions. First, an *assembly calculus* can supply operations by which lexical assemblies could generate reusable composite assemblies (Papadimitriou et al., 2020), and in which sparse neuronal assemblies can be projected, associated and merged to generate new assemblies. Second, *short-term synaptic plasticity* permits completed but currently inactive constituents to remain available for later coproduct extraction (Mongillo et al., 2008), whereby short-term changes in synaptic efficacy can preserve recent information in an activity-silent or intermittently active state and permit its later reactivation. Third, *communication subspaces* provide a natural mechanism by which one selected workspace object can be exported to another cortical system while other objects remain locally active but output-null (Semedo et al., 2019); specifically, communication-subspace analyses show that only selected dimensions of a source population predict activity in a target population and that these dimensions need not coincide with the source's largest internal fluctuations. This would sharpen our presently discussed partial directed coherence (PDC) predictions (Murphy et al., 2026): PDC asks *whether and in which direction* areas interact; communication-subspace analyses ask *which population dimensions are actually transmitted*.

Each topic here remains subject to the same admissibility criterion: it must preserve the defining algebraic contrast, rather than merely improve prediction of sentence-level neural activity.

## 10. Discussion

> "The hidden harmony is better than the obvious."
> —Heraclitus (Fragment B123)

A proposed neural mechanism should be judged not merely by whether it predicts neural activity, but by whether its dynamics preserve the defining invariants of the computation it claims to implement. Richer feature annotations and more expressive regression models can sharpen localization, tighten control of confounds, and improve effect estimates. What they cannot do, by themselves, is explain how cortical circuits build language. That requires a mechanistic framework capable of generating testable linking hypotheses, not a longer list of predictors. The Neural Admissibility Program (NAP) is one such framework: instead of only asking where and when linguistic variables are encoded, or which artificial model best predicts neural activity, it asks which neural dynamics are admissible realizations of the invariants that define compositional structure.

Applying the NAP to the composition step of syntactic Merge yielded a number of surprising results. Beginning from a thermodynamic gate (the Marcolli–Berwick gate) that could not be implemented as stated, correcting it, and then asking what else satisfied the conditions that survived, we arrived at a law that the conditions themselves specify. We call it Meld:

$$Meld(u, v) = \tanh(\kappa W[\lambda \star u + (1 - \lambda \star)v]),$$

where we state

$$\lambda \star = \mathrm{argmin}_{\lambda \in [0,1]} \{\lambda u + (1 - \lambda)v - \frac{S_2(\lambda)}{\beta}\},\ S_2(\lambda) = -\log(\lambda^2 + (1 - \lambda)^2),$$

and where $u$ and $v$ are the two constituent states, $\lambda\star$ is the pointwise entropy-optimized mixture (a soft minimum: it rewards agreement between the daughters and penalizes disagreement, with $\beta$ setting how strongly), $W$ is a single set of shared synaptic weights through which both daughters pass, $\kappa$ is an output gain, and tanh is a saturating rate transfer that bounds the composite however deeply structures are nested.

Meld is, to our knowledge, the closest neurally plausible composition law to the binding step of Merge. It preserves every invariant the NAP demands, adds none the competence object (Merge) lacks, and does so with operations that are neurobiologically plausible. The laws that decode structure better keep information Merge discards; the laws that are equally plausible discard information Merge keeps.

Simulations indicated that an entropy-optimized mixture alone preserves grouping only within a band of commitment – committing too early to one constituent erases grouping, never committing reduces composition to averaging – and that passing the mixture through shared, saturating synapses removes the constraint: Meld recovers structure at every temperature and depth tested while remaining blind to the order of its parts, and the dimensionality prediction tracks its content-dependent weighting. Within the broader framework of the language sciences, Merge specifies what the computation must accomplish, while Meld is a candidate for how its binding step could be physically realized.

The simulation also exposes a broader problem with decoding-based model comparison. Wherever rival representational hypotheses are adjudicated by fitting each to neural data and backing whichever decodes best – presently standard practice in motor control, vision, reinforcement learning, and the neuroscience of language alike – the procedure presupposes that decoding accuracy is monotonically related to mechanistic correctness. It is not. A model can decode a target variable better precisely because it represents more than the computation contains. The composition comparison developed here is a worked example. The ordered role-filler law recovers bracketing more sharply than the entropy gate at every depth and signal-to-noise ratio tested. Part of that ordering at depth is gain, and the gap at the level where the contrast is formed is not; but the simulations establish two facts, not a mechanism. The role-filler law decodes bracketing better, and it preserves an order and address distinction that the competence object (Merge) discards, leaving daughter order decodable in the composite that the next composition step consumes. Whether the first fact is caused by the second would require ablating the order-bearing part of the code and showing the advantage disappears, which we have not done. What the result licenses is the following: the empirically superior decoder is nevertheless formally inadmissible as a direct realization of the completed Merge state – inadmissible not because order is represented somewhere in the system, which a real-time parser cannot avoid, but because it is represented there. And the margin is now small: Meld, which represents nothing the competence object lacks, is within a few points of it at every depth. The remedy to this is not better decoders, but a different objective. A model comparison is diagnostic only when the comparison space is restricted in advance to models satisfying the invariants of the computation, and when the discriminating measurement is the conjunction of what the correct model must recover and what it must fail to recover. Admissibility is a prior on the model space, not a posterior verdict on the fit. As such, any field within the neurosciences that races representational models on decodability requires a specification of what its correct model should be *unable* to do, and in the absence of one the race naturally selects for expressiveness rather than for mechanism. Linguistics is unusually well placed to supply such a specification, because its invariants are already stated as algebra.

Three results reported here change what should be measured, independently of whether the wider NAP is adopted. First, as above, composition laws do not separate on bracketing. What separates them is commutativity, and the discriminating measurement is a *conjunction*: bracketing decodable, daughter order at chance, and a bracketing separation on contrasts that a content-independent mixture cannot resolve – a corner that no additive, role-filler, recurrent or fixed-pooling law occupies, that the entropy gate occupies only within a window of temperature, and that Meld occupies at every temperature tested. Second, a neural signature earns the status of implementing an algebraic operation only by tracking that operation's defining contrast. A signature that varies with a coarse, monotone quantity (e.g., sentence processing 'ramping' effects) is a correlate or a consequence, irrespective of its frequency band, anatomical source, or recording scale. Third, $\Delta\varphi$ is not a free parameter. The β troughs are the addresses, so their number is the carrier ratio, and an angular code therefore does not degrade with depth at all below its bound. The widespread expectation that a phase code should decline gracefully with embedding depth is wrong, not merely imprecise, so the carrier ratio is what an experiment can actually vary ($\Delta\varphi$ follows from it). That result concerns the phase address as one coordinate of the grouping code; constituency itself is carried by phase address together with composite identity and sealing history (§5.1, §5.3), and it is the time-resolved composite, not the static phase map, that an experiment on grouping should decode.

A natural rival to Meld is role-filler binding. McCoy and colleagues show that network representations, including those of large language models, are well approximated by tensor-product representations, and that editing the recovered fillers changes behavior (McCoy et al., 2026). Our divergence is over what such a fit meaningfully shows. The role scheme is, of course, supplied by the analyst rather than generated by the composition operation, and in their own results the linear-position scheme beats the syntactic one, and so fidelity selects the scheme that carries least structure, just as decoding accuracy selects the law that carries most. In any event, fit is not admissibility. Moreover, a role-filler code composed along a path tags each terminal with a full tree address, leaving daughter order recoverable from the composed object, which is precisely what §3.1 disallows.

A second rival account is Martin's compositional neural architecture (Martin, 2020). Martin rejects tensor products and proposes instead that composition is additive, with grouping carried by desynchronization and phase sets. While we agree that grouping is written in time (§5.3), the additive law leaves nothing for the next step to consume: a phase set is a transient coordination among daughters, not a composite that can itself be a daughter, and a time-based code is bounded by the number of distinguishable phases (§5.1). Sealing is what the additive account needs and lacks. A commentary on Martin's model (Murphy, 2020a) objected to gain modulation specifically presented as an *unconstrained, all-purpose mechanism*, borrowed from sensorimotor physiology and applied as a single cascade from syllables to clauses, with predictions – low-frequency power increasing with structure – that would not uniquely support the model. Here, we placed one gain-like operation, sublinear integration followed by saturation, at one factor of Merge, as a consequence of admissibility, since these are the operations that survive the conditions the algebra of Merge imposes. The commentary also speculated that additive models might capture integration but that structural separation would require multiplicative modulation (Murphy, 2020a) – the present simulations bear this prediction out, since vector addition is associative, hence bracketing-blind, and performs at chance on every contrast, while the content-dependent weighting $\lambda\star$ of Meld is exactly the input-dependent modulation that the commentary suggested syntax would need.

A deeper divergence with Martin is over *what composition is*. Martin's DORA composes by vector addition and carries grouping in temporal asynchrony, so a phrase is a phase set: a transient coordination that keeps its daughters separable (Martin, 2020). Multiplexing of this kind is certainly important, and ROSE relies on it, but it is not composition – a phase set is not an object of the same type as its members, so there is nothing for the next step to consume and no recursion. Meld returns an object of the input type, and the seal writes it to a record from which the phase set's information is recoverable. Where DORA runs one mechanism through every level, the NAP factorizes Merge into extraction, selection, binding and reassembly, assigns each its own dynamics, and confines the gain-like operation to aspects of binding. This also fixes the two links the commentary found missing, since Meld is a population-level operation with a stated readout (§9.2) and its relation to oscillations follows from the factorization – Meld at ROSE's O level, the seal at S. Critically, none of this excludes role-filler binding, which DORA handles well and language requires; rather, it relocates it to the record the coproduct reads and the seal writes (§4.7, §5.3), the typed filters of §4.5, and the externalization layer of §8, rather than to the composite binding hands to the next Merge step. Indeed, since the sealing history is itself a sequence of role-tagged composites, the two accounts unify there rather than compete. Martin's one prediction unique to her model is that composition is additive rather than interactive; Meld predicts the opposite, a composite that depends on how far its constituents disagree at fixed sum (§7.5).

Turning to broader themes, it may be that the prior unavailability of a program like the NAP explains why many leading researchers have recently begun to conclude that the brain does not in fact compose linguistic information at all (Pylkkänen, 2019, 2026), due to the fact that the measures they have been restricted to, like scalp-derived event-related potentials/components and band-specific power fluctuations, do not always reliably offer composition-specific signatures. The NAP offers a different response to this difficulty: if the brain does compose hierarchical linguistic structure, then without the NAP it is difficult to pin down the precise form of composition and candidate neural mechanisms. Indeed, it is likely that syntactic processes have simply been overshadowed by a residue of more powerful signatures pertaining to semantics and brute-force statistical regularities. For instance, one of the issues raised here has been the possibility that constituency-preserving lexical representations may occupy a relatively disentangled subspace and that second- and higher-order population geometry could carry information not visible in ordinary activation analyses of the kind most contemporary neurolinguistic theories are built around (Friederici, 2017; Hagoort, 2005; Hickok, 2025). Relatedly, the phase-address model we discussed introduces a concrete performance bound from the carrier-frequency ratio, rather than leaving ‘phase synchronization’ as a vague implementation metaphor.

Crucially, NAP is not strictly tied to, nor identical with, Hopf algebra, colored operads, or function-space embeddings. These are current exemplars of the program because they generate unusually explicit admissibility conditions. Future work could investigate other formal feature spaces, depending on one’s philosophical bent. NAP should also be distinguished from several influential programs in the neurobiology of language. Some have made a powerful case that the core language network is a left-lateralized, modality-independent, and language-selective system distinct from perceptual, motor, and domain-general reasoning systems (Fedorenko et al., 2024). Others have supplied highly detailed neuroanatomical models of syntactic and semantic processing, organized around frontotemporal networks (Friederici, 2012, 2017; Hickok, 2025; Maran et al., 2022; Matchin & Hickok, 2020). Others have shown how formal descriptions of symbolic cognition (Martin, 2020), cultural recycling, and program complexity can guide the search for neural codes (Dehaene et al., 2015, 2022). These programs have been highly productive, and NAP provides what we see as a complementary next step for the field: changing the explanatory target.

Relatedly, the NAP should be understood as both a generalization and a stress test of the ROSE neurocomputational architecture for language (Murphy, 2024, 2025, 2026a), and other oscillatory phase codes for language (Kazanina & Tavano, 2023; Murphy, 2020b). ROSE proposes one specific multiscale implementation in which local feature codes, high-frequency bundling, low-frequency structural coordination, and inter-areal broadcasting jointly realize syntactic computation. The present framework moves one explanatory level upward: it asks which formal invariants any such implementation (including ROSE) must preserve, and uses those invariants to compare oscillatory, attractor-based, population-geometric, and other candidate mechanisms. ROSE (focusing on oscillatory regimes) therefore becomes a falsifiable member of a wider class of neurally admissible implementations.

Another consequence of introducing Meld, and of the multi-stage neural regime for Merge within which it sits, is a sharper placement of ROSE. In ROSE, the O level was reserved for elementary operations but specified no composition law – Meld now supplies one. Its inputs are R-level constituent ensembles, made co-active and kept distinct by spike-phase coupling at O, and its output is a new ensemble that can itself be a daughter. What ROSE identified as the phase-amplitude coupling signature of structure at S is then the seal (Figure 6B): the composite Meld returns is stamped with a low frequency (e.g., δ–θ) phase address, and the sequence of such stamps is the grouping record. ROSE thus occupies the operation and the closure of Merge, with the workspace and sequencer at E, and Meld adds to it a claim ROSE on its own could not make – that binding involves a rate-level nonlinearity in the target population, sensitive to disagreement between the constituents, which no phase code carries.

We also introduced a number of specific linking hypotheses, which we summarize below, whereby the algebraic invariants of Merge should be realized as measurable properties of neural dynamics. We stress that this is purely a proof-of-concept discussion, and the following naturally require further experimental exploration:

(i) We demonstrate that embedding depth can be carried by a slow-phase address code and grouping by the sequence of composites written at each seal, while $\gamma$-phase slots keep daughters distinct without encoding their surface order.
(ii) We show that head selection can be determined by differential $\gamma$-to-slow-phase coupling, and a low-frequency $\beta$-mediated 'sealing' operation converts the result into a stable, reusable constituent.
(iii) Hopf-algebraic workspace access predicts constituent-specific reactivation and directed inter-areal flow rather than a single accumulating sentence signal.
(iv) Colored phrase-structure and semantic-role filters predict a non-additive neural interaction, while head ambiguity predicts competition followed by stabilization.
(v) Derivations should appear as structured transitions between neural workspace states, and successful composition should be measurable in the effective dimensionality of the composing population, which should distinguish alternative bracketings even when lexical content, statistics, and surface order are controlled.
(vi) The binding step should be identifiable as Meld: a target population receiving both constituents through shared synapses, integrating them sublinearly and saturating. Its composite should carry bracketing with daughter order at chance in the constituency-preserving subspace and – most distinctively – should depend on the disagreement between the constituents at fixed summed drive, with target neurons tracking the smaller input where the two disagree.

The linking hypothesis that carries the most weight concerns the composition law. Neural composition may be implemented by locally heterogeneous or interdigitated populations expressing different algebraic profiles. Some populations may show the Meld signature: bracketing preserved, daughter order at chance in the constituency-preserving subspace, and a composite sensitive to disagreement between the constituents. Nearby populations may preserve bracketing and order together, consistent with ordered role-filler or recurrent composition; others may be order-blind and bracketing-sensitive but insensitive to disagreement, which is the signature of a saturating unit with tied daughter weights rather than of Meld; and others may show additive accumulation with no recoverable grouping. Alternatively, no recorded population may exhibit the Meld signature at all, and the binding step may be better explained by one of these rivals – a result the Neural Admissibility Criterion would count as evidence, since each rival is inadmissible or incomplete in a stated way.

A recently proposed function-space realization of Merge (Marcolli & Berwick, 2026) was shown here to be unimplementable as stated: every application leaves an entropy offset that is tree-shape dependent and diverges precisely in the regime where structure is best preserved. An offset-corrected mixture was proposed, and offers a better structural code, recovering bracketing while remaining blind to daughter order. That conjunction is a class rather than an individual signature: fixed mean-pooling and any saturating unit with tied daughter weights satisfy it too. Meld – the corrected mixture passed through shared saturating synapses – is the member of the class we propose as the neural binding operation: it satisfies every condition the NAP has produced here, has no temperature window, and decodes structure to within a few points of the ordered role-filler law at every depth. Merge is the formal operation; Meld is the proposed neural binding operation.

Importantly, our simulations do not directly overturn Marcolli and Berwick's central proof-of-concept that commutative, non-associative composition can be represented in a function space (Marcolli & Berwick, 2026). But they do, however, expose a recursive normalization problem in the particular thermodynamic output chosen for that representation. The entropy offset survives nesting, becomes tree-shape dependent, and invalidates the naïve high-temperature limit as a bounded neural code. Our offset-normalized entropy gate avoids that implausibility, performs substantially better under noise, and remains commutative and non-associative. The correction has a price, which we have stated as a result rather than as an open question: the corrected gate is idempotent, so it collapses self-merge, and it reduces to mean-pooling as $\beta \to 0$, where balanced four-leaf trees collapse as well. Boundedness is bought with faithfulness on the diagonal and in the high-temperature limit; what the gate keeps, at intermediate temperature, is a content-dependent mixture that no fixed pooling law has. It is not itself a thermodynamic-semiring addition, and its Hopf-algebraic compatibility remains to be proved. No pointwise translation-equivariant gate can be both bounded and faithful on the full magma, since boundedness in that class forces idempotence (§4.9.1); Meld is bounded and not idempotent, and pays in contraction with depth instead. Whether a bounded, non-idempotent gate faithful on the full magma without such contraction exists is open; such a law would supersede Meld on faithfulness, but it would still have to be built from circuit motifs.

Of course, a real-time parser is necessarily sensitive to arrival order, and so finding daughter-order decoding *somewhere* in the complete neural state does not disqualify a given model. The more acute prediction should concern a structural subspace, stabilized code, or quotient representation after order-dependent input dynamics have been factored out.

The case for Meld is not that it wins the comparisons above but what its proposed circuit is made of. Two inputs converging on a shared target population, integrated sublinearly and passed through a saturating rate nonlinearity are operations that cortex is known to perform (Carandini & Heeger, 2012; Ohshiro et al., 2011, 2017); what is new is the claim that this conjunction is what the binding step of natural language syntax requires, and that the admissibility conditions pick it out – commutativity is weight-sharing, non-associativity and content-dependence are the soft-minimum mixture, and boundedness at any gain is saturation. A law reached from algebraic constraints that lands on standard circuit components is a better candidate than one that had to be invented. Some important topics remain open, that we leave for future research: what the pointwise domain of the soft minimum corresponds to in a cortical population, how the shared weights are learned rather than drawn, and a proof of injectivity beyond the contrasts tested.

Read this way, Meld is a proposal about the granularity mismatch that Poeppel and Embick identified as the central obstacle to a neurobiology of language (Embick & Poeppel, 2015; Poeppel, 2012): a linguistic primitive, the binding step of Merge, matched to a circuit primitive, sublinear integration through shared synapses, at the same grain and with a stated operation between them. It is also a different kind of proposal from the view that structured cognition reduces to "compression" under a program-complexity prior (Dehaene et al., 2022). Compression describes what a system's representations *achieve* (Murphy, Holmes, et al., 2024); Meld describes an operation that produces them, and it predicts something compression alone does not – that the composite depends on the disagreement between the constituents it combines.

What are some other consequences of how we have illustrated the NAP? Merge, as formalized in §3.2 and §4.7, is not one operation but a composite of four: extraction of accessible terms by the coproduct $\Delta$, selection of the pair by $\delta$, binding of the selected pair by $B$, and reassembly of the workspace by $\mu$. The composition laws compared in §4.9.1 and §7.2 are candidates for $B$ alone. The other three factors are realized, on the account of §6.7, by different dynamics – extraction and reassembly by the content-addressable store and the coproduct read it supports (§4.7, §5.3), selection and scheduling by the threshold-linear sequencer and the algorithmic layer of §8, and closure by the sealing operator of §5.3. What the present work has found, then, is a candidate for *one factor* of Merge together with an explicit division of labor for the other three. Any claims about 'a neural signature of Merge' are well defined only relative to that factorization: a signal that tracks Merge is tracking a composite, and a signal that tracks binding is tracking $B$, in the state that $\mu$ hands to the next application of $\Delta$. Consequently, a binding law cannot be tested in isolation, because its inputs are produced by $\Delta$ and its output is consumed by $\mu$. The order-blindness test of §7.2 is a statement about the consumed state, and requires the constituency-preserving subspace to be fixed first. Relatedly, the dimensionality readout of §9.2 requires the atom directions to be identified independently, and the grouping prediction of §5.1 is time-resolved because it is read at the closure events that sealing writes. Testing the binding law therefore requires independent constraints on the other factors. A decoder trained only on the final state can conflate extraction, selection, binding, and reassembly, so superior fit cannot be assigned uniquely to $B$. Another consequence of this state of affairs is that the two ways in which the corrected gate falls short of the magma are absorbed by factors other than $B$: the diagonal is never presented to the gate, because copies are produced by $\Delta$ and re-merged by Internal Merge rather than by binding an object to itself, and the balanced four-leaf pair, which a static readout of $B$ cannot separate at the exploratory limit, is separated by the sealing sequence. The difficulty of finding a neural signature of Merge and the difficulty of testing a binding law that appears to comply with it are therefore the same difficulty: neither is well posed until the four factors are assigned to particular dynamics.

Returning to the two general research programs discussed earlier (Figure 1), we can now conclude that NAP is compatible with probabilistic and predictive neural computation: its claim is not that syntax is non-probabilistic, but that probability distributions and prediction errors must be defined over structured workspace states whose dynamics preserve non-associativity, typed compatibility, and access to constituent substructure. Furthermore, the NAP affords a number of novel predictions that go beyond localized activation maps – most notably, structure-preserving composition should be readable as the effective dimensionality of the composing population, which should separate the two bracketings of a matched-terminal contrast at the disambiguating region, where an additive null is bracketing-blind – a within-depth contrast, and not a divergence from an additive null with depth, which we show carries no grouping information (§4.9.1).

The theoretical trajectory we have pursued here begins with the observation that much of theoretical linguistics was first described using relatively minimal naïve set-theoretic machinery and tools from recursive function theory (Chomsky, 1957, 1995, 2014): syntactic objects were built by recursive combination, but the internal structure of that combination was left only coarsely specified (Gärtner, 2022). More recent formulations give this architecture sharper structural content (Marcolli, Chomsky, et al., 2025). This added precision is what makes the present neurobiological translation possible: linguistic operations can now be recast not as vague cognitive labels, but as admissibility conditions on neural dynamics. A central challenge is to show that neural structure is not merely covarying with a formal property, but is actually being exploited by the system (Krakauer & Ramsey, 2026). The NAP offers one route: identify the relevant invariant in advance, recover its neural realization, demonstrate that downstream computation depends specifically on that invariant, and then test its causal role through selective perturbation.

It should also be noted that the NAP is compatible with a fully pluralistic research agenda. We will likely develop only a partially adequate account of the neural foundations of language by focusing exclusively on one favored construct (e.g., 'mechanism', circuit, pathway, cascade, topology). While this may be ambitious, such pluralism will also serve to highlight the inherent limitations of each respective causal and non-causal structure. For example, although continuous attractor-manifolds seem capable of handling some particulars of workspace address systems (Gardner et al., 2022) (possibly capturing elements of phrasal embedding depth or attentional priority), at the same time a one-dimensional ring cannot by itself encode arbitrary branching structure, and might consequently represent a candidate implementation of the pointer, depth coordinate or present focus of composition, not of the whole tree-structure. Likewise, topology alone cannot *determine* linguistic interpretation; a torus or loop can arise for many reasons. The recovered neural structure must be aligned with independent formal contrasts.

The factorization outlined here also bears on which part of Merge is plausibly innate, and human-specific. Three of its four factors are capacity-like. A workspace that holds more objects, a sequencer that schedules more steps, and a closure operation that seals more constituents involve quantitative extensions of memory, scheduling and oscillatory coordination that other species likely possess, and an account on which human uniqueness consists in an expanded information capacity (Cantlon & Piantadosi, 2024) can in principle accommodate them. Yet this information-capacity account of human evolution may be unable to accommodate Meld. Non-associative, order-blind, disagreement-sensitive composition is not a 'quantity' of anything; a larger store of the same representations composed by addition remains bracketing-blind at any size, as §4.9.1 shows, and no amount of capacity turns an associative law into a non-associative one. If any factor of Merge is the core of what has been termed Universal Grammar – or within the neurosciences, Universal Neural Grammar (Murphy, 2025) – it is therefore the binding step, and Meld is a concrete hypothesis about its physical realization. The hypothesis also meets the usual demands of evolvability, since each of Meld's components – convergent afferents onto a common target, sublinear integration, rate saturation – is a conserved cortical motif present in nonhuman primates, so what would have to be new is not a mechanism but a re-routing: the tying of the two daughters' weights onto one target whose output can itself re-enter as a daughter. That is a small change to conserved parts with a large computational consequence. It is also testable comparatively. Sublinear integration and saturation should be found in the brains of non-human primates; a population that returns a composite of the input type, order-blind and disagreement-sensitive, and feeds it back as a constituent, should not. Meld is plausibly innate, in that the motif and its routing could be developmentally canalized: the shared weights W are learned, and what would be genetically fixed is the circuit that constrains what they can compute.

Importantly, this is also the sense of 'innate' that survives Quilty-Dunn and Wood's careful disambiguation of learning into subsymbolic parameter tuning and symbolic hypothesis testing (Quilty-Dunn & Wood, 2026): W is acquired by the first, and Meld is not acquired by the second, since it is the operation that makes there be compositional symbols to test hypotheses over. Their framework also leaves open under what conditions subsymbolic attractors fail to yield compositional structure, and §4.9.1 gives one condition: when the law the landscape settles on is associative, further tuning moves the weights and leaves bracketing unrecoverable.

Finally, the NAP has the potential to sharpen neurobiological assessments of the relation between language and other cognitive systems. Hierarchy, compatibility-constrained composition, structured retrieval, and recurrent workspace dynamics are not unique to language. Related organizations occur in program execution and theorem proving, hierarchical action control, visual scene parsing, music, and event cognition. Some of these domains also preserve non-associative grouping distinctions, although not necessarily through the same primitive algebra. What may distinguish language is therefore not any single property in isolation, but the particular conjunction of open-ended lexical generation, non-associative constituency, typed interface filtering, structured access to subterms, and modality-independent externalization. By requiring such comparisons to be stated in terms of formal invariants and admissible neural mechanisms, the NAP can transform vague claims about the (non-)uniqueness of language into precise and falsifiable hypotheses. The decisive question is not simply whether language is special, but which neural operations make its formal architecture possible – and which of those operations it shares with the rest of cognition.

## 11. Conclusion

The search for the neural code for linguistic composition has been handed a deceptively simple algebraic target in Merge. Yet, viewed as a dynamical system, that one elegant operation resolves into contributions from regimes that look nothing like one another: oscillatory coordination, a content-addressable workspace, attractor dynamics over a structured state space, population geometry, and a specific neural binding operation that we term Meld. The NAP provides a principle by which these disparate mechanisms can be unified: a common criterion of admissibility specifying what each must preserve, and therefore what role each can play in realizing the computation. In this sense, the NAP proceeds with the familiar business of "reducing complex visibles to simple invisibles" – in this instance, allowing mathematical invariants to reveal the mechanistic order beneath heterogeneous neural dynamics.

**Acknowledgements**: This work would not be possible without the contributions of three figures to the modern language sciences, each of whom contributed to the intellectual backbone of the NAP, and who over the years have offered generous, kind, and extremely valuable guidance through correspondence. First, Noam Chomsky. Since 2010, his correspondence with me on issues relating to the neurobiology of language has helped constrain my sense of what an explanatory account of neurolinguistics must achieve. I have a tendency to favor late-career works of artists over their earlier work (e.g., Lynch, Dostoevsky, Tarkovsky), and Chomsky is no exception. His most recent theory articulated since ~2013 offers a substantially more elegant and cognitively plausible set of foundations for language, and has been invaluable for helping me articulate the very notion of formal linguistic properties. Second, Matilde Marcolli. Her work on mathematical linguistics offers a comprehensive and exciting new approach to formalizing theoretical syntax. In June 2025, Marcolli pushed me to formalize my ROSE model within a much more explicit mathematical framework. Although it has taken me over a year to deliver on that challenge, the present work has also developed into a much broader project, and I am extremely grateful to her for pushing me in this direction. And third, David Poeppel. Marcolli provided a more recent form of impetus to expand my work, but it was Poeppel who originally inspired me to move towards neurolinguistics. In 2012, I read 'The maps problem and the mapping problem' shortly after it appeared, which introduced me to the fascinating conceptual gaps separating linguistics and neuroscience. I hope that the present work offers testable directions toward resolving some of the problems that Poeppel's writings have so clearly articulated for the field.

**Disclosure of interest**: The author reports there are no competing interests to declare.

**Code availability**: All scripts used to run the simulations reported in this work are available on GitHub (github.com/ElliotMurphy91/NeuralAdmissibilityProgram).

**Declaration of generative AI and AI-assisted technologies in the manuscript preparation process**: During the preparation of this work, the author used Claude Opus 5 (Anthropic) to assist in checking the notation and internal consistency of the equations and to generate initial drafts of scripts used for simulations. The author reviewed, verified, and edited all AI-assisted output and takes full responsibility for the content of the work.